\documentclass{article}
\usepackage{graphicx,natbib}
\usepackage[noend]{algpseudocode}
\usepackage{lineno}
\usepackage{algorithmicx,algorithm}
\usepackage{geometry}
\usepackage{color,graphicx}
\usepackage{subfigure}
\usepackage[section]{placeins}
\usepackage{changepage}
\usepackage{amssymb}
\usepackage{amsmath}
\usepackage{mathrsfs}
\usepackage{amsthm}
\usepackage{array}
\usepackage{mathabx} 

\usepackage{multirow}

\newcommand{\tabincell}[2]{\begin{tabular}{@{}#1@{}}#2\end{tabular}}

\begin{document}

\title{Influence of Initialization on the Performance of Metaheuristic Optimizers}
\author{Qian Li$^{a,*}$ and San-Yang Liu$^{a}$ and Xin-She Yang$^{b}$ \\
$^{a}$School of Mathematics and Statistics, \\
Xidian University, Xi'an, Shaanxi 710071, P.R. China \\
$^{b}$School of Science and Technology,\\
 Middlesex University, London NW4 4BT, UK \\
*Corresponding author}

\date{}

\maketitle

\noindent {\bf Citation Detail:} \\
\noindent Qian Li, San-Yang Liu, Xin-She Yang,
Influence of initialization on the performance of metaheuristic optimizers,
Applied Soft Computing, 2020, Volume 91, Article 106193. \\

\noindent https://doi.org/10.1016/j.asoc.2020.106193 \\
\hrule

\begin{abstract}

All metaheuristic optimization algorithms require some initialization, and the initialization for such optimizers is usually carried out randomly. However, initialization can have some significant influence on the performance of such algorithms. {{This paper presents a systematic comparison of 22 different initialization methods on the convergence and accuracy of five optimizers: differential evolution (DE), particle swarm optimization (PSO), cuckoo search (CS), artificial bee colony (ABC) algorithm and genetic algorithm (GA). We have used 19 different test functions with different properties and modalities to compare the possible effects of initialization, population sizes and the numbers of iterations. Rigorous statistical ranking tests indicate that 43.37\% of the functions using the DE algorithm show significant differences for different initialization methods, while 73.68\% of the functions using
 both PSO and CS algorithms are significantly affected by different initialization methods. The simulations show that DE is less sensitive to initialization, while both PSO and CS are more sensitive to initialization. In addition, under the condition of the same maximum number of function evaluations (FEs), the population size can also have a strong effect. Particle swarm optimization usually requires a larger population, while the cuckoo search needs only a small population size. Differential evolution depends more heavily on the number of iterations, a relatively small population with more iterations
can lead to better results. Furthermore, ABC is more sensitive to initialization, while such initialization has little effect on GA. }}
Some probability distributions such as the beta distribution, exponential distribution and Rayleigh distribution can usually lead to better performance. The implications of this study and further research topics are also discussed in detail.

\end{abstract}

\section*{Acronyms}

\begin{tabular}{ll}
CEC & Congress of Evolutionary Computation \\
CS & Cuckoo Search \\
DE & Differential Evolution \\
DE-a & Adaptive Variant of DE \\
GA & Genetic Algorithm \\
LHS & Latin Hypercube Sampling \\
PSO & Particle Swarm Optimization \\
PSO-w & PSO with an Inertia Weight \\
\end{tabular}

\section{Introduction}
\label{S:1}

{{
Many real-world optimization problems are very complex, subject to multiple nonlinear constraints.
 Such nonlinearity and multimodality can cause difficulties in solving these optimization problems. Both empirical observations and numerical simulations suggest that the final solution may depend on the initial starting points for multimodal optimization problems~\citep{Yang2018, eskandar2012water}. This is especially true for gradient-based methods. In addition, for problems with non-smooth objective functions and constraints, gradient information may not be available.}} Hence, most traditional optimization methods struggle to cope with such challenging issues. {{A good alternative is to use metaheuristic optimization algorithms, such as particle swarm optimization (PSO) and cuckoo search (CS). These metaheuristic optimizers are gradient-free optimizers, which do not require any prior knowledge or rigorous mathematical properties, such as continuity and smoothness~\citep{Yang2018,li2016novel}.}}

In the past decade, various studies have shown that these metaheuristic algorithms are effective in solving different types of optimization problems, including noisy and dynamic problems~\citep{Yang2018,sun2019hybrid, fan2018auto, cheng2018improved}. For example, engineering design problems can be solved by an improved variant of the PSO~\citep{isiet2019self} and the connectivity of the internet of things (IoT) can be enhanced by a multi-swarm optimization algorithm~\citep{hasan2019optimization}. In addition, the optimized energy consumption model for smart homes can be achieved by differential evolution (DE)~\citep{essiet2019optimized}, while the optimal dam and reservoir operation can be achieved by a hybrid of the bat algorithm (BA) and PSO~\citep{yaseen2019hybrid}.  {{A fuzzy-driven genetic algorithm~\citep{jacob2009fuzzy} was used to solved a sequence segmentation problem, and a fuzzy genetic clustering algorithm was used to solve a dataset partition problem~\citep{nguyen2019partition}.}}

Almost all algorithms for optimization require some forms of initialization, where some educated guess or random initial solutions are generated. Ideally, the final optimal solutions found by algorithms should  be independent on their initial choices. This is only true for a few special cases such as  linear programs and convex optimization; however, a vast majority of problems are not linear or convex, thus such dependency can be a challenging issue. In fact, most algorithms will have different degrees of dependency on their initial setting, and the actual dependency can be problem-specific and algorithm-specific~\cite{Yang2014SI,kondamadugula2016accelerated}. For large-scale and multimodal problems, the effect of initialization is more obvious, and many algorithms may show differences in the probability of finding global optima on different initialization~\citep{ElsayedSequence}.

However, it still lacks a systematical study of initialization and how the initial distributions may affect the performance of algorithms under a given set of problems. The good news is that researchers start to realize the importance of initialization and have started to explore other possibilities with the aim to increase the diversity of the initial population~\citep{Yang2014SI}. For example, based on the guiding principle of covering the search space as uniformly as possible, some studies have preliminarily explored certain ideas of different
initialization methods, including quasi-random initialization~\citep{kimura2005genetic, ma2012impact, kazimipour2014review, maaranen2004quasi}, chaotic systems~\citep{Gao2012A, alatas2010chaotic},  anti-symmetric learning methods~\citep{rahnamayan2008opposition}, and Latin hypercube sampling~\citep{Ran2017Evolutionary, Zaman2016Evolutionary}. In some cases, these studies have improved the performance of algorithms such as PSO and genetic algorithms (GA), but there are still some serious issues. Specifically, quasi-random initialization is simple and easy to implement, but it suffers from the curse of dimensionality~\citep{maaranen2004quasi}; for chaos-based approaches, random sequences are generated by a few chaotic maps and fewer parameters (initial conditions), but they can inevitably have very sensitive dependence upon their initial conditions under certain conditions~\citep{dos2008use}. In addition, in the anti-symmetric learning method, twice the number of the population as the solution cohorts are used so as to select the solutions for the next generation, which doubles the computational cost.
Though the Latin hypercube sampling is
very effective at low dimensions, its performance can deteriorate significantly for
higher-dimensional problems. We will discuss this issue in more detail later in this paper.

On the other hand, some researchers attempted to design some specific type of initialization in combination with a certain type of algorithm so as to solve a particular type of problems more efficiently. For example, Kondamadugula et al.~\citep{kondamadugula2016accelerated} used a special sampling evolutionary algorithm and random sampling evolutionary algorithm to estimate parameters concerning digital integrated circuits; Li et al.~\citep{li2015knowledge} applied knowledge-based initialization to improve the performance of the genetic algorithm for solving the traveling salesman problem; Li et al.~\citep{Qian2019eswa} used the degrees of nodes to initialization for network disintegration problem, and Puralachetty et al.~\citep{puralachetty2016differential} proposed a two-stage initialization approach for a PID controller tuning in a coupled tank-liquid system.
However, these approaches do have some drawbacks. Firstly, such initialization requires sophisticated allocation of points, which may not be straightforward to implement and can thus increase the computational costs. Secondly, they may be suitable only for a particular type of problems or algorithms. Thirdly, such initialization is largely dependent on the experience of the user. Finally, there is no mathematical guidance about the ways of initialization in practice.

{{This motivates us to carry out a systematic study of different initialization methods and their effects on the algorithmic performance. The choice of 22 probability distributions are based on rigorous probability theory with the emphasis on different statistical properties. In addition, we have used five different metaheuristic optimization algorithms for this study, and they are differential evolution (DE), particle swarm optimization (PSO), cuckoo search (CS), artificial bee colony (ABC) algorithm and genetic algorithm (GA). There are over 100 different algorithms and variants in the literature~\citep{Yang2018,eskandar2012water,Zaman2016Evolutionary}, it is not possible to compare a good fraction of these algorithms. Therefore, the choice of algorithm has to focus on different search characteristics and representativeness of algorithms in the current literature. Differential evolution is a good representative of evolutionary algorithms, while particle swarm optimization is considered as the main optimizer of swarm intelligence based algorithms. In addition, the cuckoo search uses a long-tailed, L\'evy flights-based search mechanism that has been shown to be more efficient in exploring the search space.
Furthermore, artificial bee colony is used to represent the bee-based algorithms, while the genetic algorithm has been considered as a cornerstone for a vast majority of evolutionary algorithms.}}

Based on the simulations and analyses below, we can highlight the features and contributions of this paper as follows:

\begin{enumerate}
  \item Numerical experiments show that, under the same condition of the maximum number of fitness evaluations(FEs), some algorithms require a large number of populations to reach the optimal solution, while others can find the optimal solution through multiple iterations under a small number of populations. In this paper, we make some recommendations concerning the number of the initial population and the maximum number of iterations of the five algorithms.
  \item The initialization of 22 different probability distributions and their influence on the performance of the algorithm are studied systematically. It is found that some algorithms such as the differential evolution are not significantly affected by initialization, while others such as the particle swarm optimization are more sensitive to initialization. This may be related to the design mechanisms of these algorithms themselves, which is also an important indicator to measure the robustness of algorithms.
  \item For the five algorithms under consideration, we have used a statistical ranking technique, together with a correlation test, to gain insight into the appropriate initialization methods for given benchmark functions.

\end{enumerate}

Therefore, the rest of this paper is organized as follows. Section 2 briefly introduces the fundamentals of the three metaheuristic optimizers with some brief discussions of the other two optimizers, followed by the discussion of motivations and  details of initialization methods in Section 3. {{Experimental results are presented in Section 4, together with the comparison of different initialization methods on some benchmark functions, including commonly used benchmarks and some recent CEC functions. Further experiments concerning key parameters of different algorithms are also carried out. Then, Section 5 discusses the correlation between the distributions of the initial population and their corresponding final solutions. Finally, Section 6 concludes with discussions about further research directions.}}

\section{Metaheuristic Optimizers}

{{Though traditional optimization algorithms can work well for local search, metaheuristic optimization algorithms have some main advantages for global optimization because they usually treat the problem as a black-box and thus can be flexible and easy to use~\citep{Yang2014}.}} Furthermore, such optimizers do not have strict mathematical requirements (e.g., differentiability, smoothness), so they are suitable for problems with different properties, including discontinuities and nonlinearity. Various studies have shown their effectiveness in different applications~\citep{Yang2014, aljarah2020multi, yin2019integrated}.

{{The initialization of a vast majority of metaheuristic optimization algorithms has been done by using uniform distributions. Although this approach is easy to implement, empirical observations suggest that uniform distributions may not be the best option in all applications. It is highly needed to study initialization systematically using different probability distributions. As there are many optimization algorithms, it is not possible to study all of them. Thus, this paper will focus on five algorithms: differential evolution (DE), particle swarm optimization (PSO), cuckoo search (CS), artificial bee colony (ABC) and genetic algorithm (GA). These algorithms are representative, due to the different search mechanisms and their richer characteristics.}}

\subsection{Differential Evolution}

Differential evolution (DE) is a representative evolutionary and heuristic algorithm~\citep{Storn1997}, which has been used in many applications such as optimization, machine learning and pattern recognition~\citep{liu2005fuzzy}.
Though differential evolution has a strong global search capability with a relatively high convergence rate for unimodal problems, the performance of DE can depend on its parameter setting. For highly nonlinear problems, its convergence rate can be low. {{To overcome such limitations, various mutation strategies and adaptive parameter control for $F$ have been proposed to improve its performance\citep{Zhang2009JADE}. In the DE algorithm, each individual is a candidate solution or a point in the $D$-dimensional search space, and the $i$-th individual can be represented as
${x_i} = ({x_{i,1}},{x_{i,2}}, \cdots ,{x_{i,D}})$.}} In essence, different mutation strategies typically generate a mutation vector $({v_{i,1}},{v_{i,2}}, \cdots  \cdots ,{v_{i,D}})$ by modifying the current solution vector in different ways.

Crossover is another strategy of modifying a solution. For example, the binomial crossover is a component-wise modification, controlled by a crossover parameter $CR$, which takes the following form:
\begin{equation}\label{rand2}
{u_{i,j}} = \left\{ \begin{array}{lll}
{v_{i,j}}, & {\rm{if  }} \quad \textrm{rand}(0,1) < CR & {\rm{or }} \quad j = {j_{rand}},\\
{x_{i,j}}, & \textrm{otherwise,} \end{array} \right.
\end{equation}
where ${x_{i,j}}$ is the $j$-th dimension of the $i$-th individual solution. The updated vector can be  expressed as ${v_{i,j}}$ after the mutation step, and  ${u_{i,j}}$ corresponds to the $j$-th dimension of the $i$-th individual after crossover.

Among various variants of DE, Qin et al.~\citep{Adaptation2009} proposed a self-adaptive DE (SaDE) variant with four mutation strategies in its pool, which can be selected at different generations by a given criterion.
More specifically, according to the success and failure of each mutation, a fixed learning period (LP)
was used to update the probability of each mutation strategy being selected for the next generation.
In addition, $F$ was drawn from a normal distribution with a mean of $0.5$ and standard deviation of $0.3$; that is $N(0.5,0.3^2)$. Similarly, $CR$ was drawn from  a normal distribution $N(CR{m_k},0.1)$, where $CR{m_k}$ was calculated from previous LP generations. Though the performance of SaDE was good, its complexity had increased.

For the ease of implementation and comparison in this paper, we use a simplified adaptive DE (DE-a).
Based on the idea of the SaDE algorithm, a simple adaptive DE (DE-a) algorithm is proposed in this paper.
In the mutation pool, we use five mutation strategies as follows:
\begin{itemize}

\item DE/rand/1~\citep{Storn1997}
\begin{equation}\label{rand1}
{v_{i,j}} = {x_{{r_1},j}} + F \cdot ({x_{{r_2},j}} - {x_{{r_3},j}}),
\end{equation}

\item DE/best/1
\begin{equation}\label{best1}
{v_{i,j}} = {x_{best,j}} + F \cdot ({x_{{r_1},j}} - {x_{{r_2},j}}).
\end{equation}

\item DE/current-to-best/1~\citep{Zhang2009JADE}
\begin{equation}\label{ctobest1}
{v_{i,j}} = {x_{i,j}} + F \cdot ({x_{best,j}} - {x_{i,j}}) + F \cdot ({x_{{r_2},j}} - {x_{{r_3},j}}).
\end{equation}

\item DE/best/2
\begin{equation}
v_{i,j}=x_{best,j} + F \cdot (x_{r_1,j}-x_{r_2,j}) + F \cdot (x_{r_3,j}-x_{r_4,j}).
\end{equation}

\item DE/rand/2
\begin{equation}
v_{i,j}=x_{r_1,j} + F \cdot (x_{r_2,j}-x_{r_3,j}) + F \cdot (x_{r_4,j}-x_{r_5,j}).
\end{equation}

\end{itemize}
where $F \in [0,2]$ is a parameter for mutation strength, and ${x_{best,j}}$ is the $j$-th dimension of the current best solution. Here, $x_{r_1,j}$, $x_{r_2,j}$, $x_{r_3,j}$, $x_{r_4,j}$ and $x_{r_5,j}$ represent 5 different individuals, which are selected randomly from the current population.

Both parameters $CR$ and $F$ are initialized to a set of discrete values. That is, $CR \in [0.4,0.5,0.6,0.7,0.8]$ and $F \in [0.5,0.6,0.7,0.8,0.9]$. The current mutation strategy and parameter settings are not updated if better solutions are found during the iterations. Otherwise, mutation strategies and parameters are randomly selected from the above sets or ranges. Our simplified variant becomes easier to implement and the performance is much better than the original DE, as observed from our simulations later. Therefore, we will use this variant for later simulations.

\subsection{Particle Swarm Optimization}

{{Particle swarm optimization (PSO) is a well-known swarm intelligence optimizer with good convergence~\citep{Clerc2002}, which is widely used in many applications~\citep{Kennedy2011Particle}.}} However, it can have premature convergence for some problems, and thus various variants have been developed to remedy it with different degrees of improvement. {{Among different variants, an improved PSO with an inertia weight (PSO-w), proposed by Shi and Eberhart~\citep{shi1998modified}, is efficient and its main steps can be summarized as the following update equations:}}

\begin{equation}\label{pso1}
\begin{array}{l}
v_i^{t + 1} = w \cdot v_i^t + {c_1} r_1^t (p_i^t - x_i^t) + {c_2} r_2^t (p_g^t - x_i^t)
\end{array}
\end{equation}
\begin{equation}\label{pso2}
\begin{array}{l}
x_i^{t + 1} = x_i^t + v_i^{t + 1}
\end{array}
\end{equation}
where $v_i^{t}$ and $x_i^{t}$ are the velocity vector and
position vector, respectively, for particle $i$ at iteration $t$.
{{Here, $p_i^{t}$ is the individual best solution of $i$-th individual in the previous $t$ iterations, and $p_g^{t}$ is the best solution of the current population.
In Eq.~(\ref{pso1}), $c_1$ and $c_2$ are the two learning parameters,
while $r_1^{t}$ and $r_2^{t}$ are two random numbers at the current iteration, drawn from a uniform distribution.}} In a special case when the inertia weight $w=1$, this variant becomes the original PSO.

The value of $w$ can affect the convergence rate significantly. If $w$ is large, the algorithm can have a faster convergence rate, but it can easily fall into local optima, leading to premature convergence. Studies showed that a dynamically adjusted $w$ with iteration $t$ can be more effective. That is
\begin{equation}\label{weight}
w = {w_{\max }} - \frac{{({w_{\max }} - {w_{\min }}) \cdot t}}{{{T_{\max }}}}
\end{equation}
where ${T_{\max }}$ represents the maximum number of iterations, ${w_{\min }}$ and ${w_{\max }}$ are the minimum inertia weight and the maximum inertia weight, respectively. we will use PSO-w in the later experiments.

\subsection{Cuckoo Search}

Cuckoo search (CS) algorithm is a metaheuristic algorithm, developed by Xin-She Yang and Suash Deb~\citep{yang2009cuckoo}, which was based on the behavior of some cuckoo species and their interactions with host species in terms of brooding parasitism. CS also uses L\'evy flights instead of isotropic random walks, which can explore large search spaces more efficiently. As a result, CS has been applied in many applications such as engineering design~\citep{Gandomi2013Cuckoo}, neural networks~\citep{vazquez2011training}, semantic Web service composition~\citep{chifu2011optimizing}, thermodynamic calculations~\citep{Bhargava2013Cuckoo} and so on.

Briefly speaking, the CS algorithm consists of two parts: local search and global search. The current individual $x_i^t$ is modified to a new solution $x_i^{t + 1}$ by using the following global random walk:
\begin{equation}\label{cuckoo1}
x_i^{t + 1} = x_i^t + \alpha  \oplus L(s, \lambda),
\end{equation}
where $\alpha $ is a factor controlling step sizes, and $s$ is the step size. $L$ is a random vector drawn from a L\'evy distribution~\citep{Yang2014}. That is
\begin{equation}\label{cuckoo2}
L(s,\lambda ) \sim \frac{{\lambda \Gamma (\lambda )\sin (\pi \lambda /2)}}{\pi } * \frac{1}{{{s^{1 + \lambda }}}}
\end{equation}
Here, `$\sim$' means that $L$ is drawn as a random-number generator from the distribution on the right-hand of the equation. $\Gamma$ is the Gamma function, while $1<\lambda \le 3$ is a parameter. {{One of the advantages of using L\'evy flights is that it has a small probability of long jumps, which enables the algorithm to escape from any local optima and thus increases its exploration capability~\citep{Yang2018,Viswanathan1999Optimizing}.}} The local search is mainly carried out by
\begin{equation}\label{CS}
x_i^{t + 1} = x_i^t + \alpha s \otimes H(p_a - \varepsilon ) \otimes (x_j^t - x_k^t)
\end{equation}
where $H(u)$ is the Heaviside function. This equation modifies the solution $x_i^t$ using two
other solutions $x_j^t$ and $x_k^t$. Here, the random number $\varepsilon$ is drawn from a uniform distribution and $s$ is the step size. A switching probability $p_a$ is used to switch between these two search mechanisms, intending to balance global search and local search.

\subsection{Other Optimizers}
{{
There are other optimizers that can be representative for the purpose of comparison.
The genetic algorithm (GA) has been a cornerstone of almost all modern evolutionary algorithms,
which consists of crossover, mutation and selection mechanisms. The GA has a wide range of applications
such as pattern recognition~\citep{pal2017genetic}, neural networks and control system optimization~\citep{back1996evolutionary} as well as discrete optimization problems~\citep{guerrero2017adaptive}. The literature on this algorithm is vast, thus we will not introduce it in detail here.

On the other hand, the artificial bee colony (ABC) algorithm was inspired by foraging behaviour of honey bees~\citep{karaboga2005idea}, and this algorithm has been applied in many applications~\citep{li2017artificial,gao2018artificial,gao2010sne}. A multi-objective version also
exists~\citep{xiang2015elitism}. Due to the page limit, we will not introduce this algorithm in detail. Readers can refer to the relevant literature~\citep{karaboga2007advances}. }}

We will use the above five algorithms in this paper for different initialization strategies.

\section{Initialization Methods}

The main objective of this paper is to investigate different probability distributions for initialization and their effects on the performance of the algorithms used.

\subsection{Motivations of this work}

Both existing studies and empirical observations suggest that initialization can play an important role in the convergence speed and accuracy of certain algorithms. A good set of initial solutions, especially, when the initial solutions that are near the true optimality by chance, can reduce the search efforts and thus increase the probability of finding the true optimality. As the location of the true optimality is unknown in advance, initialization is largely uniform in a similar manner as those for Monte Carlo simulations. However, for problems in higher dimensions, a small initial population may be biased and could lie sparsely in unpromising regions. In addition, the diversity of the initial population is
also important, and different distributions may have different sampling emphasis, leading to different degrees of diversity. For example, some studies concerning genetic algorithms have shown some effects of initialization ~\citep{Burke2004Diversity,Chou2000genetic}.

\begin{figure}[ht]
\centering
\includegraphics[width=6cm]{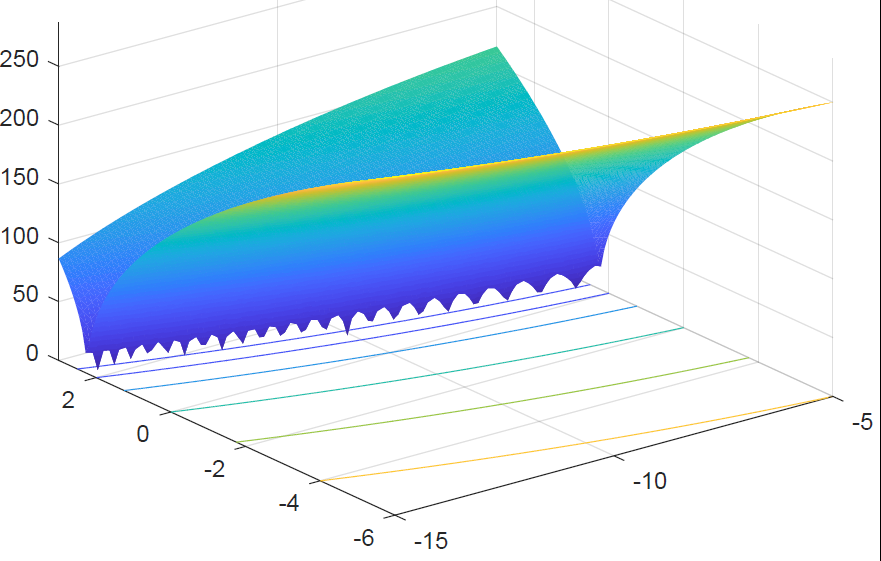}
\caption{The landscape of Bukin Function N.6. \label{Fig-1} }
\end{figure}

\begin{figure*}[htbp]
\centering
\subfigure[]{\includegraphics[width=6cm]{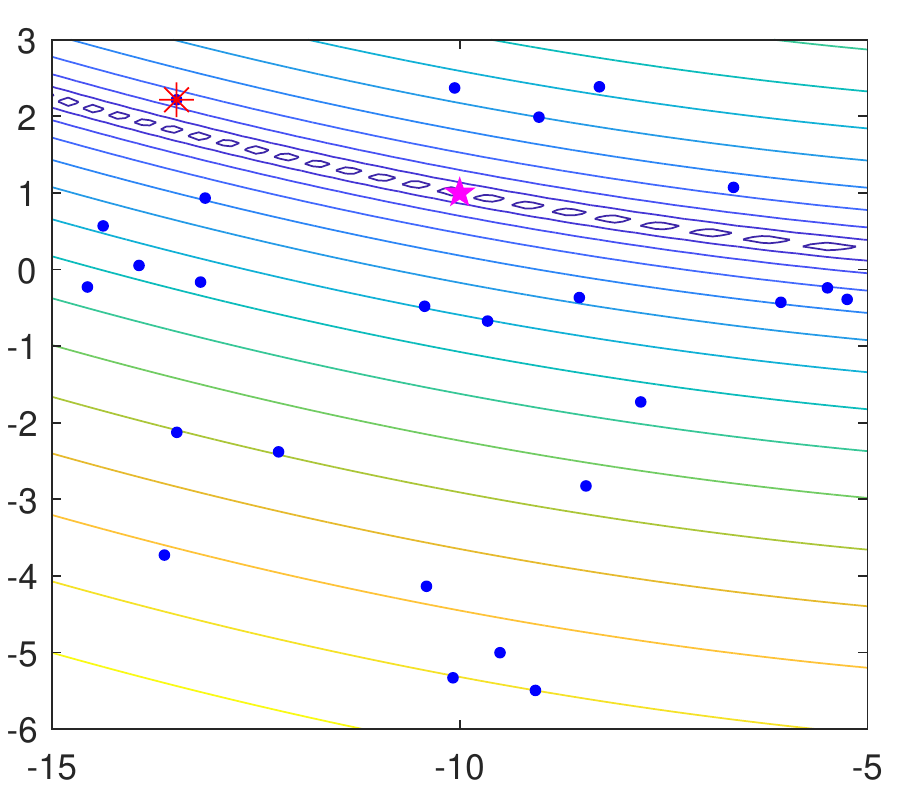}{\label{B_unfi_init}}}
\subfigure[]{\includegraphics[width=6cm]{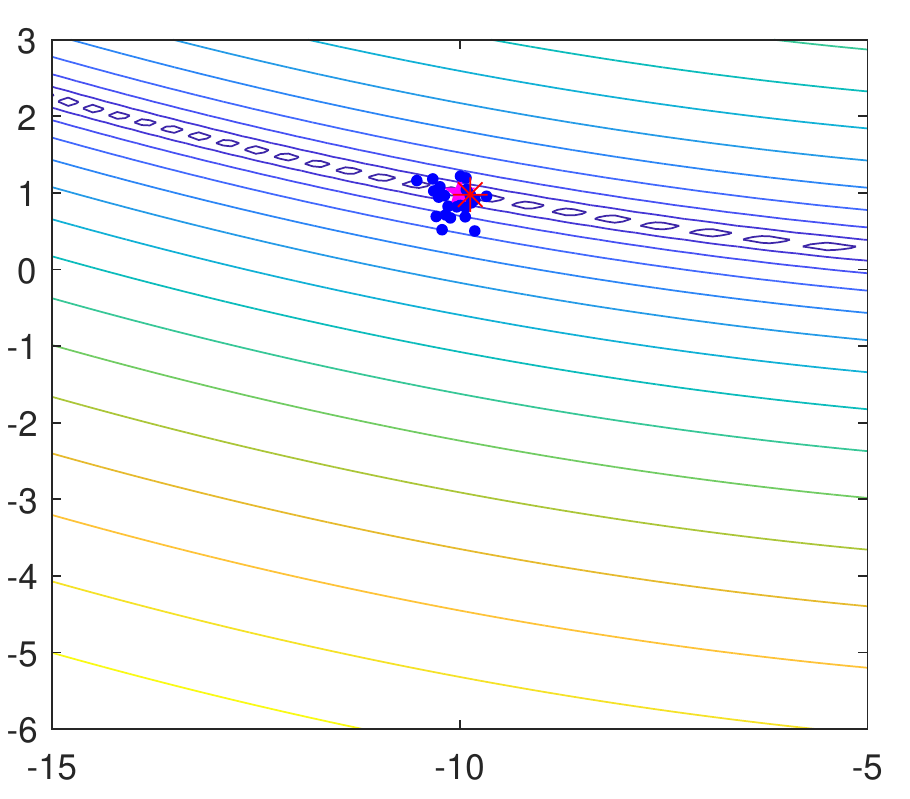}{\label{B_unfi5}}}
\caption{(a) The initial population drawn from a uniform distribution where the blue dots are the locations of the initial population, and the red $\convolution$ indicates the best solution found by the current population. The real optimal solution of this function is represented by $\bigstar$.  (b) Distribution of the same population after 5 iterations by PSO-w, the population converges near the real optimal solution.}
\label{Fig-2}
\end{figure*}

\begin{figure*}[htbp]
\centering
\subfigure[]{\includegraphics[width=6cm]{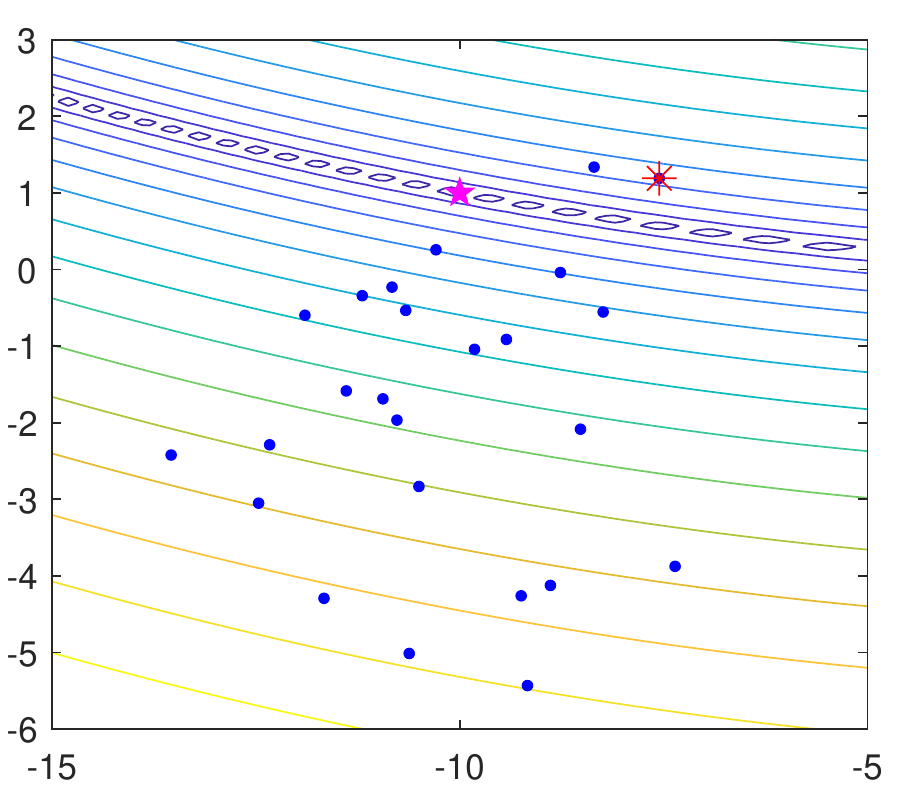}{\label{B_beta_init}}}
\subfigure[]{\includegraphics[width=6cm]{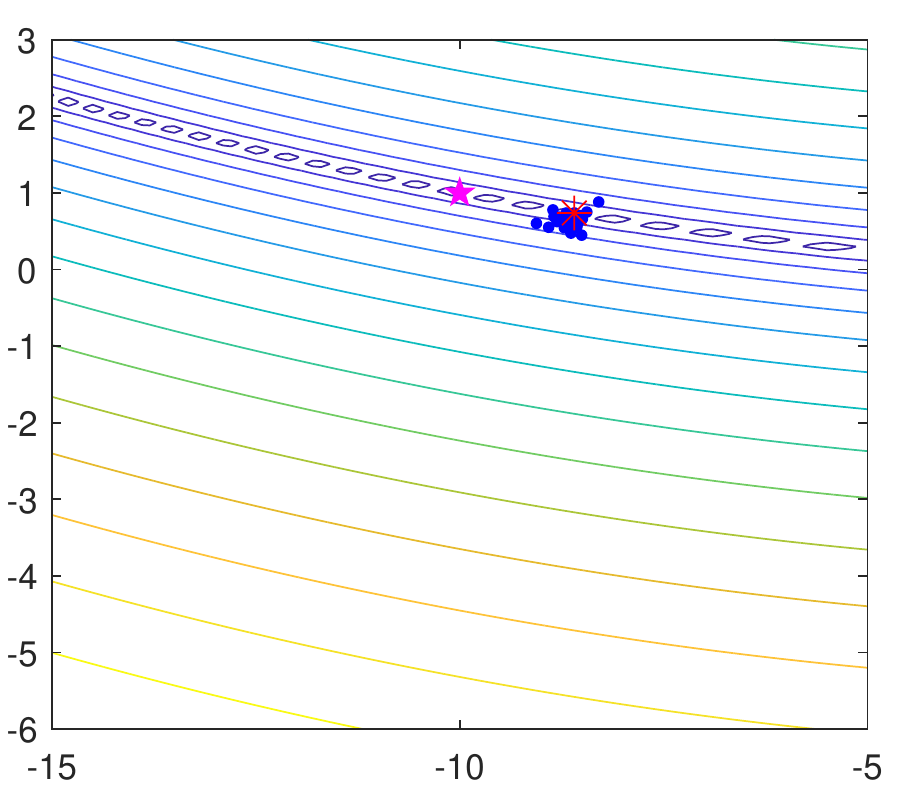}{\label{B_beta5}}}
\caption{(a) Initial population drawn from a beta distribution where the locations are marked with dots and the true optimality is marked with $\bigstar$. (b) The best solution $\convolution$ found by PSO-w after 5 iterations is far from the true optimal solution, indicating premature convergence.}
\label{Fig-3}
\end{figure*}

Many initialization methods such as the Latin hypercube sampling (LHS) in the literature are mainly based on the idea of uniform spreading in the search space. They are easy to implement and can work well sometimes. For example, the two-dimensional landscape of the Bukin function is shown in Fig.~\ref{Fig-1}. When the search space is in the area of $[-15,-5] \times [-6,3]$, the PSO-w algorithm with an initial population obeying a uniform distribution can find the optimal solution in a few iterations. The distribution of the particles is shown in Fig.~\ref{Fig-2}. For comparison, another run with an initial beta distribution has also been carried out as shown in Fig.~\ref{Fig-3}. Specifically, the $\bigstar$ indicates the real optimal solution at (-10,1), while the dots show the locations of the current population and ($\convolution$) indicates the best solution in current population. Fig.~\ref{B_unfi_init} shows the initial population with a uniform distribution in the search domain, while these population converged near the optimal solution after 5 iterations by the PSO-w algorithm, as shown in Fig.~\ref{B_unfi5} where the current best solution of the population is close to the real optimal solution. However, the initial population (as shown in Fig.~\ref{B_beta_init}) drawn from a beta distribution could fall into a local optimum after 5 iterations as shown in Fig.~\ref{B_beta5}. This clearly shows the effect and importance of initialization.

\begin{figure}[ht]
\centering
\includegraphics[width=6cm]{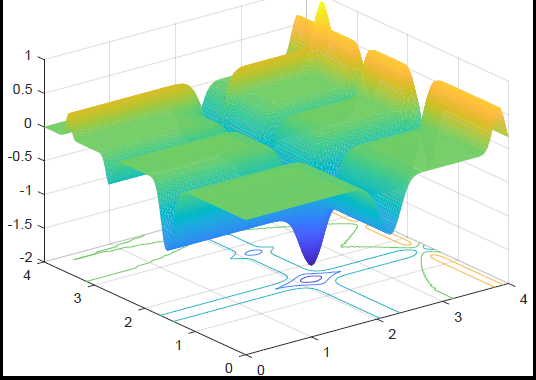}
\caption{The landscape of the Michalewicz Function. \label{Fig-4} }
\end{figure}

\begin{figure*}[htbp]
\centering
\subfigure[]{\includegraphics[width=6cm]{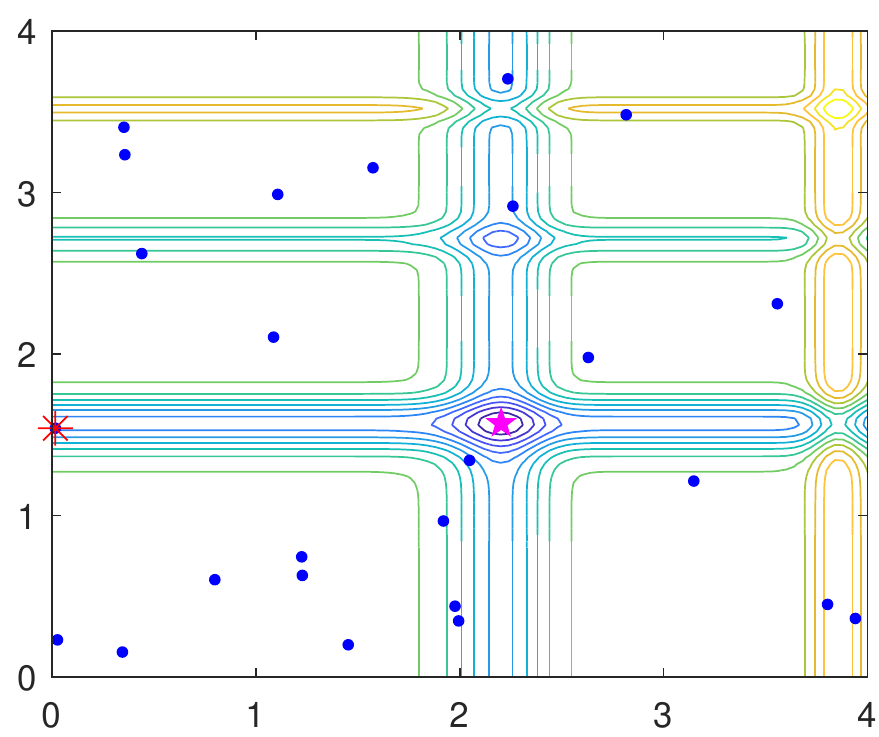}{\label{M_rand_init}}}
\subfigure[]{\includegraphics[width=6cm]{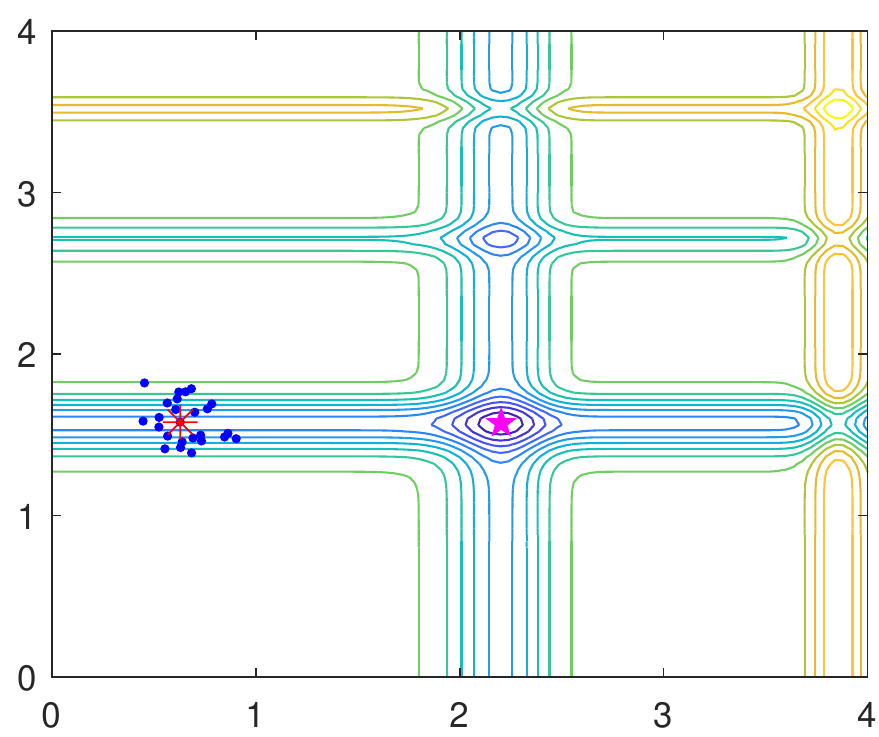}{\label{M_rand5}}}
\caption{(a) Initial population drawn from a uniform distribution. (b) The location of the best solution $\convolution$ found by PSO-w after 5 iterations is far from the true optimal solution $\bigstar$, leading to premature convergence.}
\label{Fig-5}
\end{figure*}

\begin{figure*}[htbp]
\label{fig_bug}
\centering
\subfigure[]{\includegraphics[width=6cm]{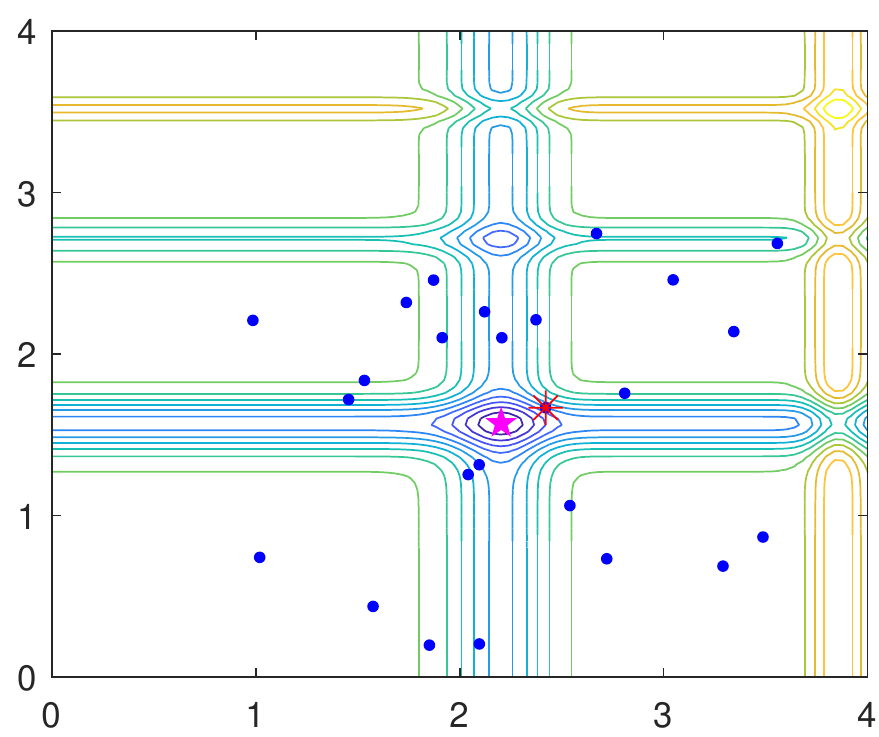}{\label{M_beta_init}}}
\subfigure[]{\includegraphics[width=6cm]{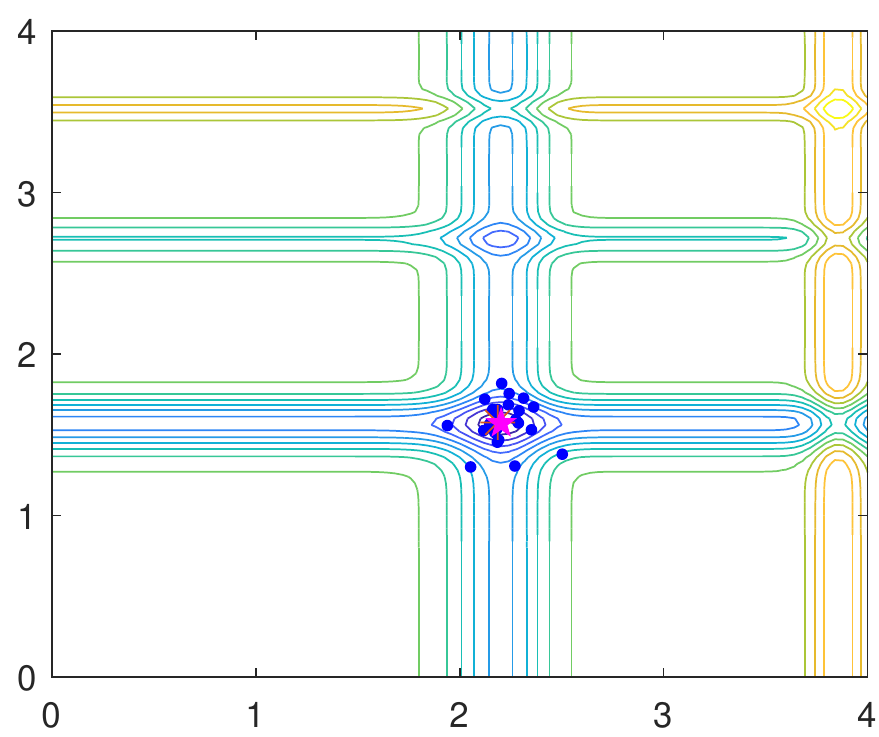}{\label{M_beta5}}}
\caption{(a) Initial population drawn from a beta distribution. (b) The best solution $\convolution$ found by PSO-w after 5 iterations is close to the true optimal solution $\bigstar$. }
\label{Fig-6}
\end{figure*}

For the above function, initialization by a uniform distribution seems to give better results. However, for another function, uniform distributions may give worse results, even though uniform distributions are widely used. As an illustrative example, the best solution of the Michalewicz function is $f_{\min}=-1.801$ in two-dimensional space at [2.20319,1.57049] (see Fig.~\ref{Fig-4}). If the initialization was done by a uniform distribution, it can lead to premature convergence as shown in Fig.~\ref{Fig-5}, while the initialization by a beta distribution can lead to the global optimal solution after 5 iterations as shown in Fig.~\ref{Fig-6}.
Clearly, this shows that uniform distributions are not the best initialization method for all functions. For the same algorithm (such as PSO-w), different initialization methods can lead to different accuracies for different problems. This suggests that different initialization methods should be used for different problems. We will investigate this issue further in a more systematically way.

In order to study the effect of initialization systematically, we will use a diverse range of different initialization methods such as Latin hypercube sampling and different probability distributions. We now briefly outline them in the rest of this section.

\subsection{Details of initialization methods}

Before we carry out detailed simulations, we now briefly outline the main initialization methods.

\subsubsection{Latin hypercube sampling}

\begin{figure}[ht]
\centering
\includegraphics[width=3.5cm]{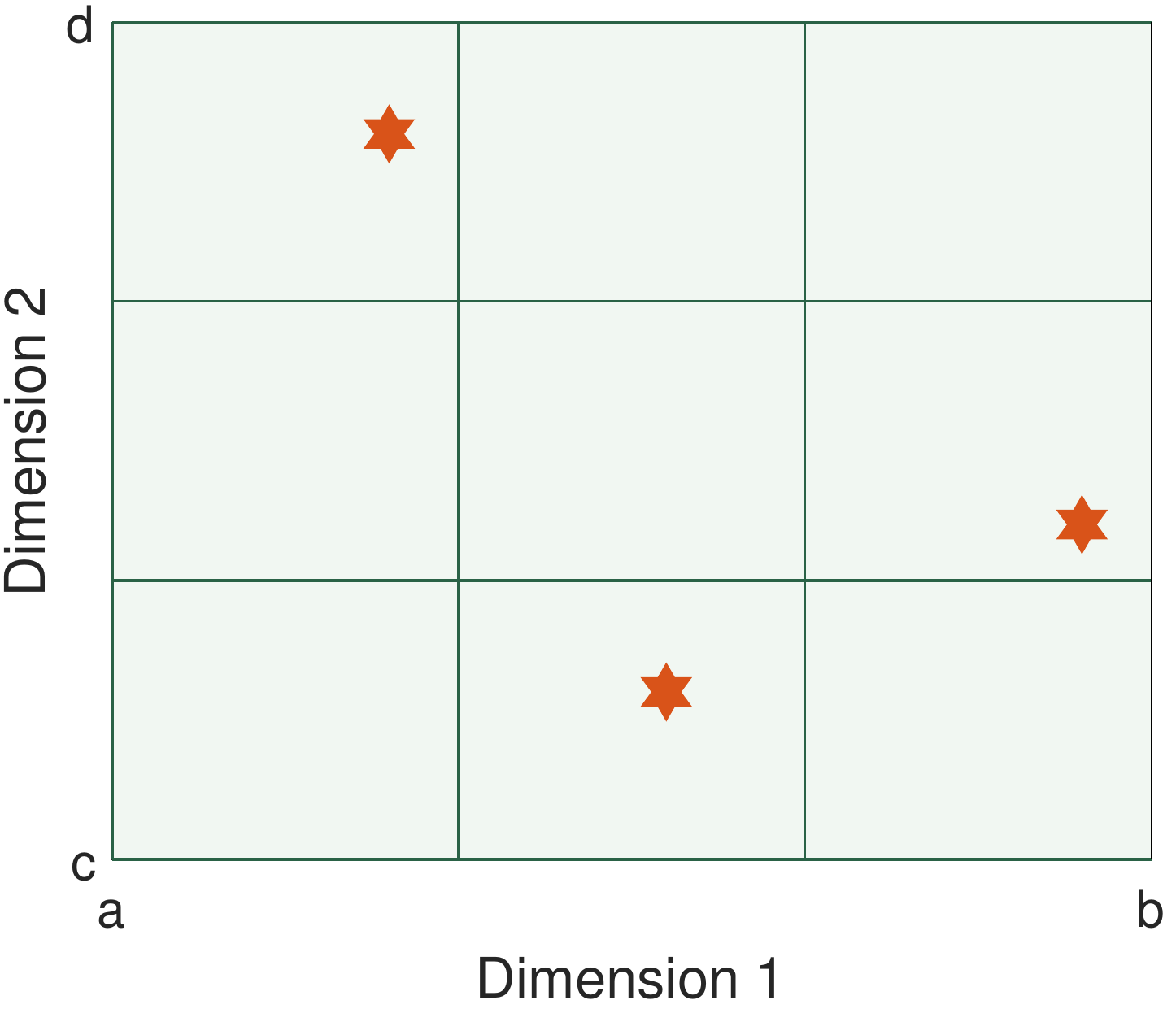}
\includegraphics[width=3.5cm]{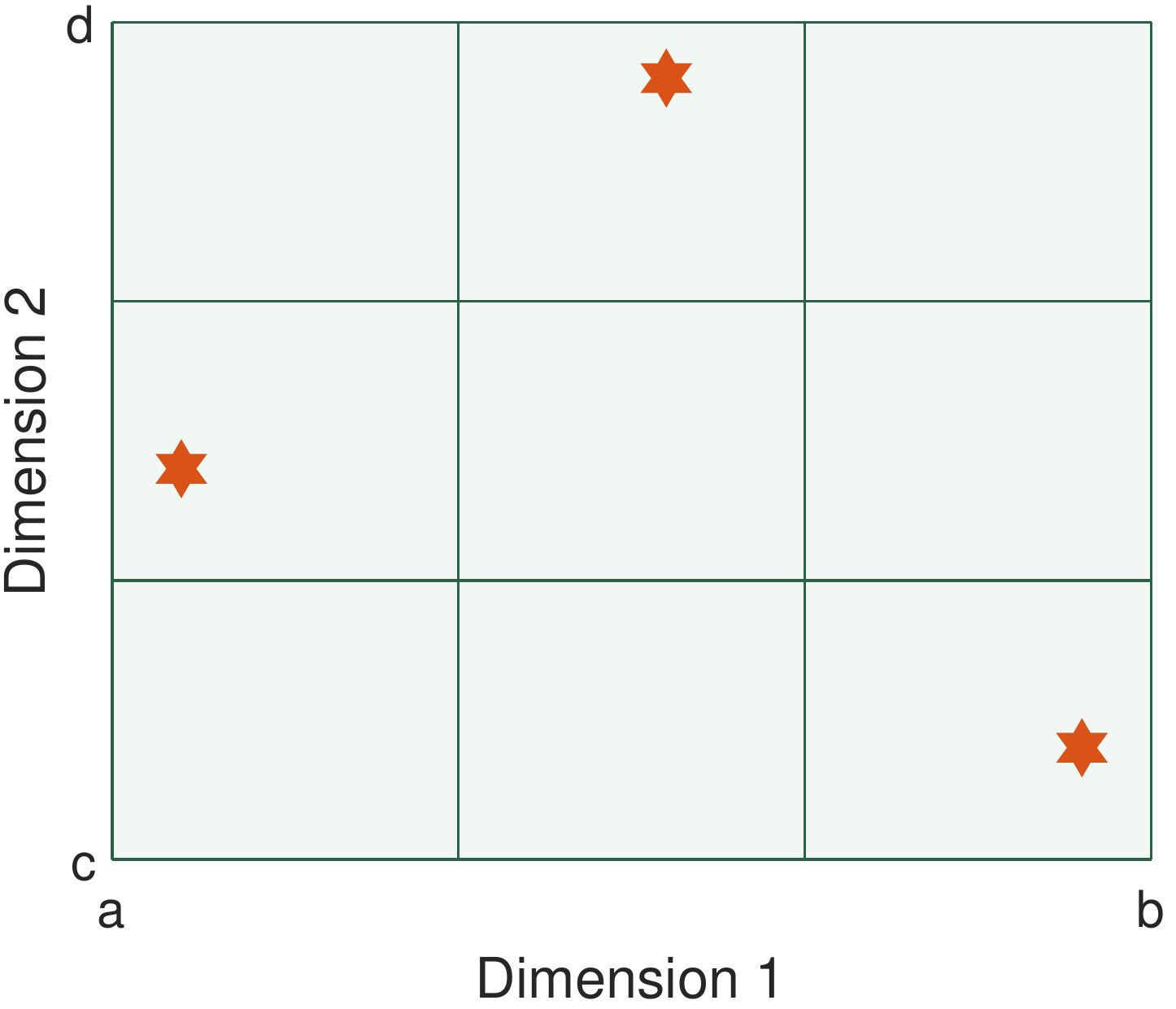}
\includegraphics[width=3.5cm]{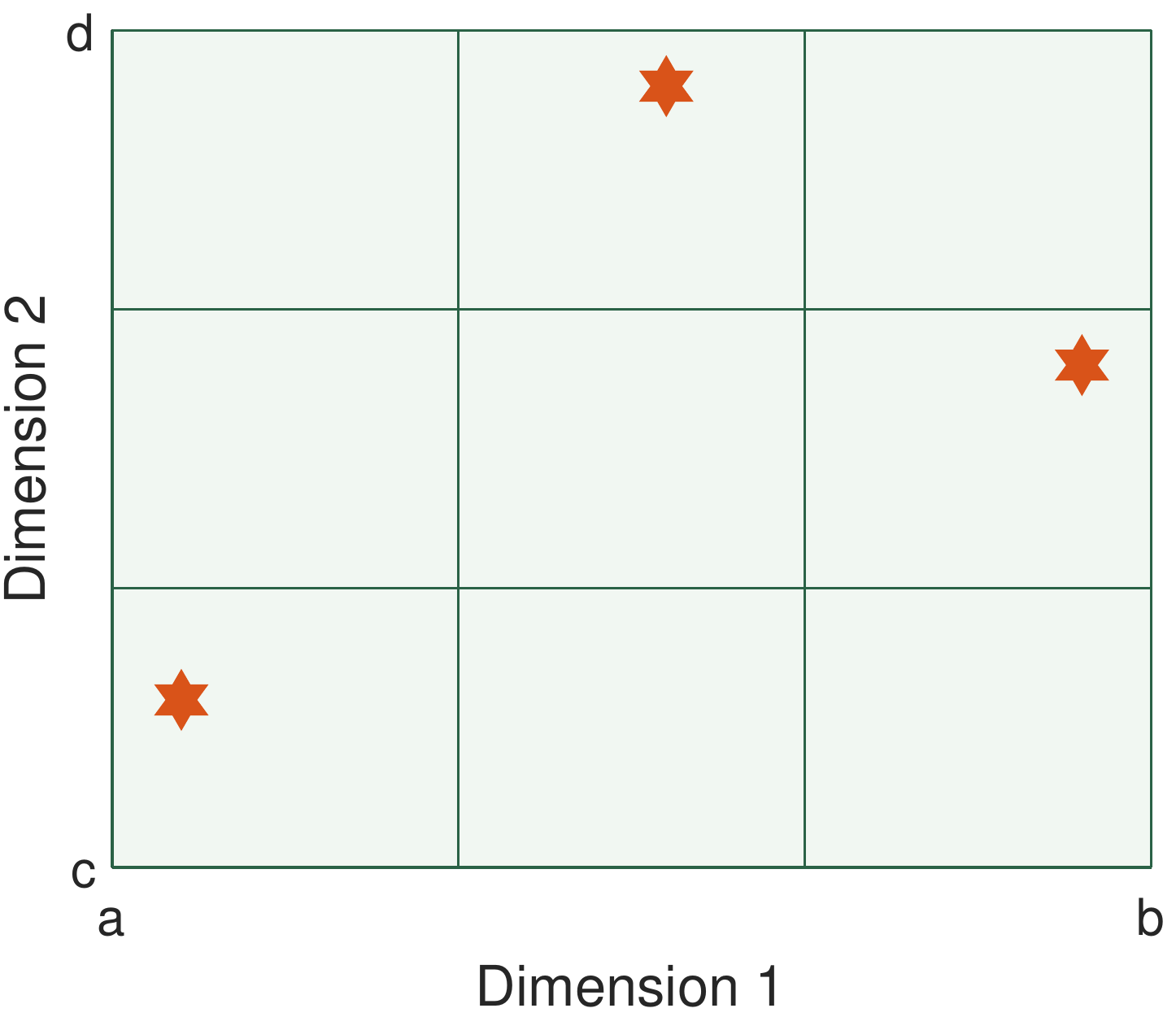}
\includegraphics[width=3.5cm]{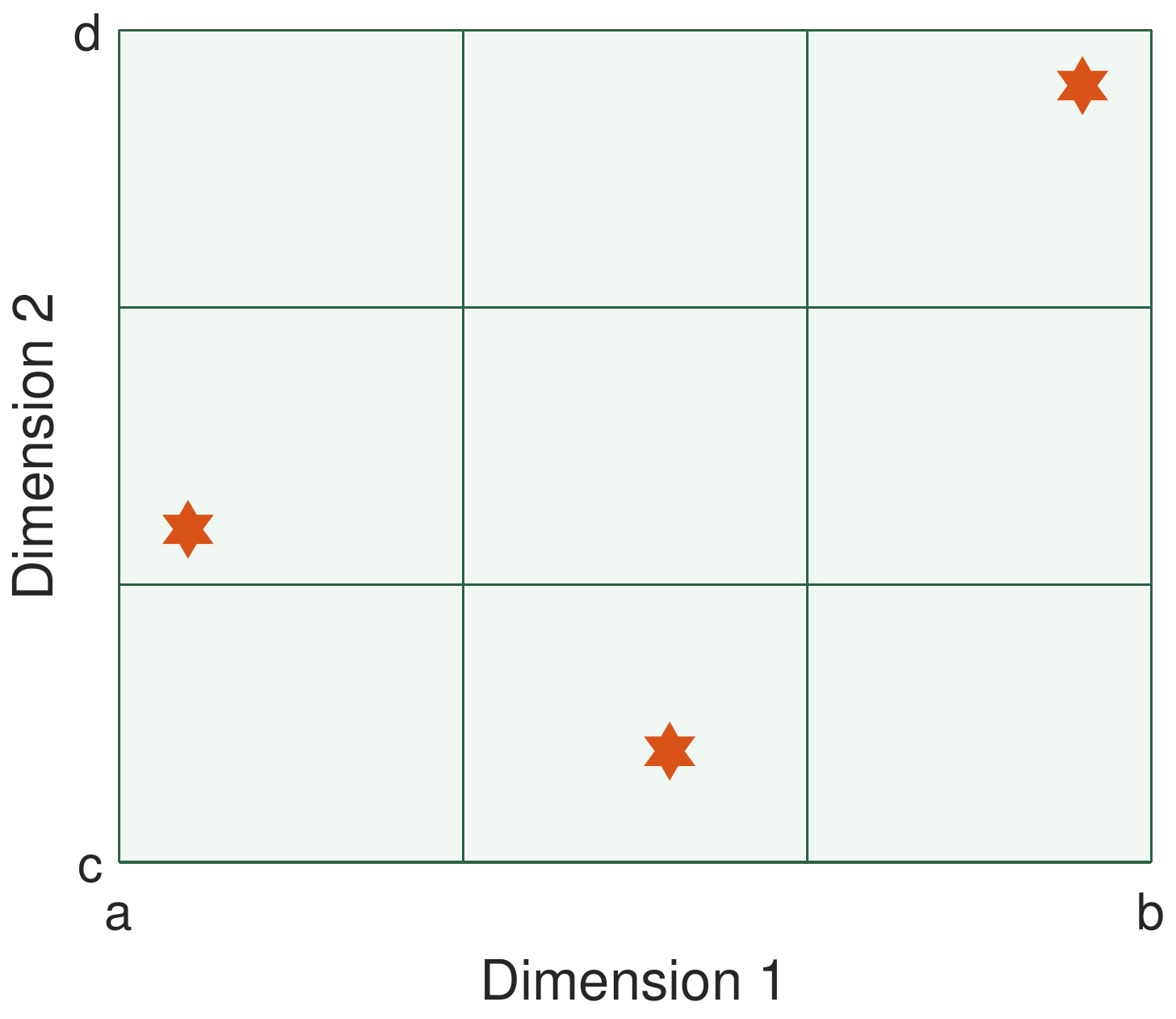}
\caption{ A 2D example of the LHS where three sampling points are distributed in four possible scenarios.}
\label{Fig-7}
\end{figure}

Latin hypercube sampling (LHS) is a spatial filling mechanism. It creates a grid in the search space by dividing each dimension into equal interval segments, and then generates some random points within some interval. It utilizes ancillary variables to ensure that each of the variables to be represented is in a fully stratified feature space~\citep{mckay1979comparison}. {{For example, if three sample points are needed in a two-dimensional (2D) parameter space, the three points may have four location scenarios (shown in Fig.~\ref{Fig-7}). Obviously, these three points can also be scattered in the diagonal subspace of the 2D search space.}}

{{In the LHS, a set of samples are distributed so that they can sparsely distribute in the search space so as to effectively avoid the problem of over aggregation of sampling points. Studies show that such sampling can provide a better spread than uniform distributions, but it does not show a distinct advantage for higher-dimensional problems.}} So we will investigate this issue further.

\subsubsection{Beta distribution}

A beta distribution is a continuous probability distribution over the interval (0,1).
Its probability density function (PDF) is given by
\begin{equation}\label{bate}
{\rm{p}}(x;a,b ) = \frac{{\Gamma (a  + b )}}{{\Gamma (a ) + \Gamma (b )}}{x^{a  - 1}}{(1 - x)^{b  - 1}},
\end{equation}
where $\Gamma (a)$ is the standard Gamma function. This distribution has two shape parameters ($a>0,b>0$)  that essentially control the shape of the distribution.
Its notation is usually written as $X \sim Be(a ,b)$. Its expected value is  $\mu=\frac{a }{{a {\rm{ + }}b }}$ and its variance is $\frac{{a b }}{{(a +b )(a + b + 1)}}$.

\subsubsection{Uniform distribution}

Uniform distributions are widely used in initialization, and a uniform distribution $U(a,b)$
on an interval $[a, b]$ is given by

\begin{equation}\label{Uni}
p(x) = \left\{ \begin{array}{l} \frac{1}{{b - a}}, \quad a < x < b, \\
0, \qquad \textrm{otherwise,}
\end{array} \right.
\end{equation}
where $a$ and $b$ are the limits of the interval. Its expectation or mean is $\frac{{a + b}}{2}$, and its variance is $\frac{(b - a)^2}{12}$.

\subsubsection{Normal distribution}

Gaussian normal distributions are among the most widely used distributions in various applications, though they are not usually used in initialization. The probability density function of this bell-shaped distribution can be written as
\begin{equation}\label{Nor}
p(x) = \frac{1}{{\sqrt {2\pi } \sigma }}\exp ( - \frac{{{{(x - \mu )}^2}}}{{2{\sigma ^2}}}),
\end{equation}
with the mean of $\mu$ and the standard deviation $\sigma$. This distribution is often written as
N($\mu ,{\sigma ^2}$) where its mean determines the central location of the probability curve and
its standard deviation $\sigma$ determines the spread on both sides of the mean~\citep{Yang2014,kizilersu2018weibull}.
Normal distributions can be approximated by other distributions
and can be linked closely with other distributions such as the log-normal distribution, Student-$t$ distribution and $F$-distribution.

\subsubsection{Logarithmic normal distribution}

Unlike the normal distribution, the Logarithmic normal distribution is an asymmetrical distribution.
Its probability density function is
\begin{equation}\label{Logn}
p(x)=\frac{1}{x \sigma \sqrt{2 \pi}} \exp\Big[-\frac{(\ln x-\mu)^2}{2 \sigma^2} \Big].
\end{equation}

A random variable $X$ obeying this distribution is often written as $\ln X \sim N(\mu ,{\sigma ^2})$. Its expectation and variance are $\exp [\mu  + {\sigma ^2}/2]$ and
$[{\exp({\sigma ^2})} - 1] {\exp[2\mu  + {\sigma ^2}]}$, respectively.

\subsubsection{Exponential distribution}

An exponential distribution is asymmetric with a long tail, and its probability density function can be written as

\begin{equation}\label{Exp}
p(x) = \left\{ \begin{array}{ll}
\lambda {e^{ - \lambda x}}, & x \ge 0, \\
0, & x < 0,
\end{array} \right.
\end{equation}
where $\lambda>0$ is a parameter. Its mean and standard deviation are $1/\lambda$ and $1/\lambda^2$, respectively.

\subsubsection{Rayleigh distribution}

The probability density function of the Rayleigh distribution can be written as
\begin{equation}\label{Ray}
p(x) = \frac{x}{{{\sigma ^2}}}{e^{ - \frac{{{x^2}}}{{2{\sigma ^2}}}}}, \quad x > 0,
\end{equation}
whose mean and variance are $\sqrt {\frac{\pi }{2}}$ and  $\frac{{4 - \pi }}{2}{\sigma ^2} \approx 0.429{\sigma ^2}$, respectively~\citep{weik2001rayleigh}.

\subsubsection{Weibull distribution}

The Weibull distribution has a probability density function~\citep{kizilersu2018weibull}
\begin{equation}\label{Wei}
f(x; \lambda, k) = \left\{ \begin{array}{lll}
\frac{k}{\lambda }{\left( {\frac{x}{\lambda }} \right)^{k - 1}}{e^{ - {{(x/\lambda )}^k}}}, &  x \ge 0, \\ 0, & x<0,  \end{array} \right.
\end{equation}
where $\lambda$ is a scale parameter, and $k$ is a shape parameter.
This distribution can be considered as a generalization of a few other distributions.
For example, $k=1$ corresponds to an exponential distribution, while $k=2$ leads to the
Rayleigh distribution. Both its mean and variance are $\lambda \Gamma (1 + \frac{1}{k})$ and ${\lambda ^2}[\Gamma (1 + \frac{2}{k}) - \Gamma {(1 + \frac{1}{k})^2}]$, respectively.

Based on the above different probability distributions, we will carry out various numerical experiments in the rest of this paper.

\section{Numerical Experiments}

\subsection{Experimental settings}

{{
In order to investigate the possible influence of different initialization methods on the five algorithms (PSO-w, DE-a, CS, ABC, GA), a series of experiments have been carried out first using a set of nine benchmark functions as shown in Table~\ref{function}. The experiments will focus first on the PSO-w, DE-a and CS, and then similar tests will be carried out for the ABC and GA.}} These benchmark functions are chosen based on their different properties such as their modal shapes and numbers of local optima. More specifically, $f_1$, $f_3$, $f_6$, and $f_8$ are continuous, unimodal functions, while $f_2$, $f_4$, $f_5$, $f_7$ and $f_9$ are multimodal functions. For example, the global minimum of $f_1$ lies in a narrow, parabolic valley, which can be difficult for many traditional algorithms. Functions $f_2$, $f_4$, $f_5$, and $f_9$ have many local minima that are widespread. The bowl-shaped function $f_3$ has $D$ local minima with only one global optimum, while the Easom function has several local minima, and its global minimum lies in a small area in a relatively large  search space. In addition, we will use 10 more recent benchmarks from CEC2014 and CEC2017 to be discussed in detail later.

\begin{table}[h]
\centering
\tiny
\caption{{{Basic Benchmark Functions.}} \label{function} }
\begin{tabular}{|c|c|c|c|c|}
\hline
Name & Function & Search Range & $x^{opt}$ & Opt \\
\hline
 Rosenbrock & ${f_1}(X) = \sum\nolimits_{i = 1}^{D - 1} {[100{{({x_{i + 1}} - x_i^2)}^2} + {{({x_i} - 1)}^2}]} $ & ${[ - 5,5]^D}$ & $(1,1, \cdots ,1)$    &0 \\
 Ackley  & ${f_2}(X) =  - 20\exp ( - 0.2\sqrt {\frac{1}{D}\sum\nolimits_{i = 1}^D {x_i^2} } )$ & & &\\
 &  \qquad \qquad $- \exp (\frac{1}{D}\sum\nolimits_{i = 1}^D {\cos } (2\pi {x_i})) + 20 + e$ & ${[ - 10,10]^D}$ & $(0,0, \cdots ,0)$ &0\\
 Sphere  &  ${f_3}(X) = \sum\nolimits_{i = 1}^D {x_i^2}$ & ${[ - 5,5]^D}$ & $(0,0, \cdots ,0)$&0 \\
 Rastrigin  & ${f_4}(X) = \sum\nolimits_{i = 1}^D {[x_i^2 - 10\cos (2\pi {x_i}) + 10]} $  & ${[ - 5.12,5.12]^D}$ & $(0,0, \cdots ,0)$ &0 \\
 Griewank & ${f_5}(X) = \frac{1}{{4000}}\sum\nolimits_{i = 1}^D {x_i^2}  - \prod _{i = 1}^D\cos (\frac{{{x_i}}}{{\sqrt i }}) + 1$ & ${[ - 600,600]^D}$ & $(0,0, \cdots ,0)$ &0 \\
 Zakharov & ${f_6}(X) = \sum\nolimits_{i = 1}^D {x_i^2}  + {(\frac{1}{2}\sum\nolimits_{i = 1}^D {i{x_i}} )^2} + {(\frac{1}{2}\sum\nolimits_{i = 1}^D {i{x_i}} )^4}$ & ${[ - 100,100]^D}$ & $(0,0, \cdots ,0)$ & 0 \\
 Alpine & ${f_7}(X) = \sum\nolimits_{i = 1}^D {\left| {{x_i}\sin ({x_i}) + 0.1{x_i}} \right|}$ & ${[ - 10,10]^D}$ & $(0,0, \cdots ,0)$ & 0 \\
 Easom & ${f_8}(X) = [ - \prod\nolimits_{i = 1}^D {\cos ({x_i})]} \exp ( - \sum\nolimits_{i = 1}^D {{{({x_i} - \pi )}^2}} )$ & ${[ - 100,100]^D}$ & $(\pi ,\pi , \cdots ,\pi )$ &-1 \\
 Schwefel & ${f_9}(X) = 418.98288727243369 * n - \sum\nolimits_{i = 1}^D {{x_i}} \sin (\sqrt {\left| {{x_i}} \right|} )$ & ${[ - 500,500]^D}$ & $420.96857*(1,1, \cdots ,1)$ &0 \\
\hline
\end{tabular}
\end{table}

For a fair comparison, we have set the same termination condition for all the algorithms with the maximum number of function evaluations (FEs) of 600000, each algorithm with certain initialization has 20 independent runs.
For all the test functions, the dimensionality is set to $D=30$. As there are so many sets of data generated, we have summarized the results as the `Best', `Mean', `Var' (variance) and `Dist'. Here, `Dist' corresponds to the mean distance from the obtained solution $x^{\rm find}$ to the true global optimal solution $x^{opt}$. That is
\begin{equation}\label{dist}
{\rm Dist} = \frac{{\sum\nolimits_{i = 1}^{TN} {\sum\nolimits_{j = 1}^D {\left| {x_{i,j}^{\rm find} - {x_j}^{opt}} \right|} } }}{{TN}},
\end{equation}
where $TN=20$ denotes the total number of runs in each set of experiments. This distance metric not only measures the distance of the results, but also measures the stability of the obtained solutions.

For the algorithm-dependent parameters, after some preliminary parametric studies, we have set
$CR$ and $F$ to $[0.4,0.5,0.6,0.7,0.8]$ and $[0.5,0.6,0.7,0.8,0.9]$, respectively, for DE-a.
In the PSO-w, learning factors $c_1$ and $c_2$ are set to 1.5, and
the inertia weight $w=0.8$. For the CS, we have used $p_a=0.25$ and $\lambda=1.5$. In addition, the population size ($NP$) will be varied so as to see if it has any effect on the results.

\subsection{Influence of population size and number of iterations}

Before we can compare different initialization methods in detail, we have to figure out if there is any significant effect due to the number of the population ($NP$) used and the maximum number of iterations $T$. Many studies in the existing literature used different population sizes and numbers of iterations~\citep{akay2012modified}. Though the total number of function evaluations for all
functions and algorithms is set to 600 000, the maximum iteration $T$ will vary with $NP$. Obviously,
a larger $NP$ will lead to a smaller $T$.

In order to make a fair comparison, all the algorithms are initialized by the same random initialization. Four functions with $D=30$ are selected randomly to reduce the computational efforts. We have carried out numerical experiments and the results are summarized in Tables~\ref{Table-2} to \ref{Table-4}.

\begin{table}[h]
\centering
\scriptsize
\caption{Influence of the population size and maximum iteration number on the
DE-a algorithm. \label{Table-2} }
\begin{tabular}{|l|l|c|c|c|c|c|c|c|c|}
\hline
 Fun & value& \tabincell{c}{NP=100\\T=6000} & \tabincell{c}{NP=200\\T=3000} & \tabincell{c}{NP=300\\T=2000} & \tabincell{c}{NP=600\\T=1000} & \tabincell{c}{NP=1000\\T=600} & \tabincell{c}{NP=2000\\T=300} & \tabincell{c}{NP=3000\\T=200} \\
\hline
 \multirow{4}{*}{Rosenbrock} & Best & 0 & 5.09e-19 & 1.39e-09 & 2.53 & 12.225 &19.929 &22.198\\
                             &Mean  & 0.0987 & 0.1993 & 0.1993 & 7.2212 & 13.632 & 21.224 & 23.878\\
                             &Var   & 1.5057 & 0.7947 &0.7947 & 215.2 & 1.2934 & 1.9251 & 1.2532 \\
                             &Dist  & 0.2    & 0.0999 & 0.1003 &5.6464 &14.933 &22.638 &25.277 \\
\hline
\multirow{4}{*}{Sphere}      & Best & 5.67e-197 & 9.71e-105 & 8.64e-70  & 7.19e-36 & 1.02e-22 & 9.05e-11  &1.63e-07\\
                             &Mean  & 1.78e-187 & 1.57e-96  & 4.45e-65  & 7.74e-33 & 1.92e-19 & 1.72e-09  & 2.74e-06\\
                             &Var   & 0           & 3.63e-191 & 1.99e-128 & 2.88e-64 & 5.98e-37 & 1.18e-17 & 2.39e-11 \\
                             &Dist  & 2.0039e-48  & 2.01e-48  & 1.48e-32  & 2.34e-16 & 9.17e-10 & 1.44e-4  & 5.88e-3 \\
\hline
\multirow{4}{*}{Rastrugin}   & Best & 6.9647 & 18.271 & 91.987  & 113.07 & 112.94 & 130.32  &140.17\\
                             &Mean  & 43.547 & 96.429 & 113.77  & 122.2 & 131.08 & 142.57  & 151.7\\
                             &Var   & 1108.7  & 558.39 & 95.447 & 48.535 & 77.995 & 41.862 & 64.614 \\
                             &Dist  & 14.371  & 19.502 & 21.89  & 23.979 & 24.69 & 26.31   & 27.254 \\
\hline
\multirow{4}{*}{Griewank}      & Best & 0         & 0          & 0          & 0          & 0          & 7.21e-12  &7.07e-09\\
                             &Mean  & 1.11e-03   & 1.11e-03  & 2.22e-03  & 0       & 3.53e-03  & 1.11e-03 & 1.11e-03\\
                             &Var   & 7.34e-06  & 7.34e-06 & 1.21e-05 & 0          & 8.69e-05 & 7.34e-06& 7.35e-06 \\
                             &Dist  & 1.1368      &  1.1368    & 2.2735     & 8.60e-07 & 0.9227     & 1.1371  & 1.1572 \\
\hline
\end{tabular}
\end{table}

\begin{table}[h]
\centering
\scriptsize
\caption{Influence of the population size and maximum iteration number on the PSO-w algorithm. \label{Table-3} }
\begin{tabular}{|l|l|c|c|c|c|c|c|c|c|}
\hline
 Fun & value& \tabincell{c}{NP=100\\T=6000} & \tabincell{c}{NP=200\\T=3000} & \tabincell{c}{NP=300\\T=2000} & \tabincell{c}{NP=600\\T=1000} & \tabincell{c}{NP=1000\\T=600} & \tabincell{c}{NP=2000\\T=300} & \tabincell{c}{NP=3000\\T=200} \\
\hline
 \multirow{4}{*}{Rosenbrock} &Best & 27.141 & 17.382 & 7.7837 & 17.936 & 13.534 & 16.019 & 14.754 \\
                             &Mean  & 36.055 & 28.803 & 24.005 & 21.842 & 18.815 & 19.304 & 18.085\\
                             &Var   & 264 & 161.56 & 18.62 & 5.035 & 7.1071 & 6.5569 & 2.1004 \\
                             &Dist  & 27.465    & 26.791 & 25.076 & 23.145 & 20.193  & 20.678 & 19.607 \\
\hline
\multirow{4}{*}{Sphere}      &Best & 2.46e-04 & 1.69e-08 & 9.77e-16  & 1.33e-36 & 4.56e-28 & 3.91e-18  &4.12e-14\\
                             &Mean  & 2.32e-03 & 2.35e-07  & 1.14e-11  & 5.68e-34 & 2.84e-27 & 1.30e-17  & 1.44e-13\\
                             &Var   & 3.51e-06 & 1.30e-13 & 1.35e-21 & 1.14e-66 & 4.69e-54 & 6.21e-35 & 6.36e-27 \\
                             &Dist  & 1.91e-01 & 1.78e-03 & 7.72e-06  & 6.91e-17 & 2.18e-13 & 1.54e-08 & 1.59e-06 \\
\hline
\multirow{4}{*}{Rastrugin}   &Best & 28.59 & 17.913 & 22.884  & 19.899 & 12.935 & 12.935  & 8.9567\\
                             &Mean  & 44.819 & 35.542 & 33.732  & 32.187 & 25.073 & 22.287  & 18.26\\
                             &Var   & 91.411 & 98.105 & 41.561 & 68.803 & 85.093 & 48.604 & 30.551 \\
                             &Dist  & 27.43  & 23.591 & 23.187  & 21.74 & 18.805 & 17.91   & 15.081 \\
\hline
\multirow{4}{*}{Griewank}    &Best & 5.36e-05 & 5.12e-09  &2.92e-14 & 0          & 0          & 2.22e-16   & 0  \\
                             &Mean  & 1.14e-03   & 3.25e-03  & 5.77e-03& 2.34e-03  & 3.69e-04  & 3.70e-04  & 3.69e-04   \\
                             &Var   & 5.25e-06   & 1.28e-04 &2.76e-04& 1.09e-04   & 2.74e-06  & 2.73e-06 & 2.73e-06  \\
                             &Dist  & 1.177    & 1.2848    &  1.4333    & 5.02e-01   & 3.79e-01  & 3.79e-01  & 3.79e-01 \\
\hline
\end{tabular}
\end{table}

\begin{table}[h]
\centering
\scriptsize
\caption{Influence of the population size and maximum iteration number on the CS algorithm. \label{Table-4} }
\begin{tabular}{|l|l|c|c|c|c|c|c|c|c|}
\hline
 Fun & value& \tabincell{c}{NP=30\\T=20000} & \tabincell{c}{NP=60\\T=10000}& \tabincell{c}{NP=100\\T=6000} & \tabincell{c}{NP=200\\T=3000} & \tabincell{c}{NP=300\\T=2000} & \tabincell{c}{NP=600\\T=1000} & \tabincell{c}{NP=1000\\T=600}  \\
\hline
 \multirow{4}{*}{Rosenbrock} &Best & 0          & 3.18e-13  & 2.76e-01 & 7.01 & 12.55    & 31.904     & 92.556  \\
                             &Mean  &3.22e-30  & 4.05e-09  & 2.6204     & 11.9608 & 16.78   & 33.51      & 105.21 \\
                             &Var   &1.04e-58  & 2.67e-16  & 1.5332     & 7.8448 & 7.1492   & 8.1804e-01 & 83.05  \\
                             &Dist  &5.55e-17  & 1.90e-05   & 4.3424     & 12.113 & 15.364   & 27.867     & 27.592  \\
\hline
\multirow{4}{*}{Sphere}      &Best & 1.91e-139 & 4.48e-62 & 1.27e-32  & 5.91e-14 & 1.23e-08 & 1.42e-03  & 7.60e-02\\
                             &Mean  & 1.41e-136 & 2.54e-61  & 3.63e-32  & 9.10e-14 & 2.22e-08 & 2.03e-03  & 1.36e-01\\
                             &Var   & 1.47e-271 & 7.27e-122 & 1.86e-64 & 7.30e-28 & 5.04e-17& 1.86e-07 & 5.14e-04 \\
                             &Dist  & 3.16e-68 & 2.03e-30 & 8.20e-16  & 1.30e-06 & 6.50e-04 & 1.97e-01& 1.62 \\
\hline
\multirow{4}{*}{Rastrugin}   &Best & 0          & 12.791 & 24.333  & 47.727 & 55.124 & 77.96  & 89.599\\
                             &Mean  & 9.45e-01 & 16.8   & 34.625  & 57.695 & 68.146 & 89.385  & 102.36\\
                             &Var   & 8.83e-01 & 7.4984  & 16.161 & 36.615 & 33.234 & 38.407 & 54.137 \\
                             &Dist  & 9.45e-01  & 14.626 & 22.947  & 30.64 & 33.247 & 36.986   & 40.746 \\
\hline
\multirow{4}{*}{Griewank}    &Best & 0          & 0          &0             & 2.71e-11  & 1.28e-06  & 2.48e-02   & 2.57e-02  \\
                             &Mean  & 0          & 0          & 0            & 1.67e-10  & 2.49e-06 & 3.19e-02  & 3.66e-02   \\
                             &Var   & 0          & 0          &0             & 1.48e-20   & 2.72e-12  & 3.27e-07 & 2.05e-05  \\
                             &Dist  & 5.28e-07 & 5.46e-07 &  5.19e-07   & 2.81e-04   & 3.60e-02  & 1.3211  & 4.5692 \\
\hline
\end{tabular}
\end{table}

{{Table~\ref{Table-2} shows the experimental results of the DE-a algorithm with different $NP$ and $T$.
When $NP = 100$ and $T=6000$, DE-a shows better performance in most cases. That means the accuracy of the DE-a algorithm depends more heavily on the number of iterations, and it manages to find the optimal solution with a small population size.}}

Table~\ref{Table-3} summarizes the results for the PSO-w algorithm. We can see that the PSO-w algorithm performs well on the Rosenbrock, Rastrugin and Griewank functions when the size of population is 3000 and the number of iterations is 200. Only for the Sphere function, the PSO-w has the highest search accuracy when $NP=600$ and $T=1000$. The results show that the accuracy of the PSO-w may depend more on its population size.

Table~\ref{Table-4} shows that the CS algorithm has better performance under a small population and repeatedly iterations. Compared with DE, CS can find the optimal solution with a smaller size of population. This may be related to the design mechanism of the CS algorithm, which increases the diversity in the iteration process of the algorithm. This is one of the advantages of the CS algorithm.

Based on the above experiments, it is recommended that the population size and the number of maximum iterations be set as shown in Table~\ref{table-5}. Thus, these parameter settings will be used in all the subsequent experiments.
\begin{table}[h]
\centering
\scriptsize
\caption{Parameter settings for DE-a, PSO-w and CS. \label{table-5}}
\begin{tabular}{|l|l|l|l|}
\hline
algorithm & $NP$  &   $T$ \\
\hline
DE-a &      100   &    6000 \\
\hline
PSO-w &    3000  & 200 \\
\hline
CS &  30   &  10000 \\
\hline
\end{tabular}
\end{table}

\subsection{Numerical results}

In order to compare the possible effects of different initialization strategies for
the first three algorithms (PSO-w, DE-a and CS), 22 different initialization methods have been tested, including 9 different distributions with different distribution parameters. As before, we have used different benchmarks with $D=30$ and have run each algorithm independently for 20 times. Tables~\ref{Table-6}, \ref{Table-7} and \ref{Table-8} show the comparison results of the `Best', `Mean', `Var' and `Dist' obtained by the three algorithms.

\begin{table}[h]
\begin{adjustwidth}{-1.7cm}{}
\centering
\tiny
\caption{Comparison of DE-a for functions ${f_1}$-${f_9}$ with different initialization methods. \label{Table-6} }
\begin{tabular}{|p{0.2cm}|p{0.45cm}|p{1.06cm}<{\centering}|p{1.08cm}<{\centering}|p{1.06cm}<{\centering}|p{1.06cm}<{\centering}|p{1.06cm}<{\centering}|p{1.06cm}<{\centering}|p{1.2cm}<{\centering}|p{1.2cm}<{\centering}|p{1.3cm}<{\centering}|p{1.09cm}<{\centering}|p{1.12cm}<{\centering}|}
\hline
 Fun & Value  & $ Be(3,2)$  & $ Be(2.5,2.5)$ & $Be(2,3)$ & $U(0,1)$ & $N(0,1)$ & $N(0.5,1)$ & $N(0.5,0.5)$ & $logn(0,1)$ &$logn(.69,.25)$ & $logn(0,0.5)$ & $logn(0,2/3)$ \\
\hline
 \multirow{4}{*}{ ${f_1}$ } &Best &0 &	0 &	0 &	0 &	0 &	0 &	0 &	0 &	0 & 0 &	0 \\
                             &Mean  &0.9967 &  0.7973&	0.7973 & 0.5980 &	0.7973 & 0.7973 &	0.5980&	 0.5980 & 0.3987 &0.1993&	0.3987 \\
                             &Var   &3.1368	&2.6767	&2.6767	&2.133	&2.6767	&2.6767	&2.133	&2.133	 &1.5057	&0.79466	&1.5057 \\
                             &Dist  &0.4999 &	0.3999 &	0.3999&	0.2999&	0.3999&	0.3999&	0.2999&	 0.2999&	0.2& 0.09999&	0.2 \\
\hline
\multirow{4}{*}{${f_2}$}  &Best & 2.66e-15 &2.66e-15&	2.66e-15&	2.66e-15&	2.66e-15&	2.66e-15&	 2.66e-15&	2.66e-15&	2.66e-15&
2.66E-15&	2.66E-15\\
                             &Mean & 6.04e-15&	5.68e-15&	5.86e-15&	5.86e-15&	5.68e-15&	 5.86e-15&	5.51e-15&	6.04e-15&	5.33e-15&	5.86e-15&	5.51e-15\\
                             &Var  & 6.31e-31&	1.69e-30&	1.20e-30&	1.20e-30&	1.69e-30&	 1.20e-30&	2.13e-30&	6.31e-31&	2.49e-30&	1.20e-30&	2.13e-30 \\
                             &Dist & 5.63e-14&	5.28e-14&	5.50e-14&	5.47e-14&	5.37e-14&	 5.71e-14&	5.28e-14&	5.61e-14&	5.11e-14&	5.44e-14&	5.23e-14\\
\hline
\multirow{4}{*}{${f_3}$}   & Best & 2.35e-194&	3.08e-197&	4.91e-195&	8.12e-195&	3.43e-195&	 1.69e-195&	2.40e-195&	7.07e-194&	2.62e-191& 1.62e-193&	3.66e-193\\
	
                             &Mean  & 2.73e-189&2.67e-185&	6.97e-186&	3.27e-187&	1.87e-187&	 1.03e-187&	4.39e-188&	1.42e-187&	2.16e-185&	8.86e-187&	1.19e-186\\
                             &Var   & 0&	0&	0&	0&	0&	0&	0&	0&	0&	0&	0 \\
                             &Dist  & 1.33e-94&	5.54e-93&	4.13e-93&	1.22e-93&	1.11e-93&	 4.91e-94&	4.77e-94&	7.61e-94&	5.34e-93&	1.93e-93&	2.05e-93 \\
\hline
\multirow{4}{*}{${f_4}$}    & Best & 5.9698&	3.9798&	4.9748&	6.9647&	6.9647&	4.9748&	6.9647&	2.9849&	 5.9698&	7.9597& 4.9748  \\
                             &Mean  &42.146&	36.604&	30.935&	44.655&	48.159&	19.308&	45.89&	34.506&	 39.749&	34.128&	42.129 \\
                             &Var   & 877.55&	1124.6&	998.76&	1243.8&	1229.1&	349.48&	896.15&	943.9&	 819.12&	661.84&	1006.8  \\
                             &Dist  & 13.865&	11.948&	13.167&	13.421&	14.901&	10.458&	16.187&	12.368&	 14.037&	14.138&	13.541\\
\hline
\multirow{4}{*}{${f_5}$}    &Best & 0&	0&	0&	0&	0&	0&	0&	0&	0&	0&	0 \\
                             &Mean  &4.31e-03&	4.56e-03&	6.04e-03&	6.16e-03&	4.19e-03&	 3.82e-03&	5.67e-03&	5.67e-03&	3.82e-03&	3.57e-03&	4.68e-03 \\
                             &Var   & 3.38e-05&	1.94e-05&	5.58e-05&	8.77e-05&	4.15e-05&	 4.05e-05&	4.99e-05&	3.78e-05&	3.73e-05&	6.40e-05&	2.81e-05 \\
                             &Dist  & 3.5522&	4.3823&	4.6998&	4.5174&	3.0828&	2.9011&	4.3142&	4.5029&	 3.1007&	2.2772&	4.0719\\
\hline
\multirow{4}{*}{${f_6}$ } & best &7.14e-04&1.69e-04&7.76e-05 &8.13e-03&	2.52e-03&	1.06e-03&	 1.81e-04&	7.29e-04&	2.88e-03&	7.33e-03&	9.60e-04 \\
                             &Mean  &0.7707& 	0.2982& 0.1959 & 0.8781& 0.6424& 0.8083& 0.6176& 0.7647& 0.6395& 1.1053& 	0.3683 \\
                             &Var   &2.9675& 1.2030& 0.2235 & 1.8932& 1.3720& 2.2181& 3.4087& 1.8174 &1.1824 &7.0834& 0.2632 \\
                             &Dist  &2.2541 &1.0356& 1.1638 &2.9246 &2.5203& 2.5707& 2.0005 &2.7499& 2.5098 &3.1681 &	2.1418\\
\hline
 \multirow{4}{*}{${f_7}$ } &Best & 6.41e-179&	2.30e-171&	3.37e-202&	9.60e-158&	1.26e-175&	 3.15e-190 &	3.18e-157&	6.17e-175&	6.19e-191&	1.62e-186&	2.66e-193\\
                             &Mean  &2.69e-16&	4.97e-16&	3.28e-16&	1.81e-16&	5.25e-16&	 1.25e-16 &	5.75e-16&	2.75e-16&	2.25e-16&	3.22e-16&	2.78e-16 \\
                             &Var   &1.92e-31&	3.20e-31&	1.60e-31&	4.55e-32&	5.29e-31&	 8.40e-32 &	1.76e-31&	9.26e-32&	6.79e-32&	1.68e-31&	1.21e-31 \\
                             &Dist  &3.3825&	5.0903&	4.4269&	3.0651&	5.8807&	2.455 &	5.6116&	3.9507&	 3.7535&	3.8837&	4.34 \\
 \hline
 \multirow{4}{*}{${f_8}$ } &Best &0 &	0 &	0 &	0 &	0 &	0 &	0 &	0  &	0 &	0 &	0 \\
                             &Mean  &0 &	0 &	0 &	0 &	0 &	0 &	0 &	0 &	0 &	0 &	0 \\
                             &Var  &0 &	0 &	0 &	0 &	0 &	0 &	0 &	0 &	0 &	0 &	0 \\
                             &Dist  &1373.1&	1317.2&	1350.1&	1333.4&	1343.6&	1354.2&	1353.9&	1256.2 &	1343.2&	1315.9&	1333.6 \\
  \hline
 \multirow{4}{*}{${f_9}$ } &Best & -3.64e-12 &	-3.64e-12&	118.44&	-3.64e-12&	-1208.1&-1330.4&	 -3.64e-12&	-83319&	-9507.6&	-2585.2&	-3167.7 \\
                             &Mean  & 159.89 &	324.72&	379& 342.48&	1716.7&	2918.7&	225.03&	-22174&	 -3801&	1872.5&	732.47\\
                             &Var  & 18125&	64990&	42231&	46360&	2.92e+06&	4.24e+06&	38244&	 2.89e+08&	7.38e+06&	3.89e+06&	3.45e+06 \\
                             &Dist  & 976.72 &	1928.1&	2315.2&	2036.6&	34256&	24285&	1374.6&	60249&	 34954&	8528.4&	22632 \\
\hline
\hline
Fun &   & $ E(0.5)$  & $E(0.1) $ & $E(0.8)$ & $Rayl(0.4)$ & $Rayl(0.8)$ & $Rayl(0.1)$ & $Weib(1,1.5)$ & $Weib(1.5,1)$ & $Weib(1,1)$ & $random$ & $LHS$ \\
\hline
 \multirow{4}{*}{${f_1}$ } & Best &0 &	0 &	0 &	0 &	0 &	0 &	0 &	0 &	0 &	0 &	0 \\
                             &Mean & 0.5980 & 0.7973 &	1.3953 & 0.3987 & 0.1993 &	0.3987 & 0.7973 & 0.3987 &	0.1993 & 0.7973 & 0.7973 \\
                             &Var  & 2.133 & 2.6767	& 3.806 & 1.5057 &	0.7947 & 1.5057	& 2.6767 &	 1.5057 & 0.79466 &	2.6767 & 2.6767 \\
                             &Dist &0.2999 & 0.3999 & 0.6999 &	0.2	& 0.09999& 	0.2 &	0.3999 &	0.2 &	0.09999 &	0.3999 & 0.3999 \\
 \hline
 \multirow{4}{*}{${f_2}$}   &Best &2.66e-15&	2.66e-15&	2.66e-15&	2.66e-15&	2.66e-15&	 2.66e-15&	2.66e-15&	2.66e-15&	6.22e-15&	2.66e-15&	6.22e-15\\
                             &Mean &5.86e-15&	5.68e-15	&6.04e-15&	5.86e-15&	5.51e-15&	 5.51e-15&	5.68e-15&	5.68e-15&	6.22e-15&	6.04e-15&	6.22e-15\\
                             &Var & 1.20e-30&	1.69e-30&	6.31e-31&	1.20e-30&	2.13e-30&	 2.13e-30&	1.69e-30&	1.69e-30&	0	& 6.31e-31&	0 \\
                             &Dist & 5.55e-14&	5.34e-14&	5.71e-14&	5.39e-14&	5.35e-14&	 5.36e-14&	5.27e-14&	5.43e-14&	5.79e-14&	5.66e-14&	5.70e-14\\
\hline
\multirow{4}{*}{${f_3}$}   & Best & 1.72e-192 &	3.16e-193&	2.30e-194&	2.25e-194&	1.10e-194&	 1.73e-194&	1.95e-193&	1.29e-194&	1.18e-195&	2.14e-195&	6.84e-195\\
                             &Mean  & 4.73e-187&	1.06e-185&	8.21e-187&	4.64e-188&	1.06e-184&	 6.70e-188&	1.53e-186&	4.18e-188&	2.44e-187&	9.04e-188&	3.05e-185\\
                             &Var   & 0&	0&	0&	0&	0&	0&	0&	0&	0&	0&	0 \\
                             &Dist  & 1.36e-93&	5.21e-93&	1.25e-93&	3.99e-94&	1.09e-92&	 4.95e-94&	1.54e-93&	3.80e-94&	8.66e-94&	6.81e-94&	6.29e-93 \\
\hline
\multirow{4}{*}{${f_4}$}    &Best &4.9748& 2.9849&	4.9748&	2.9849&	5.9698&	5.9698&	4.9748&	5.9698&	 4.9748&	3.9798&	4.9748  \\
                             &Mean &42.129 &33.681&	25.801&	36.415&	35.321&	41.014&	31.766&	33.768&	 40.713&	35.916&	44.553  \\
                             &Var  &1148 &790.32&	782.65&	1050.8&	1068.3&	983.76&	1127.1&	1090.5&	 1115.5&	1184&	1290.4 \\
                             &Dist & 14.202&12.842&	12.735&	13.968&	12.871&	14.386&	12.713&	13.309&	 13.435&	13.335&	14.058\\
\hline
\multirow{4}{*}{${f_5}$}    &Best & 0&	0&	0&	0&	0&	0&	0 &	0&	0&	0&	0 \\
                             &Mean  &6.40e-03&	4.44e-03&	3.70e-03&	2.46e-03&	6.16e-03&	 2.96e-03&	2.34e-03&	3.82e-03&	5.18e-03&	6.16e-03&	3.08e-03 \\
                             &var   & 5.19e-05&	2.91e-05&	2.46e-05&	2.30e-05&	5.78e-05&	 2.28e-05&	1.82e-05&	4.12e-05&	2.93e-05&	4.31e-05&	3.25e-05 \\
                             &Dist  & 4.6522&	3.7506&	3.3293&	2.106&	4.5919&	2.5793&	1.9941&	2.5779&	 4.1793&	4.9742&	2.4003\\
\hline
\multirow{4}{*}{${f_6}$ } & Best &3.23e-03&1.09e-03&3.89e-03&5.40e-04&	1.19e-03&3.29e-03&	5.35e-04&	 2.03e-03&	3.35e-03&	5.32e-04&	5.10e-04\\
                             &Mean & 1.0868& 0.3263 &0.3506 &0.0958 &0.6177& 0.2773 &0.7498 &0.9980 &0.6868 &3.3764 &0.6231 \\
                             &Var   &3.8065 &0.4366 &0.3968 &0.0149& 2.1068 &0.2543& 0.9155 &4.2390 &3.0450 &	128.9300& 	1.0124 \\
                             &Dist  &3.1982& 1.5674& 1.9328&1.0814 &2.3737& 1.7252& 2.9348& 3.0664& 2.3291 &3.7989 &2.5342 \\
\hline
 \multirow{4}{*}{${f_7}$ } & Best &1.23e-151&	1.19e-166&	6.09e-174&	1.75e-182&	1.64e-150&	 5.93e-190&	3.33e-168&	5.81e-173&	3.47e-168&	4.69e-175&	7.21e-187 \\
                             &Mean  &4.11e-16&	4.27e-16&	2.69e-16&	2.03e-16&	3.86e-16&	 1.50e-15&	2.86e-16&	3.05e-16&	3.58e-16&	1.61e-16&	3.14e-16\\
                             &Var   &2.80e-31&	3.79e-31&	1.63e-31&	9.48e-32&	1.72e-31&	 7.74e-30&	1.89e-31&	9.61e-32&	1.93e-31&	6.00e-32&	1.05e-31\\
                             &Dist  &4.5089&	4.716&	4.3668&	3.5245&	4.3919&	8.8051&	4.1528&	4.1528&	 4.554&	3.1052&	4.0508 \\
  \hline
 \multirow{4}{*}{${f_8}$ } &Best &0 &	0 &	0 &	0 &	0 &	0 &	0 &	0 &	0 &	0 &	0 \\
                             &Mean  &0 &	0 &	0 &	0 &	0 &	0 &	0 &	0 &	0 &	0 &	0 \\
                             &Var  &0 &	0 &	0 &	0 &	0 &	0 &	0 &	0 &	0 &	0 &	0 \\
                             &Dist  &1348.7&	1319.5&	1317&	1340.6&	1317.1&	1295.9&	1386.8&	1322&	 1354.3&	1310.2&	1333.8 \\
  \hline
 \multirow{4}{*}{${f_9}$ } &Best &-3.64e-12&	118.44&	-3.64e-12&	-3.64e-12&	-3.64e-12&	236.88&	 -3.64e-12&	-13182&	-4568.3&	-3.64e-12&	-3.64e-12\\
                             &Mean  &265.5&	484.61&	2642&	370.12&	1599.4&	983.04&	1034.8&	-6474.3&	 1346.3&	222.07&	318.8\\
                             &Var  &39026&	88492&	3.88e+06&	66401&	3.02e+06&	6.10e+05&	 2.86e+06&	1.40e+07&	6.37e+06&	53850&	48821\\
                             &Dist  &1566.4&	2904.8&	12783&	2197.8&	4777.1&	6005&	3977.6&	40105&	 25694&	1293.4&	1891.9\\
 \hline
\end{tabular}
\end{adjustwidth}
\end{table}

\begin{table}[h]
\begin{adjustwidth}{-1.7cm}{}
\centering
\tiny
\caption{Comparison of PSO-w for functions ${f_1}$-${f_9}$ with different initialization methods. \label{Table-7} }
\begin{tabular}{|p{0.2cm}|p{0.45cm}|p{1.06cm}<{\centering}|p{1.08cm}<{\centering}|p{1.06cm}<{\centering}|p{1.06cm}<{\centering}|p{1.06cm}<{\centering}|p{1.06cm}<{\centering}|p{1.2cm}<{\centering}|p{1.2cm}<{\centering}|p{1.3cm}<{\centering}|p{1.09cm}<{\centering}|p{1.12cm}<{\centering}|}
\hline
 Fun & Value  & $ Be(3,2)$  & $ Be(2.5,2.5)$ & $Be(2,3)$ & $U(0,1)$ & $N(0,1)$ & $N(0.5,1)$ & $N(0.5,0.5)$ & $logn(0,1)$ & $logn(.69,.25)$ & $logn(0,0.5)$ & $logn(0,2/3)$ \\
\hline
 \multirow{4}{*}{${f_1}$} & Best &1.90e-08&	16.383&	17.351&	14.452&	17.431&	16.145&	17.047&	14.375&	 15.049&	1.9259&	8.11e-05\\
                             &Mean  &12.924&	18.405&	18.932&	18.724&	19.349&	18.744&	21.668&	20.457&	 18.064&	15.289&	14.776 \\
                             &Var   &614.41&	1.5182&	1.3712&	2.8465&	2.5575&	3.8338&	189.06&	134.23&	 2.7577&	32.079&	37.001 \\
                             &Dist  &4.0164&	20.043&	20.546&	20.248&	20.682&	20.308&	20.005&	18.914&	 19.624&	16.927&	16.334 \\
\hline
\multirow{4}{*}{${f_2}$}  &Best& 4.75e-07&	5.49e-07&	3.59e-07&	6.52e-07&	7.02e-07&	7.75e-07&	 4.56e-07&	7.76e-07&	1.09e-06&	4.39e-07&	4.40e-07\\
                             &Mean &1.53e-01&	1.07e-01&	4.70e-02&	8.50e-04&	3.66e-03&	 3.20e-02&	1.16e-01&	7.52e-02&	6.06e-02&	5.96e-02&	1.89e-01\\
                             &Var  & 1.37e-01&	1.28e-01&	4.37e-02&	8.82e-06&	6.29e-05&	 2.04e-02&	1.26e-01&	1.13e-01&	6.65e-02&	6.65e-02&	2.21e-01 \\
                             &Dist & 3.09e-01&	2.79e-01&	9.16e-02&	4.78e-03&	1.96e-02&	 4.92e-02&	2.68e-01&	2.28e-01&	1.47e-01&	1.42e-01&	5.06e-01\\
\hline
\multirow{4}{*}{${f_3}$}   & Best &4.77e-14&	4.50e-14&	8.96e-14&	6.51e-14&	8.40e-14&	 6.46e-14&	1.02e-13&	4.78e-14&	6.51e-14&	5.85e-14&	7.13e-14\\
	
                             &Mean  & 1.49e-13&	1.26e-13&	1.98e-13&	1.39e-13&	1.42e-13&	 1.72e-13&	2.09e-13&	1.81e-13&	1.50e-13&	1.57e-13&	1.74e-13\\
                             &Var   & 6.53e-27&	2.35e-27&	7.91e-27&	3.49e-27&	1.37e-27&	 3.42e-27&	1.63e-26&	8.98e-27&	3.88e-27&	3.65e-27&	5.97e-27 \\
                             &Dist  & 1.68e-06&	1.53e-06&	1.92e-06&	1.60e-06&	1.63e-06&	 1.79e-06&	1.97e-06&	1.81e-06&	1.67e-06&	1.70e-06&	1.82e-06\\
\hline
\multirow{4}{*}{${f_4}$}    &Best &16.915&	6.9659&	17.91&	9.9548&	10.947&	12.938&	15.923&	13.931&	 12.936&	15.92&	13.931   \\
                             &Mean  &27.668& 16.933 &	28.767&	22.39&	24.428&	21.397&	22.79&	23.14&	 21.165&	27.765&	25.275 \\
                             &Var   &43.693& 35.101&	62.087&	39.554&	37.565&	41.602&	28.926&	42.563&	 25.969&	35.863&	47.994 \\
                             &Dist  &19.907& 14.339&	20.704&	17.617&	18.361&	16.823&	17.618&	17.468&	 16.475&	19.509&	18.91 \\
\hline
\multirow{4}{*}{${f_5}$}    &Best & 1.43e-10&	1.48e-10&	2.01e-10&	2.99e-10&	1.62e-10&	 1.36e-10&	4.01e-10&	2.38e-10&	2.22e-10&	9.77E-11&	2.41E-10 \\
                             &Mean  &8.99e-03&	8.62e-03&	8.00e-03&	1.12e-02&	8.62e-03&	 9.85e-03&	9.36e-03&	1.02e-02&	1.03e-02&	9.11e-03&	9.59e-03\\
                             &Var   &6.34e-05&	3.09e-05&	9.18e-05&	1.28e-04&	7.49e-05&	 1.12e-04&	8.82e-05&	9.96e-05&	1.07e-04&	1.14e-04&	1.31e-04 \\
                             &Dist  & 7.3018&	7.0998&	5.8384&	7.0651&	6.3561&	6.5636&	7.0443&	7.454&	 6.9119&	6.0607&	6.2669\\
\hline
\multirow{4}{*}{${f_6}$} &Best &9.18e-05&	4.88e-06 &	6.42e-05&	4.53e-05&	7.86e-05&	3.58e-05&	 5.92e-05&	7.84e-05&	4.38e-05&	1.87e-04&	1.82e-04\\
                             &Mean  &9.54e-04&	1.57e-04 &	1.35e-03&	4.32e-04&	3.94e-04&	 8.71e-04&	9.79e-04&	9.28e-04&	6.74e-04&	1.30e-03&	1.78e-03 \\
                             &Var   &1.05e-06&	2.82E-08 &	1.29e-06&	6.05e-08&	8.45e-08&	 1.41e-06&	1.08e-06&	1.01e-06&	3.98e-07&	1.46e-06&	1.28e-05 \\
                             &Dist  &1.19e-01&	4.75e-02 &	1.46e-01&	8.63e-02&	8.08e-02&	 1.15e-01&	1.23e-01&	1.19e-01&	1.05e-01&	1.47e-01&	1.40e-01\\
\hline
 \multirow{4}{*}{${f_7}$} &Best &1.29e-02&	1.76e-03&	1.78e-02&	3.82e-03&	5.48e-03&	5.26e-03&	 6.30e-03&	9.67e-03&	3.52e-03&	1.23e-02&	1.30e-02 \\
                             &Mean  &6.99e-02&	1.85e-02&	4.51e-02&	2.58e-02&	3.26e-02&	 2.32e-02&	5.06e-02&	3.72e-02&	3.13e-02&	5.30e-02&	5.97e-02 \\
                             &Var   &3.52e-03&	1.58e-04&	6.32e-04&	4.34e-04&	3.76e-04&	 3.02e-04&	3.33e-03&	3.05e-04&	6.72e-04&	4.64e-03&	1.54e-03 \\
                             &Dist  &39.429&	14.155 &	35.415&	19.585&	21.599&	19.558&	26.302&	 23.057&	19.963&	29.271&	32.17 \\
 \hline
 \multirow{4}{*}{${f_8}$} &Best &0 &	0  &	0 &	0 &	0 &	0 &	0 &	0 &	0 &	0 &	0 \\
                             &Mean  &0 &	0 &	0 &	0 &	0 &	0 &	0 &	0 &	0 &	0 &	0 \\
                             &Var  &0 &	 0  &	0 &	0 &	0 &	0 &	0 &	0 &	0 &	0 &	0 \\
                             &Dist  &1091.3&	1012.2 &	1158.5&	1514.6&	5500.2&	4678.4&	2247.6&	 6917.9&	9429.8&	3703&	4829.9\\
  \hline
 \multirow{4}{*}{${f_9}$} &Best & 2053.3 &	2112.8&	2546.5&	2151.6&	-6422.2&	-3147.9&	 4490.9&-1.04e+05&	-13842&	-2612.6&	-18992\\
                             &Mean  & 2811.1 &	3320.5&	3670.8&	2859.7&	-2435.2&	-1415.8&	5569&	 -51602&	-10275&	57.926&	-7186.7\\
                             &Var  & 1.70e+05&	4.02e+05&	1.91e+05&	1.54e+05&	2.11e+06&	 1.65e+06&	2.69e+05&	4.07e+08&	3.66e+06&	1.91e+06&	1.25e+07\\
                             &Dist  & 6586.7&	9418.9&	14068&	9433.5&	37369&	30269&	17740&	98148&	 37286&	18758&	31444 \\
\hline
\hline
Fun &   & $ E(0.5)$  & $E(0.1) $ & $E(0.8)$ & $Rayl(0.4)$ & $Rayl(0.8)$ & $Rayl(0.1)$ & $Weib(1,1.5)$ & $Weib(1.5,1)$ & $Weib(1,1)$ & $random$ & $LHS$ \\
\hline
 \multirow{4}{*}{${f_1}$} &Best &16.449&	13.908&	13.012&	17.453&	7.95e-04&	13.064&	2.46e-01&	 6.1102&	1.3975&	14.909&	16.213 \\
                             &Mean &19.28&	18.472&	17.983&	18.637&	16.005&	19.003&	20.822&	18.077&	 20.893&	18.03&	18.369 \\
                             &Var  & 1.6701&	3.7361&	6.84837&	1.3217&	25.016&	4.4588&	358.48&	 13.88&	183.51&	3.565&	2.2941 \\
                             &Dist &20.954&	19.898&	19.395&	20.132&	17.446&	20.556&	15.883&	19.229&	 19.228&	19.393&	19.624 \\
 \hline
 \multirow{4}{*}{${f_2}$}   &Best &4.60e-07&	5.37e-07&	5.14e-07&	3.88e-07&	3.89e-07&	 5.01e-07&	3.82e-07&	3.92e-07&	4.43e-07&	4.80E-07&	6.11E-07\\
                             &Mean &1.36e-01&	6.11e-02&	4.66e-02&	4.67e-02&	8.24e-02&	 1.88e-01&	1.94e-03&	1.52e-01&	1.97e-01&	6.15e-02&	2.26e-05\\
                             &Var & 1.70e-01&	6.64e-02&	4.34e-02&	4.34e-02&	1.36e-01&	 2.46e-01&	5.95e-05&	1.37e-01&	1.87e-01&	6.70e-02&	1.42e-09 \\
                             &Dist &3.71e-01&	1.48e-01&	8.26e-02&	8.35e-02&	2.77e-01&	 5.62e-01&	1.06e-02&	3.07e-01&	5.34e-01&	1.59e-01&	1.35e-04 \\
\hline
\multirow{4}{*}{${f_3}$}   &Best & 9.53e-14&	4.69e-14&	8.81e-14&	4.12e-14&	6.00e-14&	 3.43e-14&	4.76e-14&	7.89e-14&	8.11e-14&	4.34e-14&	7.52e-14\\
                             &Mean  & 1.61e-13&	1.53e-13&	1.66e-13&	1.38e-13&	1.63e-13&	 1.71e-13&	1.48e-13&	1.65e-13&	1.68e-13&	1.66e-13&	1.21e-13\\
                             &Var   & 3.76e-27&	5.19e-27&	2.13e-27&	4.08e-27&	4.76e-27&	 5.07e-27&	4.41e-27&	4.80e-27&	6.63e-27&	5.33e-27&	1.70e-27 \\
                             &Dist  & 1.76e-06&	1.67e-06&	1.76e-06&	1.59e-06&	1.73e-06&	 1.74e-06&	1.64e-06&	1.74e-06&	1.78e-06&	1.77e-06&	1.53e-06\\
\hline
\multirow{4}{*}{${f_4}$}    &Best &12.934&	20.931&	14.929&	10.959&	14.006&	26.864&	17.91&	14.935&	 11.94&	11.944&	15.921 \\
                             &Mean &25.526&	38.759&	21.196&	19.801&	25.846&	38.663&	29.556&	23.736&	 22.745&	20.113&	22.641  \\
                             &Var  &60.968&	113.88&	21.165&	28.328&	53.31&	63.94&	63.415&	25.204&	 61.3&	24.206&	34.464\\
                             &Dist &18.662&	23.836&	16.721&	16.122&	19.268&	23.936&	20.801&	17.967&	 17.626&	16.326&	17.47 \\
\hline
\multirow{4}{*}{${f_5}$}    &Best &1.13e-10&	3.15e-10 &	1.47e-10&	2.58e-10&	2.43e-10&	 3.14e-10&	2.19e-10&	2.86e-10&	1.38e-10&	1.19e-10&	2.52e-10 \\
                             &Mean  &6.64e-03&	7.14e-03 &	9.11e-03&	1.29e-02&	1.22e-02&	 1.08e-02&	1.41e-02&	8.37e-03&	9.96e-03&	8.74e-03&	1.24e-02 \\
                             &Var   &1.32e-04&	7.52e-05&	5.68e-05&	1.95e-04&	1.62e-04&	 1.44e-04&	1.86e-04&	1.77e-04&	2.39e-04&	9.21e-05&	1.90e-04\\
                             &Dist  &4.153&	5.4522&	6.5607&	8.0532&	7.823&	6.8695&	8.2484&	5.5116&	 6.1723&	5.7461&	7.3272 \\
\hline
\multirow{4}{*}{${f_6}$} &Best &1.04e-04&	7.86e-05&	4.82e-05&	2.44e-05&	1.76e-04&	2.22e-04&	 5.56e-05&	4.07e-05&	4.57e-05&	2.27e-05&	7.34e-05\\
                             &Mean &7.91e-04&	4.89e-03&	7.57e-04&	1.86e-04&	1.47e-03&	 3.98e-03&	1.04e-03&	4.37e-04&	7.40e-04&	4.28e-04&	3.26e-04  \\
                             &Var   &7.00e-07&	4.87e-05&	3.41e-07&	2.93e-08&	1.48e-06&	 2.01e-05&	8.13e-07&	7.49e-08&	7.59e-07&	1.45e-07&	5.26e-08 \\
                             &Dist  &1.07e-01&	2.59e-01&	1.09e-01&	5.41e-02&	1.56e-01&	 2.41e-01&	1.28e-01&	8.73e-02&	1.06e-01&	8.21e-02&	7.31e-02 \\
\hline
 \multirow{4}{*}{${f_7}$} &Best &4.07e-03&	2.29e-02&	2.09e-03&	2.17e-03&	9.49e-03&	1.61e-02&	 7.42e-03&	1.11e-02&	1.20e-02&	7.28e-03&	6.92e-03\\
                             &Mean  &3.42e-02&	6.99e-02&	2.67e-02&	2.77e-02&	3.65e-02&	 7.81e-02&	5.73e-02&	2.87e-02&	4.88e-02&	3.32e-02&	2.88e-02\\
                             &Var   &8.60e-04&	1.34e-03&	3.08e-04&	8.30e-04&	6.89e-04&	 2.96e-03&	2.39e-03&	6.56e-04&	1.36e-03&	8.67e-04&	2.02e-04\\
                             &Dist  &21.761&	44.023&	18.272&	17.403&	31.41&	51.998&	32.112&	22.349&	 25.514&	20.956&	20.691 \\
  \hline
 \multirow{4}{*}{${f_8}$} &Best &0 &	0 &	0 &	0 &	0 &	0 &	0 &	0 &	0 &	0 &	0 \\
                             &Mean  &0 &	0 &	0 &	0 &	0 &	0 &	0 &	0 &	0 &	0 &	0 \\
                             &Var  &0 &	0 &	0 &	0 &	0 &	0 &	0 &	0 &	0 &	0 &	0 \\
                             &Dist  &2144.8&	2460.6&	3292.1&	1221.1&	3123.6&	2346.7&	3268.4&	6778.9&	 4051.7&	1487&	1562.7\\
  \hline
 \multirow{4}{*}{${f_9}$} &Best &3851.7&	2151.8&	-3501.2&	2211.4&	478.1&	2526.7&	495.23&	-19232&	 -8430.5&	1934.6&	1638.4\\
                             &Mean &5721.5&	2866.3&	-686.19&	3238.6&	2372.5&	2988.6&	2509.5&	-13907&	 -4593.5&	2693.6&	2569.2\\
                             &Var  &1.08e+06&2.47e+05&2.19e+06&3.00e+05&1.41e+06&93053&	 1.04e+06&6.64e+06&3.64e+06&1.48e+05&1.84e+05\\
                             &Dist  &18835&	11125&	25640&	9832.4&	14121&	15581&	17327&	48313&	 30333&	9012.9&	7984\\
 \hline
\end{tabular}
\end{adjustwidth}
\end{table}

\begin{table}[h]
\begin{adjustwidth}{-1.7cm}{}
\centering
\tiny
\caption{Comparison of CS for functions ${f_1}$-${f_9}$ with different initialization methods. \label{Table-8} }
\begin{tabular}{|p{0.2cm}|p{0.45cm}|p{1.06cm}<{\centering}|p{1.08cm}<{\centering}|p{1.06cm}<{\centering}|p{1.06cm}<{\centering}|p{1.06cm}<{\centering}|p{1.06cm}<{\centering}|p{1.2cm}<{\centering}|p{1.2cm}<{\centering}|p{1.3cm}<{\centering}|p{1.09cm}<{\centering}|p{1.12cm}<{\centering}|}
\hline
 Fun & Value  & $ Be(3,2)$  & $ Be(2.5,2.5)$ & $Be(2,3)$ & $U(0,1)$ & $N(0,1)$ & $N(0.5,1)$ & $N(0.5,0.5)$ & $logn(0,1)$ & $logn(.69,.25)$ & $logn(0,0.5)$ & $logn(0,2/3)$ \\
\hline
 \multirow{4}{*}{${f_1}$} &Best &7.49e-25&	2.51e-12&	4.78e-12&	5.09e-12&	4.43e-13&	3.88e-12&	 5.26e-12&	1.25e-23&	9.93e+05&	5.85e-23&	9.71e-29\\
                             &Mean  &6.52e-03&	9.11e-10&	4.08e-08&	1.99e-01&	1.99e-01&	 1.52e-08&	4.30e-09&	3.00e-07&	1.10e+06&	2.02e-01&	1.39e-04 \\
                             &Var   &6.24e-04&	5.33e-18&	2.01e-14&	7.95e-01&	7.95e-01&	 2.58e-15&	1.35e-16&	1.08e-12&	2.58e+09&	7.94E-01&	1.89E-07 \\
                             &Dist  &4.93e-02&	1.29e-05&	7.41e-05&	1.00e-01&	1.00e-01&	 5.69e-05&	3.88e-05&	1.34e-04&	1.16e+02&	1.28e-01&	2.42e-03\\
\hline
\multirow{4}{*}{${f_2}$}  &Best & 2.66e-15&	2.66e-15&	2.66e-15&	2.66e-15 &	2.66e-15 &	2.66e-15&	 2.66e-15&	2.66e-15 &	1.70e+01&	2.66e-15 &	2.66e-15\\
                             &Mean &9.31e-02&	2.66e-15&	2.66e-15&	2.66e-15 &	2.66e-15 &	 2.66e-15 &	9.31e-02&	2.66e-15 &	1.71e+01&	2.66e-15&	4.66e-02\\
                             &Var  & 8.22e-02&	0 &	0 &	0 &	 0 & 0 &	8.22e-02&	0&	7.48e-03&	0 &	 4.34e-02 \\
                             &Dist &1.64e-01&	3.15e-14 &	3.22e-14&	3.20e-14&	3.13e-14&	 3.17e-14 &	1.64e-01 &	3.22e-14 &	2.84e+02&	3.12e-14&	8.21e-02 \\
\hline
\multirow{4}{*}{${f_3}$}  & Best &5.16e-69&	5.03e-70&	3.44e-69&	3.26e-69&	1.78e-69&	9.30e-70&	 3.90e-69&	3.10e-69&	625&	1.27e-68&	8.41e-69\\
	
                             &Mean  &4.59e-67&	9.48e-68&	1.43e-67&	1.83e-67&	5.45e-67&	 2.18e-67&	1.91e-67&	4.46e-67&	680&	5.06e-67&	2.31e-67 \\
                             &Var   &1.88e-132&	1.98e-134&	9.08e-134&	5.36e-134&	1.80e-132&	 1.59e-133&	1.10e-133&	3.59e-133&	894.74&	1.04e-132&	5.64e-134  \\
                             &Dist  &1.81e-33&	1.01e-33&	1.27e-33&	1.56e-33&	2.09e-33&	 1.66e-33&	1.52e-33&	2.29e-33&	136&	2.38e-33&	1.84e-33 \\
\hline
\multirow{4}{*}{${f_4}$}    &Best &3.3086&	2.9396&	3.7509&	2.0714&	5.7085&	2.0919&	0.27923&	4.5855&	 743.94&	7.8894&	7.0934  \\
                             &Mean &8.5715&	6.2083&	8.9598&	8.4141&	10.208&	8.6322&	7.9627&	10.67&	 819.93&	12.461&	11.592 \\
                             &Var  &8.845&	6.8603&	9.6093&	7.8935&	12.2&	10.466&	11.069&	5.7495&	 982.17&	5.9648&	7.5269 \\
                             &Dist  &7.3607& 5.1551&	7.7512&	7.3389&8.9385&	7.5962&	7.0386&	9.3112&	 145.97&	10.095&	10.356 \\
\hline
\multirow{4}{*}{${f_5}$}    &Best & 0&	0&	0&	0&	0&	0&	0&	0&	2071&	0&	0\\
                             &Mean  &7.39e-04&	4.93e-04&	0&	0&	3.69e-04&	0&	3.69e-04&	0&	 25.5&	0& 7.39e-04\\
                             &Var   &5.18e-06&	4.86e-06&	0&	0&	2.74e-06&	0&	2.74e-06&	0&	 28030&	0&	5.18e-06 \\
                             &Dist  &7.57e-01&	4.29e-01&	5.37e-07&	5.54e-07&	3.79e-01&	 5.51e-07&	3.79e-01&	5.79e-07&	16350&	5.60e-07&	7.58e-01\\
\hline
\multirow{4}{*}{${f_6}$ } &Best &7.74e-04&	1.70e-04 &	5.83e-04&	1.93e-03&	4.82e-03&	4.74e-03&	 3.50e-03&	3.64e-03&	4.73e+16&	2.61e-03&	3.75e-03\\
                             &Mean  &1.46e-02&	1.40e-03 &	1.54e-02&	1.33e-02&	7.30e-02&	 8.48e-02&	6.46e-02&	5.06e-02&	1.63e+17&	4.02e-02&	1.20e-01 \\
                             &Var   &1.56e-04&	8.78e-07 &	3.87e-04&	9.05e-05&	1.42e-02&	 7.70e-03&	1.40e-02&	2.82e-03&	3.98e+33&	1.41e-03&	4.09e-02 \\
                             &Dist  &4.78e-01&	1.52e-01&	4.30e-01&	4.68e-01&	9.64e-01&	1.14&	 9.09e-01&	8.81e-01&	3.00e+03&	7.52e-01&	1.18\\
\hline
 \multirow{4}{*}{${f_7}$ } &Best &3.26e-03&	1.45e-04&	9.22e-02&	3.62e-01&	3.53e-01&	5.04e-01&	 7.64e-01&	1.50&	1.11e+02&	6.13e-01&	8.81e-01\\
                             &Mean  &5.12e-01&	3.98e-01&	5.68e-01&	8.89e-01&	1.82&	1.89&	 1.66&	2.39&	1.23E+02&	1.79& 1.91 \\
                             &Var   &1.17e-01&	1.35e-01&	1.17e-01&	2.61e-01&	9.15e-01&	 4.46e-01&	3.13e-01&	2.63e-01&	5.00e+01&	3.30e-01&	2.97e-01 \\
                             &Dist  &5.13e+01&	3.28e+01&	5.30e+01&	5.77e+01&	9.44e+01&	 9.15e+01&	8.32e+01&	1.16e+02&	2.99e+02&	1.27e+02&	1.22e+02 \\
 \hline
 \multirow{4}{*}{${f_8}$ } &Best &-1.65e-01&	-2.52e-01&	-1.70e-01&	-2.89e-01&	-2.12e-01&	 -1.45e-01&	-1.49e-01&	-9.56e-02 &	-1.76e-01&	-1.44e-01&	-1.26e-01 \\
                             &Mean  &-8.40e-02&	-9.80e-02&	-8.44e-02&	-8.85e-02&	-9.34e-02&	 -7.81e-02&	-7.59e-02&	-6.32e-02&	-7.99e-02&	-7.47e-02&	-6.86e-02 \\
                             &Var  & 8.11e-04&	1.62e-03&1.41e-03&2.67e-03&	1.72e-03&1.09e-03&7.88e-04&	 2.26e-04&1.33e-03&7.58e-04&5.41e-04\\
                             &Dist & 1.70e+03&	1.51e+03&	1.82e+03&	1.78e+03&	1.85e+03&	 1.86e+03&	1.83e+03&	1.79e+03&	1.85e+03&	1.93e+03&	1.84e+03\\
  \hline
 \multirow{4}{*}{${f_9}$ } &Best &3.95e-04&	385.34&	724.82&	9.0398&	-38.177&	243.62&	143.57&	-25579&	 -4394.8& 3.82e-04& 3.82e-04\\
                             &Mean  &280.18&	722.53&	1174.2&	613.81&	1459.4&	820.01&	749.11&	 -4934.2&	-266.16&	51.42&	307.49\\
                             &Var  &53694&	55598&	86599&	93668&	3.50e+05&	1.22e+05&	89298&	 3.88e+07&	3.81e+06&	24883&	3.10e+05\\
                             &Dist  &1222.4&	3459.5&	5595.9&	3023.7&	13250&	4955.3&	3812.9&	38282&	 30198&	229.96&	1692.6 \\
\hline
\hline
Fun &   & $ E(0.5)$  & $E(0.1) $ & $E(0.8)$ & $Rayl(0.4)$ & $Rayl(0.8)$ & $Rayl(0.1)$ & $Weib(1,1.5)$ & $Weib(1.5,1)$ & $Weib(1,1)$ & $random$ & $LHS$ \\
\hline
 \multirow{4}{*}{${f_1}$ } &Best &4.24e-12&	1.00e-11&	8.65e-12&	1.25e-12&	1.44e-25&	7.96e-12&	 2.28e-25&	1.18e-21&	1.01e-13&	1.38e-12&	3.11e-12 \\
                             &Mean &5.47e-10 &	1.99e-01&	1.99e-01&	6.55e-10&	1.37e-01&	 1.99e-01&	1.07e-02&	5.04e-04&	9.87e-02&	1.99e-01&	1.99e-01 \\
                             &Var  & 1.14e-18 &	7.95e-01&	7.95e-01& 1.63e-18 &	3.72e-01&	 7.95e-01&	2.25e-03&	5.07e-06&	1.89e-01&	7.95e-01&	7.95e-01  \\
                             &Dist &1.72e-05 &	1.00e-01&	1.00e-01&	1.25e-05 &	2.34e-01&	 1.00e-01&	7.05e-02&	1.30e-02&	2.15e-01&	1.00e-01&	1.00e-01\\
 \hline
 \multirow{4}{*}{${f_2}$}   &Best &2.66e-15&	2.66e-15&	2.66e-15&	2.66e-15&	2.66e-15 &	 2.66e-15&	2.66e-15&	2.66e-15& 2.66e-15&	2.66e-15&	2.66e-15\\
                             &Mean &4.66e-02&	4.66e-02&	4.66e-02&	5.78e-02 &2.66e-15 &	 5.78e-02&	5.78e-02&	4.66e-02& 2.66e-15 &	9.31e-02&	2.66e-15\\
                             &Var & 4.34e-02&	4.34e-02&	4.34e-02&	6.67e-02& 0 &	6.67e-02&	 6.67e-02&	4.34e-02& 0 &	8.22e-02& 0 \\
                             &Dist &8.21e-02&	8.21e-02&	8.21e-02&	1.31e-01&	3.16e-14&	 1.31e-01&	1.31e-01&	8.21e-02& 3.21e-14&	1.64e-01&	3.16e-14\\
\hline
\multirow{4}{*}{${f_3}$}   &Best &3.27e-69&	7.05e-69&	3.35e-70&	2.84e-70&	2.82e-69&	1.38e-69&	 3.33e-69&	3.47e-69&	2.32e-69&	6.93e-69&	5.24e-70 \\
                             &Mean  &3.29e-67&	7.26e-67&	1.51e-67&	1.06e-67&	1.03e-66&	 7.00e-67&	1.78e-67&	4.17e-67&	1.29e-66&	1.85e-67&	1.18e-67 \\
                             &Var   &1.86e-133&	2.26e-132&	3.04e-134&	5.92e-134&	1.13e-131&	 1.90e-132&	1.46e-133&	2.45e-133&	1.76e-131&	1.27e-133&	1.24e-134  \\
                             &Dist  &2.01e-33&	2.71e-33&	1.47e-33&	1.01e-33&	2.59e-33&	 2.65e-33&	1.46e-33&	2.37e-33&	2.44e-33&	1.51e-33&	1.32e-33 \\
\hline
\multirow{4}{*}{${f_4}$}    &Best &6.9979&	6.6867&	3.886&	5.1891&	5.5713&	6.5398&	2.8385&	4.268&	 2.9355&	2.1314&	1.4139 \\
                             &Mean &10.761&	11.877&	9.3383&	8.1427&	10.574&	11.839&	7.9303&	10.917&	 9.9632&	7.3846&	8.8921  \\
                             &Var  &4.4353&	6.3303&	8.5221&	3.2042&	10.417&	6.4774&	6.2093&	11.118&	 9.9866&	6.608&	10.793\\
                             &Dist &9.1505&	10.289&	8.0989&	7.0184&	8.9637&	10.381&	6.834&	9.4743&	 8.6558&	6.575&	7.8213 \\
\hline
\multirow{4}{*}{${f_5}$}    &Best &0&	0&	0&	0&	0&	0&	0&	0&	0&	0&	0\\
                             &Mean  &0&	0&	0&	0&	0&	0&	0&	0&	3.69e-04&	3.69e-04&	0 \\
                             &Var   &0&	0&	0&	0&	0&	0&	0&	0&	2.74e-06&	2.74e-06&	0\\
                             &Dist  &5.39e-07&	5.29e-07&	5.37e-07&	5.48e-07&	5.84e-07&	 5.47e-07&	5.50e-07&	5.22e-07&	3.79e-01&	3.79e-01&	5.65e-07 \\
\hline
\multirow{4}{*}{${f_6}$ } &Best &8.88e-04&	5.10e-03&	2.52e-03&	6.73e-04&	1.79e-03&	2.01e-03&	 1.58e-03&	3.00e-03&	3.98e-03&	9.87e-04&	2.70e-03\\
                             &Mean &4.04e-02&	6.87e-02&	3.49e-02&	4.46e-03&	3.41e-02&	 1.07e-01&	5.22e-02&	3.59e-02&	4.57e-02&	1.61e-02&	1.46e-02  \\
                             &Var   &1.16e-03&	7.29e-03&	1.25e-03&	1.61e-05&	7.68e-04&	 2.57e-02&	5.11e-03&	1.61e-03&	1.58e-03&	2.89e-04&	1.81e-04 \\
                             &Dist  &7.96e-01&	9.85e-01&	6.99e-01&	2.58e-01&	7.53e-01&	1.13&	 8.43e-01&	7.03e-01&	8.26e-01&	4.88e-01&	4.69e-01 \\
\hline
 \multirow{4}{*}{${f_7}$ } &Best &2.75e-01&	1.63e-02&	5.31e-01&	2.51e-03&	6.82e-01&	1.28e-03&	 6.44e-01&	8.43e-01&	4.79e-01&	9.15E-02&	1.06E-02\\
                             &Mean  &1.22&	1.31&	2.06&	6.86e-01&	1.88&	1.22&	1.97&	2.37&	 2.17&	9.86&	9.49e-01\\
                             &Var   &3.07e-01&	3.52e-01&	7.59e-01&	2.90e-01&	4.29e-01&	 2.14e-01&	8.69e-01&	5.54e-01&	6.51e-01&	2.29e-01&	3.14e-01\\
                             &Dist  &7.26e+01&	1.58e+02&	9.52e+01&	4.94e+01&	1.26e+02&	 1.44e+02&	1.06e+02&	1.21e+02&	9.47e+01&	6.20e+01&	5.93e+01 \\
  \hline
 \multirow{4}{*}{${f_8}$ } &Best & -1.24e-01 &	-2.16e-01&	-1.34e-01 &	-2.31e-01&	-1.44e-01&	 -1.92e-01&	-1.64e-01&	-1.49e-01&	-1.48e-01&	-1.42e-01&	-2.43e-01 \\
                             &Mean  & -7.16e-02 &	-8.33e-02& -6.78e-02&	-7.88e-02&	-7.21e-02 &	 -6.99e-02&	-8.58e-02&	-7.88e-02&	-7.05e-02&	-7.94e-02&	-1.01e-01 \\
                             &Var  & 6.77e-04 &	2.24e-03& 7.63e-04 &	1.70e-03& 7.19e-04&	1.30e-03&	 1.11e-03&	1.07e-03& 9.00e-04&	1.06e-03&	3.08e-03 \\
                             &Dist  & 1.82e+03 &	1.73e+03&	1.89e+03 &	1.74e+03 &	1.80e+03&	 1.83e+03&	1.81e+03&	1.80e+03& 1.84e+03 &	1.74e+03&	1.60e+03\\
 \hline
 \multirow{4}{*}{${f_9}$ } &Best &520.68&	1793.7&	175.86&	437.43& 3.82e-04 &	1594& 3.82e-04&	-12349&	 -1523.4&	459.12&	182.46\\
                             &Mean &968.22&	2160.1&	578.87&	810.59&	 50.817 &	1998.4&	295.42&	 -2165.4&	615.8&	771.45&	791.25\\
                             &Var  &71115&	46239&	98524&	66299&	13395 &	53307&	1.12e+05&	 1.71e+07&	5.24e+05&	53597&	94734\\
                             &Dist  &4857.6&	11135&	2806.9&	3851.3&	203.1 &	11048&	1385.4&	33789&	 8227.6&	3437.3&	3903.5\\
 \hline
\end{tabular}
\end{adjustwidth}
\end{table}

As presented in these three tables, for nine benchmark functions and three different algorithms, four indicators with 22 different initialization methods have been tested and analyzed. It can be seen that initialization methods can have a great influence on the results of these algorithms. For some functions, some initialization methods are significantly better than others.

But for other functions, a quick look seems to give the impression that the results are not quite consistent. {{For example, it seems that some initialization methods can find the `Best' fitness values with a higher accuracy, but their `Mean' fitness values are less accurate. To make sense of the results, we have carried out some statistical analyses, including the Friedman ranking test~\citep{lopez2017extended}, and the results are presented in the next subsection. }}

\subsection{Comparison and Friedman rank test}

{{Four indicators (`Best', `Mean', `Var', and `Dist') are used to analyze the results of each algorithm from different perspectives. In order to compare the effects of the 22 initialization approaches, a useful non-parametric test, known as the Friedman rank test, is used for statistical analyses. }}

Let us start with function $f_1$ in Table~\ref{Table-6}. For the Friedman rank test to compare different initialization methods,  the null hypothesis is that the effects of initialization methods for DE-a are all equal, while its alternative hypothesis is that at least one of the initialization methods may differ from at least one of the others. That is
\begin{equation}\label{}
\begin{array}{l}
{H_{\rm{0}}}:{I_{1}} = {I_{2}} = {I_{3}} =  \cdots  = {I_{22}} \\
{H_1}: \textrm{Not all the initialization effects are equal.}
\end{array}
\end{equation}

In essence, the Friedman rank test begins by ordering the ranks of the initialization methods for test indicators. For indicator `Best', as can be seen from Fig.~\ref{Table-6}, different initializations produce the same results. So the first row of Table~\ref{Table-9} is $11.5 = (1 + 2 +  \cdots  + 21 + 22)/22$. Calculate the rank values for each row with this method (Different values are sorted by traditional sorting methods). The rank values of different initialization methods for $f_1$ are listed in Table~\ref{Table-9}.

\begin{table}[h]
\begin{adjustwidth}{-1.1cm}{}
\centering
\tiny
\caption{Friedman rank values of different initialization methods for $f_1$.  \label{Table-9} }
\begin{tabular}{|p{0.5cm}<{\centering}|p{0.8cm}<{\centering}|p{1.1cm}<{\centering}|p{0.9cm}<{\centering}|p{0.9cm}<{\centering}|p{0.9cm}<{\centering}|p{1cm}<{\centering}|p{1.3cm}<{\centering}|p{1.3cm}<{\centering}|p{1.2cm}<{\centering}|p{1.5cm}<{\centering}|p{1.1cm}<{\centering}|p{1.2cm}<{\centering}|}
\hline
  Rank& $ Be(3,2)$  & $ Be(2.5,2.5)$ & $Be(2,3)$ & $U(0,1)$ & $N(0,1)$ & $N(0.5,1)$ & $N(0.5,0.5)$ & $logn(0,1)$ & $logn(.69,.25)$ & $logn(0,0.5)$ & $logn(0,2/3)$ \\
\hline
Best& 11.5& 11.5& 11.5& 11.5& 11.5& 11.5& 11.5& 11.5& 11.5& 11.5& 11.5 \\
Mean&  21   &  16.5 &  16.5 &  10.5&  16.5 & 16.5 &  10.5 & 10.5  &  6   &  2   &  6     \\
Var&    21  &  16.5   &  16.5   &  10.5   &  16.5   &  16.5   &  10.5 &  10.5 &  6   &   2  &  6   \\
Dist&  21    & 16.5    &   16.5  &   10.5  &  16.5   &  16.5   &  10.5   & 10.5    &  6   &  2   & 6    \\
sum&  74.5   &  61   &   61   &   43  &  61  &   61  &  43    &   43   & 29.5    &  17.5    &  29.5   \\
mean& 18.63& 15.25& 15.25& 10.75& 15.25&15.25 &10.75 &10.75 &7.38 &4.38 &7.38 \\
\hline
\hline
 Rank& $ E(0.5)$  & $E(0.1) $ & $E(0.8)$ & $Rayl(0.4)$ & $Rayl(0.8)$ & $Rayl(0.1)$ & $Weib(1,1.5)$ & $Weib(1.5,1)$ & $Weib(1,1)$ & $random$ & $LHS$ \\
\hline
Best& 11.5& 11.5& 11.5& 11.5& 11.5& 11.5& 11.5& 11.5& 11.5& 11.5& 11.5\\
Mean&  10.5 &  16.5   &  22   &   6  &  2  &   6  &  16.5 &  6   &   2  &  16.5 & 16.5     \\
Var&    10.5  &  16.5    &  22   & 6    &    2  &  6  &  16.5   &  6   &  2   &  16.5   &  16.5    \\
Dist&  10.5    &  16.5    &  22  &  6  &  2   &   6   &  16.5   & 6    & 2    & 16.5    &  16.5   \\
sum&   43  &  61 &   77.5 &   29.5 &    17.5 &   29.5    &   61    &  29.5    & 17.5   &43   &   43  \\
 mean&10.75&15.25&19.38&7.38&4.38&7.38&15.25&7.38&4.38&15.25&15.25\\
\hline
\end{tabular}
\end{adjustwidth}
\end{table}

For such Friedman rank test, we usually focus on the mean rank of different indicators. In the `mean' row in Table~\ref{Table-9},  we can see that the minimum value is 4.38 and there are three such values. For the  ease of observation, the order or rank of initialization corresponding to these three values is 1. The second smallest is 7.38 that repeated five times, so their rank is 2. In this way, the orders of the initialization performance for all different functions and their corresponding two-side $p$-values are given in Table~\ref{Table-10}.

\begin{table}[h]
\begin{adjustwidth}{-1.3cm}{}
\centering
\tiny
\caption{Ranks of different initialization methods for DE-a over functions $f_1-f_9$. \label{Table-10} }
\begin{tabular}{|p{0.2cm}|p{0.8cm}|p{0.75cm}<{\centering}|p{1.1cm}<{\centering}|p{0.8cm}<{\centering}|p{0.9cm}<{\centering}|p{0.9cm}<{\centering}|p{0.9cm}<{\centering}|p{1.2cm}<{\centering}|p{1.2cm}<{\centering}|p{1.3cm}<{\centering}|p{1.15cm}<{\centering}|p{1.2cm}<{\centering}|}
\hline
 Fun&$p{\rm{-}}value$ &$ Be(3,2)$  & $ Be(2.5,2.5)$ & $Be(2,3)$ & $U(0,1)$ & $N(0,1)$ & $N(0.5,1)$ & $N(0.5,0.5)$ & $logn(0,1)$ & $logn(.69,.25)$ & $logn(0,0.5)$ & $logn(0,2/3)$ \\
\hline
${f_1}$  &0.000  & 5 &	4 &	4 &	3 &	4 &	4 &	3 &	3 &	2 &1 &	2 \\
\hline
${f_2}$  & 0.957  & 17 &5 &14 &13 &9 &20 &4 &16 &1 &12 &2 \\
\hline
${f_3}$   & 0.008 & 3& 11 & 11& 8 &7 & 2 &1 &9& 18 &12 & 15\\
\hline
${f_4}$    & 0.002 & 13 & 6 & 5 &18 & 19 & 1 &17 &4 &11 &10 & 11 \\
\hline
${f_5}$    & 0.000& 10 & 9 & 15 & 18 & 11 & 6 &14 & 13 & 6 &8 &11  \\
\hline
${f_6}$  & 0.000 &11 & 3 &1 & 16 &12 & 14 &6 & 13 & 10 &20 &5 \\
\hline
${f_7}$  & 0.000& 9 &18 & 11 & 5 & 18 & 1 & 19 & 7 &2 &10 &6 \\
\hline
${f_8}$  & 0.459& 21 &7 &17& 10 &15 &19 &18 & 1 &14 & 4 &11\\
\hline
${f_9}$  & 0.337 & 1 & 6 & 8 & 5 & 15 & 17 & 2 & 9 & 9 &13 &10\\
\hline
\hline
Fun && $ E(0.5)$  & $E(0.1) $ & $E(0.8)$ & $Rayl(0.4)$ & $Rayl(0.8)$ & $Rayl(0.1)$ & $Weib(1,1.5)$ & $Weib(1.5,1)$ & $Weib(1,1)$ & $random$ & $LHS$ \\
\hline
${f_1}$ &  &3 &4 &	6 &	2 &	1 &	2 &	4 &	2 &	1 &	4 &	4 \\
\hline
${f_2}$ &  & 15 &6& 19 &10 & 7& 8& 3& 11 & 22 & 18 & 21\\
\hline
${f_3}$ &  & 12 & 17 &10 & 4 & 16& 6 & 13 & 2 & 5 & 3 & 14\\
\hline
${f_4}$ &   & 15 & 2 & 3 & 7 & 9 & 14 & 6 & 9 & 12 & 8 &16 \\
\hline
${f_5}$  &  & 19& 9& 5 & 2 & 16 & 3 & 1 &7 & 12 & 17 & 4 \\
\hline
${f_6}$  &  & 19 & 4 & 7 &2 & 9 & 5 & 8 & 18 & 15 & 17 &6 \\
\hline
${f_7}$  &  & 18 & 20 &13 & 4 & 15 &17 & 14 & 12 & 16& 3 & 8\\
\hline
${f_8}$  &  & 16 &8 &5 & 13 & 6 & 2 &22 & 9 &20& 3 &12  \\
\hline
${f_9}$  &  & 3 &12 & 18 &7 &14 &16 &11 &9 &14 &3 &4\\
\hline
\end{tabular}
\end{adjustwidth}
\end{table}

{{It can be seen clearly in Table~\ref{Table-10} that different initialization methods may have different effects on different functions.}} Apart from functions $f_2$, $f_8$ and $f_9$, all the $p$-values of other six functions are far less than 0.05, so the null hypothesis should be rejected at the $\alpha=0.05$ level. This means that the initialization will affect 2/3 of the functions, and only three functions ($f_2, f_8$ and $f_9$) are less influenced by initialization when the DE-a algorithm is used.

For the DE-a algorithm, it is easy to see in Table~\ref{Table-10} which initialization method is more suitable for certain functions. Now the question is that which initialization method(s) may be better for the DE-a algorithm? We use the above results in Table~\ref{Table-10} and treat the nine functions as observation samples. By comparing 22 different initialization methods using the Friedman rank test, the results are summarized in Table~\ref{Table-11}. The $p$-value is $0.617$, which is much greater than $0.05$, so we cannot reject the null hypothesis at the $\alpha=0.05$ level. This means that the performance of the DE-a is not particularly sensitive to the initialization on most test functions, which implies that DE is a relatively stable and robust algorithm.

\begin{table}[h]
\begin{adjustwidth}{-0.5cm}{}
\centering
\tiny
\caption{Friedman ranks of different initialization methods for DE-a. \label{Table-11} }
\begin{tabular}{|p{0.85cm}<{\centering}|p{1.1cm}<{\centering}|p{0.9cm}<{\centering}|p{0.9cm}<{\centering}|p{0.9cm}<{\centering}|p{0.9cm}<{\centering}|p{1.2cm}<{\centering}|p{1.2cm}<{\centering}|p{1.3cm}<{\centering}|p{1.1cm}<{\centering}|p{1.1cm}<{\centering}|}
\hline
  $ Be(3,2)$  & $Be(2.5,2.5)$ & $Be(2,3)$ & $U(0,1)$ & $N(0,1)$ & $N(0.5,1)$ & $N(0.5,0.5)$ & $logn(0,1)$ & $logn(.69,.25)$ & $logn(0,0.5)$ & $logn(0,2/3)$ \\
\hline
18& 3 &14 &19& 21 & 11 & 9 & 7 & 4 & 10 &5\\
\hline
\hline
 $E(0.5)$  & $E(0.1) $ & $E(0.8)$ & $Rayl(0.4)$ & $Rayl(0.8)$ & $Rayl(0.1)$ & $Weib(1,1.5)$ & $Weib(1.5,1)$ & $Weib(1,1)$ & $random$ & $LHS$ \\
\hline
22& 12 & 16 &1 & 15 & 2 & 13 & 6 & 20 & 8 & 17 \\
\hline
\end{tabular}
\end{adjustwidth}
\end{table}

Similar to the analysis for the DE-a algorithm, we now use Friedman rank tests for the PSO-w. The sorted  results of nine functions are shown in Table \ref{Table-12}. The $p$-values for functions $f_1$, $f_8$ and $f_9$, are 0.344, 0.459, 0.984, respectively, which means that we cannot reject the null hypothesis. This indicates that different initialization methods have no obvious effect on these three functions. However, for functions $f_3$, $f_4$, $f_6$ and $f_7$, when the initial population distribution obeys $Be(3,2)$, the accuracy of the solution is higher. In this case, its `Best', `Mean', `Var' and `Dist' values are all better than those by other distributions.

\begin{table}[h]
\begin{adjustwidth}{-1.3cm}{}
\centering
\tiny
\caption{Ranks of different initialization methods for PSO-w over functions $f_1-f_9$. \label{Table-12} }
\begin{tabular}{|p{0.2cm}|p{0.8cm}|p{0.75cm}<{\centering}|p{1.1cm}<{\centering}|p{0.8cm}<{\centering}|p{0.9cm}<{\centering}|p{0.9cm}<{\centering}|p{0.9cm}<{\centering}|p{1.2cm}<{\centering}|p{1.2cm}<{\centering}|p{1.3cm}<{\centering}|p{1.15cm}<{\centering}|p{1.2cm}<{\centering}|}
\hline
 Fun & $p{\rm{-}}value$ & $ Be(3,2)$  & $ Be(2.5,2.5)$ & $Be(2,3)$ & $U(0,1)$ & $N(0,1)$ & $N(0.5,1)$ & $N(0.5,0.5)$ & $logn(0,1)$ & $logn(.69,.25)$ & $logn(0,0.5)$ & $logn(0,2/3)$ \\
\hline
${f_1}$  & 0.344 &2&	11&	17&	13&	21&	19&	22&	16&	8&	4&	1\\
\hline
${f_2}$  & 0.02 &18&	17&	5&	4&	6&	8&	14&	16&	13&	9&	20	\\
\hline
${f_3}$    & 0.000 & 10&	1&	21&	4&	5&	15&	22&	18&	9&	7&	20\\
\hline
${f_4}$    & 0.000 & 18&	1&	19&	6&	8&	7&	12&	11&	4&	16&	15\\
\hline
${f_5}$    & 0.002 & 9	&8&	3&	17&	4&	10&	15&	16&	14&	5&	12 \\
\hline
${f_6}$  & 0.000 & 16	&1	&17&	5&	7&	11&	15&	13&	8&	18&	19\\
\hline
${f_7}$  & 0.000& 20	&1&	16&	4&	8&	3&	14&	11&	7&	18&	19\\
\hline
${f_8}$  & 0.459 & 2&	1&	3&	6&	19&	17&	9&	21&	22&	15&	18\\
\hline
${f_9}$  & 0.984 &3&	17&	20&	4&	18&	7	&21&	11&	8&	12&	13\\
\hline
\hline
Fun &   & $ E(0.5)$  & $E(0.1) $ & $E(0.8)$ & $Rayl(0.4)$ & $Rayl(0.8)$ & $Rayl(0.1)$ & $Weib(1,1.5)$ & $Weib(1.5,1)$ & $Weib(1,1)$ & $random$ & $LHS$ \\
\hline
 ${f_1}$  & & 20&	10&	5&	14&	3&	18&	12&	6&	15	&7&	9 \\
 \hline
${f_2}$  & & 19&	10&	7&	3&	12&	22&	1&	15&	21&	11&	2\\
\hline
${f_3}$ &  & 16	&8&	14&	2&	12&	11&	6&	17&	19&	13&	3\\
\hline
${f_4}$  &  & 14&	21&	5&	2&	17&	22&	20&	10&	13&	3&	9 \\
\hline
${f_5}$  &  & 1	&7&	6&	22&	19&	18&	20&	11&	13&	2&	21 \\
\hline
${f_6}$  & & 12	&21&	10&	2&	20&	22&	14&	6&	9&	3&	4 \\
\hline
${f_7}$  &  & 9	&21&	2&	5&	13&	22&	17&	10&	15&	12&	6\\
\hline
${f_8}$  &  & 8	&11&	14&	4&	12&	10&	13&	20&	16&	5&	7 \\
\hline
${f_9}$  &  & 22&	14&	15&	19&	5&	16&	6&	9&	10&	1&	2\\
\hline
\end{tabular}
\end{adjustwidth}
\end{table}

Considering all the benchmark functions, the overall ranks for the PSO-w algorithm are summarized in
Table~\ref{Table-13}. The null hypothesis is that the effects of initialization methods for the PSO-w are equally well. The $p$-value is $0.001$, which is less than $0.05$, so we can reject the null hypothesis at the $\alpha=0.05$ level. Thus, we can conclude that the performance of the PSO-w is sensitive to the initialization methods on test functions, and the three best initialization methods for the PSO-w algorithm are Random, $Be(2.5,2.5)$ and LHS, respectively.

\begin{table}[h]
\begin{adjustwidth}{-0.5cm}{}
\centering
\tiny
\caption{Friedman ranks of different initialization methods for PSO-w.  \label{Table-13} }
\begin{tabular}{|p{0.85cm}<{\centering}|p{1.1cm}<{\centering}|p{0.9cm}<{\centering}|p{0.9cm}<{\centering}|p{0.9cm}<{\centering}|p{0.9cm}<{\centering}|p{1.2cm}<{\centering}|p{1.2cm}<{\centering}|p{1.3cm}<{\centering}|p{1.1cm}<{\centering}|p{1.1cm}<{\centering}|}
\hline
  $ Be(3,2)$  & $ Be(2.5,2.5)$ & $Be(2,3)$ & $U(0,1)$ & $N(0,1)$ & $N(0.5,1)$ & $N(0.5,0.5)$ & $logn(0,1)$ & $logn(.69,.25)$ & $logn(0,0.5)$ & $logn(0,2/3)$ \\
\hline
9&	2&	15&	4&	8	&10	&21&	19&	7&	12&	20	\\
\hline
\hline
 $ E(0.5)$  & $E(0.1) $ & $E(0.8)$ & $Rayl(0.4)$ & $Rayl(0.8)$ & $Rayl(0.1)$ & $Weib(1,1.5)$ & $Weib(1.5,1)$ & $Weib(1,1)$ & $random$ & $LHS$ \\
\hline
16	&17&	6&	5&	14&	22&	13&	11&	18&	1&	3\\
\hline
\end{tabular}
\end{adjustwidth}
\end{table}

Similarly, we have carried out some analysis for the results by the CS. Table~\ref{Table-14} shows the rank results of different initialization methods for the CS over functions $f_1-f_9$. Apart from functions $f_8$ and $f_9$, the $p$-values of all other functions are far less than 0.05, which indicates that the CS algorithm is significantly affected by initialization. For most test functions, the most commonly used pseudo-random method is not the best initialization method. For example, the solution of $f_6$ under the beta distribution has the highest accuracy and stability.

\begin{table}[h]
\begin{adjustwidth}{-1.3cm}{}
\centering
\tiny
\caption{Ranks of different initialization methods for CS over functions $f_1-f_9$. \label{Table-14} }
\begin{tabular}{|p{0.2cm}|p{0.8cm}|p{0.75cm}<{\centering}|p{1.1cm}<{\centering}|p{0.8cm}<{\centering}|p{0.9cm}<{\centering}|p{0.9cm}<{\centering}|p{0.9cm}<{\centering}|p{1.2cm}<{\centering}|p{1.2cm}<{\centering}|p{1.3cm}<{\centering}|p{1.15cm}<{\centering}|p{1.2cm}<{\centering}|}
\hline
 Fun & $p{\rm{-}}value$  & $ Be(3,2)$  & $ Be(2.5,2.5)$ & $Be(2,3)$ & $U(0,1)$ & $N(0,1)$ & $N(0.5,1)$ & $N(0.5,0.5)$ & $logn(0,1)$ & $logn(.69,.25)$ & $logn(0,0.5)$ & $logn(0,2/3)$ \\
\hline
${f_1}$ & 0.000&8&	2&	9&	18&	14&	6&	7&	5&	22&	16&	4\\
\hline
${f_2}$  &0.000 &19&	3&	10&	7&	2&	6&	20&	9&	22&	1&	11\\
\hline
${f_3}$    & 0.000&16	&1	&5	&6&	14&	8&	9&	13&	22&	20&	11\\
\hline
${f_4}$    & 0.001 &7&	3&	10&	5&	16&	8&	6&	13&	22&	18&	21\\
\hline
${f_5}$   & 0.000 & 20&	19&	3&	10&	15&	9&	16&	13&	22&	11&	21 \\
\hline
${f_6}$ & 0.000&4&	1&	3&	5&	19&	20&	16&	15&	22&	11&	21\\
\hline
${f_7}$ & 0.000& 2&	1&	3&	6&	13&	12	&10&	17&	22&	14&	16 \\
\hline
${f_8}$ &0.709 & 2&	1&	8&	4&	11&	21&	13&	15&	14	&18	&19\\
  \hline
 ${f_9}$ & 0.196 &3&	8&	21&	6	&18&	19&	10&	11&	12&	2&	5\\
\hline
\hline
Fun &   & $ E(0.5)$  & $E(0.1) $ & $E(0.8)$ & $Rayl(0.4)$ & $Rayl(0.8)$ & $Rayl(0.1)$ & $Weib(1,1.5)$ & $Weib(1.5,1)$ & $Weib(1,1)$ & $random$ & $LHS$ \\
\hline
\multirow{1}{*}{${f_1}$} & & 3&	19&	20&	1&	12&	21&	11&	10&	13&	15&	17 \\
\hline
${f_2}$  & &12	&13&	14&	16&	4&	17&	18&	15&	8&	21&	5\\
\hline
${f_3}$  & & 12	&21&	4&	2&	18&	17&	7	&15&	19&	10&	3\\
\hline
${f_4}$  &  & 14&	17&	11&	4&	15&	19&	1&	20&	12&	2&	9\\
\hline
${f_5}$  &  & 5	&2	&4&	7&	14&	6&	8&	1&17&	18&	12 \\
\hline
${f_6}$  &  & 9&	18&	10&	2&	8	&17&	13&	12&	14&	6&	7 \\
\hline
${f_7}$ &  & 8	&11&	18&	4&	19&	9&	20&	21&	15&	5&	7\\
\hline
${f_8}$ &  & 16&	5&	22&	6& 12&	17&	7&	9&	20&	10&	3\\
\hline
${f_9}$ &  & 17	&22&	7&	15&	1&	20&	4&	13&	14&	9&	16\\
\hline
\end{tabular}
\end{adjustwidth}
\end{table}

In Table~\ref{Table-15}, the effect of different initialization methods on the CS is sorted. The null hypothesis is that the initialization methods for the CS are equally effective. The $p$-value is $0.00276$, which is far less that $0.05$, so we can confidently reject the null hypothesis at the $\alpha=0.05$ level. In other words, the selection of initial population has a great influence on the CS algorithm. It seems that the best initialization method is the Beta distribution, followed by Rayleigh and Uniform distributions.

\begin{table}[h]
\begin{adjustwidth}{-0.5cm}{}
\centering
\tiny
\caption{Friedman ranks of different initialization methods on the CS.  \label{Table-15} }
\begin{tabular}{|p{0.85cm}<{\centering}|p{1.1cm}<{\centering}|p{0.9cm}<{\centering}|p{0.9cm}<{\centering}|p{0.9cm}<{\centering}|p{0.9cm}<{\centering}|p{1.2cm}<{\centering}|p{1.2cm}<{\centering}|p{1.3cm}<{\centering}|p{1.1cm}<{\centering}|p{1.1cm}<{\centering}|}
\hline
  $ Be(3,2)$  & $ Be(2.5,2.5)$ & $Be(2,3)$ & $U(0,1)$ & $N(0,1)$ & $N(0.5,1)$ & $N(0.5,0.5)$ & $logn(0,1)$ & $logn(.69,.25)$ & $logn(0,0.5)$& $logn(0,2/3)$ \\
\hline
6	&1	&4	&3	&17&	11&	13&	15&	22&	16&	20\\
\hline
\hline
 $ E(0.5)$  & $E(0.1) $ & $E(0.8)$ & $Rayl(0.4)$ & $Rayl(0.8)$ & $Rayl(0.1)$ & $Weib(1,1.5)$ & $Weib(1.5,1)$ & $Weib(1,1)$ & $random$ & $LHS$\\
\hline
 9	&18	&12&	2	&10	&21&	7	&14&	19&	8&	5\\
\hline
\end{tabular}
\end{adjustwidth}
\end{table}


\subsection{Test problems from CEC2014 and CEC2017 suites}

In this section, some well-known single objective real-parameter numerical optimization problems from the  CEC2014 and CEC2017 benchmark suites have also been tested. The CEC2014 test suite~\citep{liang2013problem} contains 30 test problems where $f_1$ to $f_3$ are unimodal functions, $f_4$ to $f_6$ are multi-modal, $f_7$ to $f_{22}$ are hybrid, and $f_{23}$ to $f_{30}$ are compositions. The CEC2017 suite contains mainly single objective real parameter bound-constrained numerical optimization benchmark problems~\citep{cec2017}. The 30 benchmark functions are divided into four categories: unimodal functions ($F_1$-$F_3$), multimodal functions ($F_4$-$F_{10}$), hybrid functions ($F_{11}$-$F_{20}$) and composition functions ($F_{21}$-$F_{30}$).
All these problems are the minimization of the objective function with in the regular domain of ${[ - 100,100]^D}$ where $D$ is the dimension of the problem.
Obviously, it is time-consuming to test all these functions. In order to focus on the main objectives
of this paper, we have selected some representative functions with different properties. These 10 selected
benchmarks are seven from CEC2014 functions and three from the CEC2017 suite (see Table~\ref{table-20}).

\begin{table}[h]
\centering
\scriptsize
\caption{CEC2014 and CEC2017 single objective optimization problems. \label{table-20}}
\begin{tabular}{|l|l|l|l|}
\hline
Type  &No. & Function  &   Opt \\
\hline
Unimodal & $f_1$   & Rotated High Conditioned Elliptic Function   & 100\\
\hline
\multirow{2}{*}{Multimodal}&  $ f_7$ &  Shifted and Rotated Griewank¡¯s Function    & 700 \\
                            &  $ f_{13}$ &  Shifted and Rotated HappyCat Function    &1300\\
\hline
\multirow{2}{*}{Hybrid}&   $ f_{18}$ &  Hybrid Function 2 (N=3)    & 1800 \\
                            & $ f_{20}$  &  Hybrid Function 4 (N=4)    & 2000  \\
\hline
\multirow{2}{*}{Composition}&  $ f_{23}$  &  Composition Function 1 (N=5)   & 2300 \\
                            &  $ f_{25}$ &  Composition Function 3 (N=3)  &  2500    \\
\hline
\hline
Multimodal&   $ F_8$ & Shifted and Rotated Non-Continuous Rastrigin¡¯s function      & 800 \\
\hline
Hybrid&    $ F_{14}$    &  Hybrid Function 4 (N=4)   & 1400 \\
\hline
Composition  &  $ F_{22}$ &  Composition Function 2 (N=3)  &  2300    \\
\hline
\end{tabular}
\end{table}

Each algorithm has been run independently 20 times for each initialization method, and the stopping criterion is the same for all runs with 600000 fitness evaluations. The population size $NP$ and the maximum number of iterations $T$ of DE-a are set to 300 and 2000, respectively. For the PSO-w, the $NP$ is 2000 and the $T$ is 300. The $NP$ is set to 100 and the $T$ is set to 6000 for CS. The parameters involved in three algorithms are the same as the previous settings. There is no explicit reference about the optimal solutions of CEC2014 and CEC217 suites, so the indicator `Dist' is not presented in this part. The experimental results of these 30 dimensions are shown in Tables~\ref{Table-21} to \ref{Table-23}:

\begin{table}[h]
\begin{adjustwidth}{-1.7cm}{}
\centering
\tiny
\caption{Comparison of DE-a for CEC functions with different initialization methods. \label{Table-21} }
\begin{tabular}{|p{0.2cm}|p{0.45cm}|p{1.06cm}<{\centering}|p{1.08cm}<{\centering}|p{1.06cm}<{\centering}|p{1.06cm}<{\centering}|p{1.06cm}<{\centering}|p{1.06cm}<{\centering}|p{1.2cm}<{\centering}|p{1.2cm}<{\centering}|p{1.3cm}<{\centering}|p{1.09cm}<{\centering}|p{1.12cm}<{\centering}|}
\hline
 Fun & Value  & $ Be(3,2)$  & $ Be(2.5,2.5)$ & $Be(2,3)$ & $U(0,1)$ & $N(0,1)$ & $N(0.5,1)$ & $N(0.5,0.5)$ & $logn(0,1)$ & $logn(.69,.25)$ & $logn(0,0.5)$ & $logn(0,2/3)$ \\
\hline
 \multirow{3}{*}{${f_1}$} &Best & 1.84e+06&	1.25e+06&	1.59e+06&	2.11e+06&	1.50e+06&	2.32e-06&	 2.10e+06&	2.15e+06&	2.00e+06&	1.86e+06&	1.95e+06\\
                             &Mean  &3.19e+06&	2.74e+06&	3.56e+06&	3.52e+06&	3.43e+06&	 3.55e+06&	3.80e+06&	3.74e+06&	3.79e+06&	3.31e+06&	3.58e+06 \\
                             &Var   &1.47e+12&	1.145e+12&	1.46e+12&	1.65e+12&	1.38e+12&	 1.11e+12&	1.42e+12&	1.21e+12&	2.15e+12&	1.28eE+12&	1.86e+12 \\
\hline
\multirow{3}{*}{${f_7}$}  &Best & 700&	700&	700&	700&	700&	700&	700&	700&	700&	 700&	700\\
                             &Mean &700.01&	700.00&	700.00&	700.01&	700.00&	700.01&	700.00&	700.00&	 700.00&	700.00&	700.01\\
                             &Var  & 5.60e-05&	2.52e-05&	5.10e-05&	6.36e-05&	3.61e-05&	 3.38e-05&	3.16e-05&	5.12e-05&	3.27e-05&	3.59e-05&	7.44e-05 \\
\hline
\multirow{3}{*}{${f_{13}}$}  & Best &1300.19&	1300.20&	1300.21& 1300.20& 1300.22& 1300.19& 1300.20& 1300.21& 1300.15&	1300.15& 1300.19\\
                             &Mean  &1300.26&	1300.27&	1300.27&	1300.27&	1300.27&	 1300.256&	1300.27&	1300.26&	1300.26&	1300.26&	1300.28\\
                             &Var   &1.81e-03&	1.35e-03&	1.66e-03&	1.53e-03&	1.53e-03&	 1.18e-03&	1.51e-03&	9.34e-04&	2.14e-03&	2.04e-03&	1.52e-03\\
\hline
\multirow{3}{*}{${f_{18}}$}    &Best &1866.1&	1888.3&	1867.5&	1870.2&	1864&	1883.6&	1885.5&	1881.5&	 1836.6&	1894.1&	1877  \\
                             &Mean &2105.6&	1984&	2023.5&	1966.4&	2071.5&	2199.6&	2012.3&	14458&	 1958.6&	2202.2&	1974.8 \\
                             &Var  &1.60e+05&	13063&	51432&	10016&	96318&	4.55e+05&	15738&	 3.10e+09&	3500.4&	6.84e+05&	7250.3\\
\hline
\multirow{3}{*}{${f_{20}}$}    &Best &2037.16&	2038.13&	2033.42&	2034.92&	2042.43&	 2036.96&	2043.04&	2036.83&	2037.90&	2037.88&	2039.17\\
                             &Mean  &2048.25&	2048.19&	2049.18&	2049.54&	2050.26&	 2050.48&	2051.85&	2048.66&	2050.31&	2048.58&	2049.95\\
                             &Var   & 36.18&	52.19&	51.87&	41.27&	36.47&	34.04&	17.77&	31.20&	 33.69&	28.20&	25.30\\
\hline
\multirow{3}{*}{${f_{23}}$ } &Best &2615.24& 2615.24&	2615.24	&2615.24&	2615.24	&2615.24&	2615.24	 &2615.24&	2615.24&	2615.24&	2615.24\\
                             &Mean &2615.24&	2615.24&	2615.24&	2615.24&	2615.24&	 2615.24&	2615.24&	2615.24&	2615.24&	2615.24&	2615.24 \\
                             &Var   &2.18e-25&	2.18e-25&	2.18e-25&	2.18e-25&	2.18-25&	 2.18e-25&	2.18e-25&	2.18e-25&	2.18-25&	2.18-25&	2.18e-25\\
\hline
 \multirow{3}{*}{${f_{25}}$ } &Best &2703.26&	2702.77&	2702.78&	2702.80	&2702.82&	2702.97	 &2703.03&	2702.75&	2702.92&	2703.20&	2702.91\\
                             &Mean  &2704.02&	2703.80	&2703.99&	2704.12	&2703.62&	2703.95&	 2704.11&	2703.92&	2704.12&	2704.18&	2703.76 \\
                             &Var   &3.39e-01&	2.11e-01&	7.17e-01&	3.74e-01&	1.93e-01&	 4.05e-01&	4.08e-01&	4.89e-01&	6.52e-01&	3.33e-01	&2.94e-01\\
\hline
\multirow{3}{*}{${F_{8}}$}    &Best &942.05&	945.78&	947.19&	957.68&	944.1&	948.11&	939.79&	939.35&	 918.84&	919.51&	939.27\\
                             &Mean  &961.36&	961.92&	964.58&	967.55&	964.98&	964.55&	964.48&	963.77&	 963.84&	962.48&	962.33\\
                             &Var   &100.89&	59.664&	79.826&	73.786&	167.79&	54.317&	154.33&	165.39&	 223.97&	214.74&	113.96\\
\hline
\multirow{3}{*}{${F_{14}}$ } &Best &1457&  1463.3&	1466.5&	1466.1&	1458.4&	1467.1&	1466.6&	1460.5&	 1466.6&	1457.8&	1463\\
                             &Mean &1473.8&	1474.8&	1474.3&	1476.8&	1475.4&	1478.2&	1477.7&	1474.9&	 1476.2&	1474.8&	1478.8 \\
                             &Var   &52.13&	37.66&	19.04&	37.37&	36.14&	33.11&	33.88&	51.01&	 30.92&	66.27&	64.09\\
\hline
 \multirow{3}{*}{${F_{22}}$ } &Best &2300&	2300&	2300&	2300&	2300&	2300&	2300&	2300&	 2300&	2300&	2300\\
                             &Mean  &2300.2&	2300.1&	2300.1&	2300.1&	2300.1&	2300.1&	2300&	2300&	 2300.1&	2300&	2300.1 \\
                             &Var   &5.79e-01&	3.01e-01&	3.01e-01&	3.13e-01&	3.01e-01&	 3.01e-01&	1.41e-25&	1.52e-25&	3.03e-01&	1.74e-25&	3.01e-01\\
\hline
\hline
Fun &   & $ E(0.5)$  & $E(0.1) $ & $E(0.8)$ & $Rayl(0.4)$ & $Rayl(0.8)$ & $Rayl(0.1)$ & $Weib(1,1.5)$ & $Weib(1.5,1)$ & $Weib(1,1)$ & $random$ & $LHS$ \\
\hline
 \multirow{3}{*}{${f_1}$ } &Best &1.78e+06&	2.01e+06&	1.87e+06&	1.62e+06&	2.11e+06&	1.19e+06&	 2.08e+06&	1.78e+06&	1.98e+06&	1.44e+06&	1.25e+06 \\
                             &Mean &3.94e+06&	3.62e+06&	3.22e+06&	3.41e+06&	3.46e+06&	 3.44e+06&	3.42e+06&	3.26e+06&	3.59e+06&	3.58e+06&	3.20e+06 \\
                             &Var  &2.64e+12&	8.27e+11&	1.23e+12&	1.56e+12&	8.22e+11&	 2.11e+12&	1.17e+12&	1.23e+12&	1.62e+12&	2.35e+12&	1.39e+12  \\
 \hline
 \multirow{3}{*}{${f_7}$}   &Best & 700&	700&	700&	700&	700&	700&	700&	700	&700&	 700&	700\\
                             &Mean &700.00&	700.01&	700.01&	700.01&	700.00&	700.01&	700.01&	700.01&	 700.00&	700.01&	700.00\\
                             &Var & 3.85e-05&	1.39e-04&	8.51e-05&	1.17e-04&	4.02e-05&	 7.73e-05&	5.43e-05&	7.84e-05&	4.18e-05&	4.46e-05&	5.00e-05 \\
\hline
\multirow{3}{*}{${f_{13}}$}   &Best &1300.17&	1300.18	&1300.20& 1300.19&	1300.16&	1300.16&	 1300.18&	1300.17&	1300.20&	1300.20	& 1300.20\\
                             &Mean  & 1300.27&	1300.25&	1300.27&	1300.26&	1300.26	&1300.26&	 1300.25&	1300.25	&1300.26&	1300.26	& 1300.28\\
                             &Var   &1.92e-03&	1.85e-03&	1.34e-03&	1.28e-03&	1.97e-03&	 2.15e-03&	2.66e-03&	2.07e-03&	1.34e-03&	1.58e-03&	2.05e-03 \\
\hline
\multirow{3}{*}{${f_{18}}$}    &Best &1855.9&	1854.5&	1867.8&	1857.5&	1872.6&	1850&	1887.2&	1881.2&	 1835&	1851.4&	1856.6 \\
                             &Mean &2054.1&	1936.1&	1993.3&	1958.8	&2031.2&	1962.6&	1970.6	&2043.3	 &1948.6	&1958&	3449 \\
                             &Var  &68834&	6992.8&	28191&	3191.8&	69288&	21981&	5744.2&	97484	 &4970.3&	5558.5&	4.29E+07\\
\hline
\multirow{3}{*}{${f_{20}}$}    &Best &2035.72&	2044.01&	 2029.32&2034.46&2036.93&2033.82&2034.80&2041.47&	2040.84	&2035.73&	2041.37\\
                             &Mean  &2049.25&	2052.33&	2048.86&	2048.19&	2051.34&	 2049.18&	2048.36&	2048.28&	2051.39&	2049.50&	2049.10 \\
                             &Var   &31.46&	11.49&  49.09&	62.84&	37.90&	50.68&	37.71&	16.13&	 38.44&	45.16&	44.02\\
\hline
\multirow{3}{*}{${f_{23}}$ } &Best &2615.24& 2615.24&	2615.24	&2615.24&	2615.24	&2615.24&	2615.24	 &2615.24&	2615.24&	2615.24&	2615.24\\
                             &Mean &2615.24&	2615.24&	2615.24&	2615.24&	2615.24&	 2615.24&	2615.24&	2615.24&	2615.24&	2615.24&	2615.24 \\
                             &Var   &2.18e-25&	2.18e-25&	2.18e-25&	2.18e-25&	2.18-25&	 2.18e-25&	2.18e-25&	2.18e-25&	2.18-25&	2.18-25&	2.18e-25 \\
\hline
 \multirow{3}{*}{${f_{25}}$ } &Best & 2702.83&	2702.83&	2703.16&	2702.92&	2703.19&	 2702.61&	2703.25&	2702.83&	2703.03&	2703.13&	2703.13\\
                             &Mean  &2704.11&	2703.92&	2704.15&	2703.79&	2703.98&	 2703.34&	2704.06&	2703.77&	2703.93&	2703.89&	2703.92\\
                             &Var   &5.91e-01&	5.75e-01&	4.34e-01&	3.44e-01&	3.78e-01&	 3.65e-01&	2.39e-01&	3.29e-01&	2.56e-01&	2.14e-01&	3.26e-01\\
\hline
\multirow{3}{*}{${F_{8}}$}    &Best &944.34&	941.11&	949.11&	932.02&	948.63&	943.51&	936.27&	940.6&	 945.19&	941.23&	948.74\\
                             &Mean  &963.73&	961.87&	963.83&	960.52&	965.03& 962.46&	960.09&	965.05&	 966.09&	963.98&	964.64 \\
                             &Var   &86.325&	120.04&	63.67&	135.94&	77.99&	129.63&	154.77&	140.02&	 80.76&	74.60&	103.69\\
\hline
\multirow{3}{*}{${F_{14}}$ } &Best &1466.6&	1466&	1462.9&	1462.7&	1464.6&	1453.3&	1461.6&	1462.9&	 1463.5&	1460.6&	1456.4\\
                             &Mean &1474.2&	1475.4&	1474.4&	1476.6&	1475.9&	1475.7&	1473.9&	1474.9&	 1475&	1474.1&	1475.6 \\
                             &Var   &21.49&	28.57&	26.34&	37.29&	32.21&	70.68&	51.38&	48.05&	 36.61&	42.83&	57.64 \\
\hline
 \multirow{3}{*}{${F_{22}}$ } &Best &2300&	2300&	2300&	2300&	2300&	2300&	2300&	2300&	 2300&	2300&	2300 \\
                             &Mean  &2300.2 &	2300&	2300.2&	2300&	2300.2&	2639.6&	2300&	2300&	 2300&	2300.1&	2300.1\\
                             &Var   &5.70e-01&	1.52e-25&	5.70e-01&	1.96e-25&	5.70e-01&	 2.30e+06&	1.85e-25&	1.31e-25&	1.85e-25	& 3.01e-01&	3.03e-01\\
\hline
\end{tabular}
\end{adjustwidth}
\end{table}

\begin{table}[h]
\begin{adjustwidth}{-1.7cm}{}
\centering
\tiny
\caption{Comparison of PSO-w for CEC functions with different initialization methods. \label{Table-22} }
\begin{tabular}{|p{0.2cm}|p{0.45cm}|p{1.06cm}<{\centering}|p{1.08cm}<{\centering}|p{1.06cm}<{\centering}|p{1.06cm}<{\centering}|p{1.06cm}<{\centering}|p{1.06cm}<{\centering}|p{1.2cm}<{\centering}|p{1.2cm}<{\centering}|p{1.3cm}<{\centering}|p{1.09cm}<{\centering}|p{1.12cm}<{\centering}|}
\hline
 Fun & Value  & $ Be(3,2)$  & $ Be(2.5,2.5)$ & $Be(2,3)$ & $U(0,1)$ & $N(0,1)$ & $N(0.5,1)$ & $N(0.5,0.5)$ & $logn(0,1)$ & $logn(.69,.25)$ & $logn(0,0.5)$ & $logn(0,2/3)$ \\
\hline
 \multirow{3}{*}{${f_1}$} &Best &3.09e+05&	2.08e+05&	45809&	3.20e+05&	2.05e+05&	2.81e+05&	 1.95e+05&	1.70e+05&	1.93e+05&	2.46e+05&	2.18e+05\\
                             &Mean  &5.11e+06&	3.17e+06&	1.73e+06	&2.53e+06&	2.41e+06&	 2.50e+06&	4.62e+06&	1.73e+06&	2.81e+06&	1.63e+06&	2.62e+06 \\
                             &Var   &3.43e+13&	1.11e+13&	2.81e+12&	3.49e+12&	4.12e+12&	 8.05e+12&	2.83e+13&	2.35e+12&	2.69e+13&	1.89e+12&	6.83e+12 \\
\hline
\multirow{3}{*}{${f_7}$}  &Best & 700&	700&	700&	700&	700&	700&	700&	700&	700&	 700&	700\\
                             &Mean &700.02&	700.02&	700.01&	700.01&	700.02&	700.02&	700.01&	700.02	 &700.02&	700.01&	700.02\\
                             &Var  & 3.69e-04&	2.34e-04& 2.64e-04&	1.34e-04& 1.64e-04&	3.55e-04&	 1.54e-04& 2.15e-04	&2.44e-04	&1.45e-04&	2.69e-04 \\
\hline
\multirow{3}{*}{${f_{13}}$}  & Best &1300.2	&1300.2&	1300.2&	1300.3&	1300.3&	1300.3&	1300.2&	1300.3&	 1300.2&	1300.2&	1300.2\\
                             &Mean  &1300.4 &	1300.4&	1300.4&	1300.4&	1300.4&	1300.4&	1300.4&	1300.4&	 1300.4&	1300.4&	1300.4\\
                             &Var   &7.83e-03&	1.14e-02& 8.75e-03&	6.54e-03& 1.12e-02&	1.12e-02& 1.59e-02&	7.42e-03&	8.69e-03& 7.69e-03&	9.06e-03 \\
\hline
\multirow{3}{*}{${f_{18}}$}    &Best &1866.1&	1888.3&	1867.5&	1870.2&	1864&	1883.6&	1885.5&	1881.5&	 1836.6&	1894.1&	1877  \\
                             &Mean &2105.6&	1984&	2023.5&	1966.4&	2071.5&	2199.6&	2012.3&	14458&	 1958.6&	2202.2&	1974.8 \\
                             &Var  &1.60e+05&	13063&	51432&	10016&	96318&	4.55e+05&	15738&	 3.10e+09&	3500.4&	6.84e+05&	7250.3\\
\hline
\multirow{3}{*}{${f_{20}}$}    &Best &2050.3&	2076&	2066.1&	2093&	2090&	2119.3&	2084.9&	2108.9&	 2103.6&	2107.4&	2068.1\\
                             &Mean  &2155.6&	4383.9&	2170.6&	2199.4&	2212.9&	2180.2&	2176.3&	2174.8&	 2188.7&	2184.7&	2175.6\\
                             &Var   &2155.7&	9.94e+07&	3548&	5516.9&	6635.3&	2236.4&	3755.2&	 2517.2&	2758.6&	5166.8&	1918.3\\
\hline
\multirow{3}{*}{${f_{23}}$ } &Best &2500&	2500&	2500&	2500&	2615.4&	2500&	2500&	2615.4&	 2500&	2500&	2615.4\\
                             &Mean  &2500&	2500&	2500&	2500&	2615.7&	2610&	2528.9&	2615.7&	 2586.9&	2592.7&	2615.8\\
                             &Var   &3.09E-09&	2.91E-09&	3.21E-09&	3.70E-09& 2.19e-02&	670.33&	 2639.8&	5.182e-02&	2646.8&	2261.1&	1.31e-01\\
\hline
 \multirow{3}{*}{${f_{25}}$ } &Best &2700&	2700&	2700&	2700&	2700&	2700&	2700&	2700&	 2700&	2700&	2700\\
                             &Mean  &2700&	2700&	2700&	2700&	2707.4&	2704.4&	2710.1&	2702.6&	 2702.2&	2702.7 \\
                             &Var   &4.21e-12&	1.96e-12&	3.68e-13&	1.17e-11&	18.138&	20.167&	 3.82e-08&	15.2&	14.12&	9.8308&	15.881 \\
\hline
 \multirow{3}{*}{${F_{8}}$}    &Best &888.55&	900.49&	872.63&	866.66&	888.55&	887.56&	885.57&	898.50&	 883.58&	906.47&	900.49\\
                             &Mean  &933.62&	933.48&	928.55&	932.13&	934.52&	935.56&	929.19&	931.33&	 935.07&	951.04&	941.56\\
                             &Var   &513.41&	570.29&	745.33&	773.66&	552.95&	576.55&	595.09&	483.40&	 578.83&	589.91&	450.36\\
\hline
 \multirow{3}{*}{${F_{14}}$ } &Best &1.47e+03&	1.47e+03&	1.48e+03&	1.46e+03&	1.48e+03&	 1.48e+03&	1.48e+03&	1.49e+03&	1.48e+03&	1.46e+03&	1.50e+03\\
                             &Mean  &1.96e+03&	2.01e+03&	1.55e+03&	1.53e+03&	2.44e+03&	 2.07e+03&	1.55e+03&	3.14e+03&	1.54e+03&	1.54e+03&	1.58e+03\\
                             &Var   &3.68e+06&	4.59e+06&	3.89e+03&	1.30e+03&	1.59e+07&	 5.46e+06&	2.75e+03&	2.05e+07&	2.08e+03&	1.99e+03&	7.95e+03\\
\hline
 \multirow{3}{*}{${F_{22}}$ } &Best &2.30e+03&	2.30e+03&	2.30e+03&	2.30e+03&	2.30e+03&	 2.30e+03&	2.30e+03&	2.30e+03&	2.30e+03&	2.30e+03&	2.30e+03\\
                             &Mean  &2.30e+03&	2.30e+03&	2.30e+03&	2.47e+03&	2.30e+03&	 2.30e+03&	2.30e+03&	2.30e+03&	2.61e+03&	2.30e+03	&2.30e+03 \\
                             &Var   &2.01&	2.40&	1.97&	5.85e+05&	2.57&	0.82&	0.80&	2.15&	 9.85e+05&	2.96&	2.13 \\
\hline
\hline
Fun &   & $ E(0.5)$  & $E(0.1) $ & $E(0.8)$ & $Rayl(0.4)$ & $Rayl(0.8)$ & $Rayl(0.1)$ & $Weib(1,1.5)$ & $Weib(1.5,1)$ & $Weib(1,1)$ & $random$ & $LHS$ \\
\hline
 \multirow{3}{*}{${f_1}$ } &Best &2.88e+05&	2.33e+05&	2.04e+05&56405&	3.23e+05&	1.93e+05&	 1.86e+05&	95827&	1.64e+05&	74981&	2.49e+05 \\
                             &Mean &2.96e+06&	1.53e+06&	2.52e+06&	3.51e+06&	2.22e+06&	 2.29e+06&	2.65e+06&	1.86e+06&	1.98e+06&	3.25e+06&	3.41e+06 \\
                             &Var  &1.11e+13&	2.39e+12&	3.64e+12&	1.11e+13&	2.87e+12&	 3.99e+12&	9.50e+12&	3.63e+12&	3.84e+12&	1.71e+13&	2.23e+13  \\
 \hline
 \multirow{3}{*}{${f_7}$}   &Best & 700&	700&	700&	700&	700&	700&	700&	700	&700&	 700&	700\\
                             &Mean &700.02&	700.01&	700.01&	700.01&	700.02&	700.01&	700.01&	700.02&	 700.01&	700.01&	700.01\\
                             &Var & 3.36e-04& 2.06e-04&	1.75e-04& 1.93e-04&	4.02e-04&	7.84e-05&	 8.82e-05& 3.44e-04&	1.71e-04& 2.37e-04&	1.03e-04 \\
\hline
\multirow{3}{*}{${f_{13}}$}   &Best & 1300.2&	1300.2&	1300.2&	1300.3&	1300.2&	1300.2&	1300.2&	1300.2&	 1300.3&	1300.3&	1300.2\\
                             &Mean  & 1300.4&	1300.3&	1300.4&	1300.4&	1300.4&	1300.4&	1300.4&	1300.4&	 1300.4&	1300.4&	1300.4\\
                             &Var   &8.16e-03&	8.75e-03& 1.01e-02&	7.84e-03& 1.17e-02&	1.15e-02& 8.54e-03&	8.97e-03& 9.26e-03&	5.34e-03&1.12e-02 \\
\hline
\multirow{3}{*}{${f_{18}}$}    &Best &1855.9&	1854.5&	1867.8&	1857.5&	1872.6&	1850&	1887.2&	1881.2&	 1835&	1851.4&	1856.6 \\
                             &Mean &2054.1&	1936.1&	1993.3&	1958.8	&2031.2&	1962.6&	1970.6	&2043.3	 &1948.6	&1958&	3449 \\
                             &Var  &68834&	6992.8&	28191&	3191.8&	69288&	21981&	5744.2&	97484	 &4970.3&	5558.5&	4.29E+07\\
\hline
\multirow{3}{*}{${f_{20}}$}    &Best &2094.5&	2081.9&	2078&	2068.4&	2080&	2091&	2089.2&	2110.1&	 2062.6&	2070.5&	2065.2\\
                             &Mean  &2182.1&	2200.1&	2194.8&	4627.2&	2174.5&	2193.8&	2169.4&	2174&	 2153.8&	2164.9&	2192.8 \\
                             &Var   &3888.9&	4664&	2587.6&	1.22e+08&	2677.6&	4364.7&	1978.1&	 4444.5&	2943.2&	3629.6&	3892.2\\
\hline
\multirow{3}{*}{${f_{23}}$ } &Best &2500&	2500&	2500&	2500&	2500&	2500&	2615.4&	2615.4&	 2500&	2500&	2500\\
                             &Mean &2500&	2500&	2586.8&	2500&	2598.4&	2500&	2615.7&	2615.7&	 2609.8&	2500&	2500 \\
                             &Var   &8.84e-09&	8.83e-10&	2643.6&	3.47e-09&	1798.8&	1.06e-09&	 3.9262e-02&	5.20e-02&	668.29&	2.72e-09&	7.06e-09 \\
\hline
 \multirow{3}{*}{${f_{25}}$ } &Best & 2700&	2700&	2700&	2700&	2700&	2700&	2700&	2700&	 2700&	2700&	2700\\
                             &Mean  &2700&	2700&	2702.2&	2700&	2700.7&	2700&	2702&	2710.6&	 2705.1&	2700&	2700\\
                             &Var   &6.44e-11&	3.05e-13&	11.943&	1.39e-12&	3.2384&	8.57e-13&	 10.414	&14.051&	7.8027&	4.27e-12	& 6.92e-12\\
\hline
\multirow{3}{*}{${F_{8}}$}    &Best &881.59&	908.45&	905.47&	892.53&	894.52&	920.53&	902.49&	898.06&	 881.59&	883.58&	900.49\\
                             &Mean  &928.45&	951.24&	938.39&	946.45&	939.50&	953.56&	939.54&	932.66&	 931.84&	931.64&	935.43\\
                             &Var   &482.66&	818.25&	473.99&	578.76&	846.19&	503.99 & 303.78&	 593.32&	761.96&	470.65&	453.12\\
\hline
\multirow{3}{*}{${F_{14}}$ } &Best &1.49e+03&	1.50e+03&	1.49e+03&	1.46e+03&	1.49e+03&	 1.48e+03&	1.49e+03&	1.48e+03&	1.49e+03&	1.46e+03&	1.48e+03\\
                             &Mean  &1.54e+03&	1.55e+03&	2.25e+03&	2.30e+03&	2.26e+03&	 2.84e+03&	4.14e+03&	2.12e+03&	3.25e+03& 1.54e+03&	1.55e+03\\
                             &Var   &1.22e+03&	1.49e+03&	9.58e+06&	1.15e+07&	1.05e+07&	 3.12e+07&	8.67e+07&	6.98e+06&	5.79e+07&	1.47e+03&	1.39e+03\\
\hline
 \multirow{3}{*}{${F_{22}}$ } &Best &2.30e+03&	2.30e+03&	2.30e+03&	2.30e+03&	2.30e+03&	 2.30e+03&	2.30e+03&	2.30e+03&	2.30e+03&	2.30e+03&	2.30e+03\\
                             &Mean  &2.30e+03&	4.25e+03&	2.30e+03&	2.30e+03&	2.30e+03&	 4.02e+03&	2.30e+03&	2.30e+03&	2.30e+03&	2.30e+03&	2.30e+03 \\
                             &Var   &	3.22 & 5.18e+06&	3.30&	13.27&	1.42&	5.92e+06&	4.31&	 2.12&	2.52&	1.48&	2.70 \\
\hline
\end{tabular}
\end{adjustwidth}
\end{table}

\begin{table}[h]
\begin{adjustwidth}{-1.7cm}{}
\centering
\tiny
\caption{Comparison of CS for CEC functions with different initialization methods. \label{Table-23} }
\begin{tabular}{|p{0.2cm}|p{0.45cm}|p{1.06cm}<{\centering}|p{1.08cm}<{\centering}|p{1.06cm}<{\centering}|p{1.06cm}<{\centering}|p{1.06cm}<{\centering}|p{1.06cm}<{\centering}|p{1.2cm}<{\centering}|p{1.2cm}<{\centering}|p{1.3cm}<{\centering}|p{1.09cm}<{\centering}|p{1.12cm}<{\centering}|}
\hline
 Fun & Value  & $ Be(3,2)$  & $ Be(2.5,2.5)$ & $Be(2,3)$ & $U(0,1)$ & $N(0,1)$ & $N(0.5,1)$ & $N(0.5,0.5)$ & $logn(0,1)$ & $logn(.69,.25)$ & $logn(0,0.5)$ & $logn(0,2/3)$ \\
\hline
 \multirow{3}{*}{${f_1}$} &Best &2.92e+05&	4.00e+05&	3.46e+05&	4.97e+05&	3.72e+05&	4.81e+05&	 5.12e+05&	3.72e+05&	7.01e+08&	3.82e+05&3.54e+05	\\
                             &Mean  &6.46e+05&	7.53e+05&	7.45e+05&	7.67e+05&	7.16e+05&	 7.10e+05&	7.40e+05&	6.48e+05&	6.57e+09&	5.34e+05	&6.32e+05 \\
                             &Var   &3.21e+10&	2.56e+10&	2.24e+10&	2.82e+10&	3.33e+10&	 2.79e+10&	2.51e+10&	2.15e+10&	3.21e+19&	6.96e+09&	1.48e+10 \\
\hline
\multirow{3}{*}{${f_7}$}  &Best & 700&	700&	700&	700&	700&	700&	700&	700&	1965.7&	 700&	700\\
                             &Mean &700&	700&	700&	700&	700&	700&	700&	700&	 3159.2&	700&	700\\
                             &Var  & 6.65e-25&	2.77e-24&	1.75e-24&	3.70e-22&	5.03e-23&	 2.97e-22&	6.64e-23&	1.56e-24&	2.63e+05	& 1.34e-22&	5.55e-21 \\
\hline
\multirow{3}{*}{${f_{13}}$}  & Best &1300.2&	1300.3&	1300.2&	1300.2&	1300.2&	1300.2&	1300.2&	1300.2&	 1317.7&	1300.2&	1300.2\\
	
                             &Mean  &1300.3&	1300.3&	1300.3&	1300.3&	1300.3&	1300.3&	1300.3&	1300.3&	 1328.9&	1300.3	&1300.3\\
                             &Var   &1.39e-03&	8.42e-04&	1.84e-03&	1.42e-03& 1.12e-03&	1.56e-03&	 1.72e-03& 7.30e-04&	28.63&	1.41e-03	& 9.28e-04\\
\hline
\multirow{3}{*}{${f_{18}}$}    &Best &1843.3 &	1830.3&	1842.2&	1834.6&	1838.3&	1832.9&	1827.6&	1833.4&	 2.91e+10&	1830&	1833.6  \\
                             &Mean &1859.6&	1850.9&	1856.4&	1851.4&	1856.4&	1853.5&	1853.4&	1854.7&	 7.62e+10&	1850.3&	1854.9 \\
                             &Var  &145.62&	143.06&	136.53&	89.423&	108.16&	169.72&	93.004&	120.63&	 4.19e+20&	104.59&	94.003 \\
\hline
\multirow{3}{*}{${f_{20}}$}    &Best & 2033.5&	2030.3&	2036.6&	2026&	2032.7&	2032.7&	2034.2&	2037.4&	 1.25e+05&	2036.5&	2033.7\\
                             &Mean  &2043.8	&2042.7&	2046.1&	2043.1&	2041.7&	2042.8&	2043.9&	2045.1&	 1.54e+09&	2043.8&	2044.9\\
                             &Var   &22.967&	34.75&	46.147&	52.618&	17.839&	32.583&	39.687&	36.689&	 2.96e+18&	33.398&	30.237 \\
\hline
\multirow{3}{*}{${f_{23}}$ } &Best &2615.2&	2615.2&	2615.2&	2615.2&	2615.2&	2615.2&	2615.2&	2615.2&	 12419&	2615.2&	2615.2\\
                             &Mean  &2615.2&	2615.2&	2615.2&	2615.2&	2615.2&	2615.2&	2615.2&	2615.2&	 26820&	2615.2&	2615.2\\
                             &Var   &2.18e-25&	2.18e-25&	2.18e-25&	2.18e-25&	2.18e-25&	 2.18e-25&	2.18e-25&	2.18e-25&	5.50e+07	& 2.18e-25&	2.18e-25 \\
\hline
 \multirow{3}{*}{${f_{25}}$ } &Best &2703.9&	2703.6&	2703.6&	2703.8&	2704.1&	2703.8&	2703.6&	2703.7&	 3171.9&	2703.9&	2704\\
                             &Mean  &2704.8	&2704.4&	2704.6&	2704.6&	2705.1&	2704.8&	2704.9&	2705.1&	 4143&	2705&	2705.1 \\
                             &Var   &0.28&	0.25&	0.46&	0.35&	0.28&	0.23&	0.45&	0.76&	 2.16e+05&	0.39&	0.38 \\
\hline
\multirow{3}{*}{${F_{8}}$}    &Best & 924.24&	913.1&	905.61&	874.39&	887.55&	883.15&	891.24&	904.51&	 1593.5&	865.49&	888.58\\
                             &Mean  &947.46&	937.15&	938.27&	919.17&	914.73&	912.46&	922.11&	928.04&	 1943.7&	916.34&	918.49\\
                             &Var   &287.24&	272.62&	251.48&	341.25&	213.17&	108.03&	255.46&	287.42&	 34289&	354.4&	311.77 \\
\hline
\multirow{3}{*}{${F_{14}}$ } &Best &1446.1&	1451.1&	1442.6&	1447.1&	1446.5&	1442.6&	1447&	1446.2&	 1453&	1446.9&	1446.7\\
                             &Mean  &1455.5&	1459.5&	1455.9&	1456.7&	1459.7&	1456.5&	1455.9&	1457.2&	 8.427e+09&	1456.2&	1457\\
                             &Var   &44.31&	34.55&	62.52&	34.97&	56.84&	64.83&	29.23&	39.89&	 1.07e+20&	38.85&	55.49\\
\hline
 \multirow{3}{*}{${F_{22}}$ } &Best &2300&	2300&	2300&	2300&	5509.9&	2396&	2300.1&	2301.5&	 4151&	2300.1&	2301.1\\
                             &Mean  &2300&	2300&	2677.2&	2300.8&	6055.2&	5536.8&	4460.7&	4935.1&	 6349.5&	5910.6&	5578 \\
                             &Var   &1.96e-13&	3.23e-13&	1.34e+06&	1.61&	83776&	1.82e+06&	 3.21e+06&	3.07e+06&	1.58e+06&	7.54e+05&	2.01e+06 \\
\hline
\hline
Fun &   & $ E(0.5)$  & $E(0.1) $ & $E(0.8)$ & $Rayl(0.4)$ & $Rayl(0.8)$ & $Rayl(0.1)$ & $Weib(1,1.5)$ & $Weib(1.5,1)$ & $Weib(1,1)$ & $random$ & $LHS$ \\
\hline
\multirow{3}{*}{${f_1}$ } &Best &4.36e+05&	6.13e+05&	4.44e+05&	4.05e+05&	3.06e+05&	4.20e+05&	 3.45e+05&	4.01e+05&	4.80e+05&	5.17e+05&	5.67e+05 \\
                             &Mean &7.74e+05&	8.59e+05&	7.08e+05&	7.17e+05&	5.86e+05&	 7.04e+05&	6.44e+05&	6.88e+05&	7.00e+05&	7.96e+05&	8.06e+05 \\
                             &Var  &3.21e+10&	2.56e+10&	2.24e+10&	2.82e+10&	3.33e+10&	 2.79e+10&	2.51e+10&	2.15e+10&	3.21e+19&	6.96e+09&	1.48e+10  \\
 \hline
 \multirow{3}{*}{${f_7}$}   &Best & 700&	700&	700&	700&	700&	700&	700&	700	&700&	 700&	700\\
                             &Mean & 700&	700&	700&	700	&700&	700	&700&	700&	700&	700	 & 700\\
                             &Var & 1.93e-24&	4.76e-23&	8.45e-23&	1.50e-22&	3.28e-21&	 6.39e-23&	1.25e-21&	4.44e-23&	1.53e-22	& 1.13e-20&	1.84e-22 \\
\hline
\multirow{3}{*}{${f_{13}}$}   &Best & 1300.2&	1300.2&	1300.2&	1300.2&	1300.2&	1300.2&	1300.2&	1300.2&	 1300.2&	1300.2&	1300.2\\
                             &Mean  & 1300.3&	1300.3&	1300.3&	1300.3&	1300.3&	1300.3&	1300.3&	1300.3&	 1300.3&	1300.3&	1300.3\\
                             &Var   & 1.30e-03&	1.25e-03&	1.71e-03& 1.17e-03&	9.58e-04&	 8.29e-04&2.46e-03&	1.64e-03&	2.04e-03& 1.18e-03	& 1.05e-03 \\
\hline
\multirow{3}{*}{${f_{18}}$}    &Best &1832.2&	1832.9&	1840.3&	1837.6&	1838.6&	1836&	1836.5&	1828.7&	 1838.8&	1840.1&	1835.5 \\
                             &Mean &1852.7&	1851.6&	1851.7&	1859.6&	1855.6&	1859.7&	1855.9&	1851.6&	 1855.5&	1856.9&	1851.4  \\
                             &Var  &168.69&	84.948&	79.543&	193.97&	144.25&	127.88&	152.01&	140.67&	 139.09&	124.63&	96.454\\
\hline
\multirow{3}{*}{${f_{20}}$}    &Best &2034.2&	2032.3&	2033.2&	2031.4&	2034.7&	2034.4&	2034.4&	2026.8&	 2028.9&	2030.7&	2034.8\\
                             &Mean  &2041.5	&2043.7&	2042.9&	2044.9&	2044.3&	2041.8&	2044.3&	2042.4&	 2043.6&	2044&	2043.5 \\
                             &Var   &26.93&	27.444&	40.999&	32.61&	57.805&	27.786&	37.643	&42.141&	 32.321&	51.707&	39.462\\
\hline
\multirow{3}{*}{${f_{23}}$ } &Best &2615.2&	2615.2&	2615.2&	2615.2&	2615.2&	2615.2	&2615.2&	2615.2&	 2615.2	&2615.2&	2615.2\\
                             &Mean &2615.2&	2615.2&	2615.2	&2615.2	&2615.2	&2615.2&	2615.2&	2615.2&	 2615.2	&2615.2	&  2615.2  \\
                             &Var   &2.18e-25&	2.07e-25&	2.18e-25&	2.18e-25&	2.18e-25&	 2.18e-25&	2.18e-25&	2.18e-25&	2.18e-25&	2.18e-25&	2.18e-25 \\
\hline
 \multirow{3}{*}{${f_{25}}$ } &Best & 2703.8&	2703.4&	2704.1&	2703.9&	2704.3&	2703.2	&2703.9&	 2704.1&	2703.5&	2703.5	&2703.8\\
                             &Mean  &2704.7	&2703.9&	2704.9&	2704.8&	2705&	2703.9&	2704.7&	2705&	 2704.8&	2704.8	&2704.7\\
                             &Var   &0.2876&	0.1534&	0.3632&	0.3239&	0.2161	&0.0997&	0.2584&	 0.2986&	0.2694&	0.4495	& 0.2606\\
\hline
\multirow{3}{*}{${F_{8}}$}    &Best &886.82&	884.55&	881.53&	902.53&	888.96&	871.05&	896.85&	865.81&	 892.52&	892.77&	883.74\\
                             &Mean  &917.25&	925.93&	915.66&	926.21&	923.64&	914.05&	919.95&	913.91&	 915.83&	920.02&	919.88\\
                             &Var   &232.06&	406.49&	368.09&	171.7&	391.17&	330.79&	363.33&	342.47&	 173.41&	140.17&	353.47 \\
\hline
\multirow{3}{*}{${F_{14}}$ } &Best &1445.9&	1443.5&	1445.8&	1448&	1439.1&	1447.7&	1442.7&	1448.4&	 1442.7&	1448.1&	1449.5\\
                             &Mean  &1457&	1458.6&	1454.9&	1459.3&	1455.8&	1456.3&	1452.4&	1457.5&	 1455.8&	1455.8&	1457\\
                             &Var   &34.19&	40.95&	38.71&	58.42&	84.53&	39.29&	37.57&	32.53&	 44.76&	25.117&	18.55\\
\hline
 \multirow{3}{*}{${F_{22}}$ } &Best &2300.2&	5476.3&	2301&	2300&	2309.9&	5289.8&	2302.5&	2303.1&	 2302.5&	2300&	2300\\
                             &Mean  &4251&	5851.4&	4626.8&	2493.3&	5714&	5744.1&	4728.8&	5884&	 5435.1&	3045.9&	2304.2 \\
                             &Var   &3.13e+06&	44081&	3.05e+06&	7.47e+05&	1.40e+06&	45164&	 3.28e+06&	7.72e+05&	1.86e+06&	2.26e+06&	112.71 \\
\hline
\end{tabular}
\end{adjustwidth}
\end{table}

From these tables, we can see that the CEC2014 and CEC2017 benchmark problems are indeed more difficult and challenging than the problems considered in the previous subsection. In other words, different search areas have different characteristics and most of the CEC problems have many local optima. More detailed results regarding the quality of the solutions obtained are shown in Table~\ref{Table-21}, Table~\ref{Table-22}, and Table~\ref{Table-23}. From these results, it is clear that the performance of these algorithms for most functions were influenced by different initialization methods, but for a few functions, the performance is less affected by different initialization methods.

For all the results of these ten test functions, the Friedman test has used to rank all initialization methods of each algorithm. For the DE-a, the $p$ values of ten functions are 0.191, 0.012, 0.822, 0.742,	0.912, 0.004, 0.0726, 0.858, 0.827, and 0.006, respectively. For the PSO-w, the $p$ values of ten functions are 0.021, 0.024, 0.743, 0.001, 0.031,0.009, 0.004, 0.196, 0.001, and 0.001, respectively. For the CS, the $p$ values of ten functions are 0.005, 0.036, 0.091, 0.006, 0.023,	 0.006, 0.0001, 0.061, 0.275, and 0.013, respectively. The Friedman tests of all three algorithms are summarized in Table~\ref{Table-24}. The $p$ value of the DE-a is 0.3045, which is greater than $0.05$, so we cannot reject the null hypothesis at the $\alpha=0.05$ level. It can be considered that various initialization methods have little influence on the DE-a algorithm. The $p$ values of both the PSO-w and CS are all less than $0.05$, this means that the effects of different initialization methods are significantly different. The robustness of three algorithms is completely consistent with the previous experimental results, which again shows that initialization is important and their detailed study is necessary.

\begin{table}[h]
\begin{adjustwidth}{-1.3cm}{}
\centering
\tiny
\caption{Friedman ranks of different initialization methods for three algorithms.\label{Table-24} }
\begin{tabular}{|p{0.85cm}<{\centering}|p{0.85cm}<{\centering}|p{0.85cm}<{\centering}|p{1.1cm}<{\centering}|p{0.9cm}<{\centering}|p{0.9cm}<{\centering}|p{0.9cm}<{\centering}|p{0.9cm}<{\centering}|p{1.2cm}<{\centering}|p{1.2cm}<{\centering}|p{1.3cm}<{\centering}|p{1.1cm}<{\centering}|p{1.1cm}<{\centering}|}
\hline
\multirow{4}{*}{DE}
 &$p$ & $ Be(3,2)$  & $ Be(2.5,2.5)$ & $Be(2,3)$ & $U(0,1)$ & $N(0,1)$ & $N(0.5,1)$ & $N(0.5,0.5)$ & $logn(0,1)$ & $logn(.69,.25)$ & $logn(0,0.5)$ & $logn(0,2/3)$ \\
\cline{2-13}
 &0.30&  9& 5& 11& 22& 8& 12& 21& 2& 20& 1& 14\\
\cline{2-13}
 &  &$ E(0.5)$  & $E(0.1) $ & $E(0.8)$ & $Rayl(0.4)$ & $Rayl(0.8)$ & $Rayl(0.1)$ & $Weib(1,1.5)$ & $Weib(1.5,1)$ & $Weib(1,1)$ & $random$ & $LHS$ \\
\cline{2-13}
 & & 15&4&16&7&18&17&3&6&13&10&19\\
\hline
\hline
\multirow{4}{*}{PSO-w}
&  $p$ & $ Be(3,2)$  & $ Be(2.5,2.5)$ & $Be(2,3)$ & $U(0,1)$ & $N(0,1)$ & $N(0.5,1)$ & $N(0.5,0.5)$ & $logn(0,1)$ & $logn(.69,.25)$ & $logn(0,0.5)$ & $logn(0,2/3)$ \\
\cline{2-13}
 & 0.04&3&	16&	1&	5&	22&	21&	10&	18&	14&	15&	17\\
\cline{2-13}
  &  &$ E(0.5)$  & $E(0.1) $ & $E(0.8)$ & $Rayl(0.4)$ & $Rayl(0.8)$ & $Rayl(0.1)$ & $Weib(1,1.5)$ & $Weib(1.5,1)$ & $Weib(1,1)$ & $random$ & $LHS$ \\
\cline{2-13}
  &  &8	&4&	13&	12&	19&	9	&11&	20	&6&	2&	7\\
\hline
\hline
\multirow{4}{*}{CS}
 & $p$ & $ Be(3,2)$  & $ Be(2.5,2.5)$ & $Be(2,3)$ & $U(0,1)$ & $N(0,1)$ & $N(0.5,1)$ & $N(0.5,0.5)$ & $logn(0,1)$ & $logn(.69,.25)$ & $logn(0,0.5)$ & $logn(0,2/3)$ \\
\cline{2-13}
&0.01&7	&4	&18&	10&	14&	12&	13&	15&	22&	5&	17\\
\cline{2-13}
 &&$ E(0.5)$  & $E(0.1) $ & $E(0.8)$ & $Rayl(0.4)$ & $Rayl(0.8)$ & $Rayl(0.1)$ & $Weib(1,1.5)$ & $Weib(1.5,1)$ & $Weib(1,1)$ & $random$ & $LHS$ \\
\cline{2-13}
 & & 1&	3&	11&	20&	19&	2&	21&	8&	6&	16&	9\\
\hline
\end{tabular}
\end{adjustwidth}
\end{table}

Furthermore, the top three most suitable initialization methods are: $Be(2,3)$, Random and $Be(3,2)$ for the PSO-w algorithm. For the CS, $E(0.5)$, $Rayl(0.1)$, $E(0.1)$ perform better in solving the CEC2014 and CEC2017 problems. Although initialization has little effect on the DE-a algorithm, some methods are still preferred. The recommended initialization methods are $logn(0,0.5)$, $logn(0,1)$, and $Weib(1,1.5)$. This seems to be slightly inconsistent with the earlier conclusion. The reason may be that these CEC2014 and CEC2017 problems are much more complex and multimodal. It can be expected that a combination of different initialization methods may be useful to enhance the diversity of the population.

Based on all the experimental results, for the 19 test functions in this paper, we can say that 43.37$\%$ of the functions using the DE-a algorithm show significant differences for different initialization methods, while 73.68$\%$ of the functions using the PSO-w and the CS are significantly affected by different initialization methods. In other words, initialization methods may have some effects on the performance of the algorithm, whether it is for less complex functions or more complex problems.

\FloatBarrier
\section{Discussions}

Based on the above extensive simulations and tests, {{we can now investigate any possible correlation between the average distance of the initial population from the true optimal solution and the performance of final solutions found by algorithms. We also discuss their implications. In addition, we will give the initialization suggestions for the other two algorithms: ABC and GA. }}

\subsection{Correlations and discussions}

We now conduct some further analyses and discussions to demonstrate the influence of the initial population with different distributions on the results of algorithms. To begin with, we first define the average distance $\Delta$ between the initial population and the corresponding real optimal solution as
\begin{equation}\label{M_dis}
\Delta  = \frac{{\sum\nolimits_{i = 1}^{NP} {\sum\nolimits_{j = 1}^D {\left| {{x_{i,j}} - {x_j}^{opt}} \right|} } }}{{NP}}
\end{equation}
where ${{x_{i,j}}}$ indicates the $j$-th dimension of the $i$-th individual, $NP$ represents the total population size, and ${x_j}^{opt}$ represents the $j$-th dimension of the true optimal solution.

In order to explore the relationship between the initial distribution and the performance of an algorithm (in terms of the final solution obtained), we investigate if the average distance between the generated initial population and the true optimal solution ($\Delta$) has any positive correlation to the distance between the obtained final solution from the true optimal solution.
Each initialization method for each algorithm runs 20 times independently, so as to avoid any potential bias of the experiment. Let $\overline \Delta $ be the mean value of $\Delta$ in 20 experiments. The previously defined `$Dist$' represents the average distance between the optimal solution obtained by the algorithm and the real optimal solution in $NT$ (here $NT=20$) experiments. Then, we carry out some analyses to see if there is any connection between $\overline \Delta$ and `$Dist$'. The results are summarized in Table~\ref{Table-16} to Table~\ref{Table-18}.

\begin{table}[h]
\begin{adjustwidth}{-1.7cm}{}
\centering
\tiny
\caption{The result of $\overline \Delta$ and $Dist$ on DE-a. \label{Table-16} }
\begin{tabular}{|p{0.2cm}|p{0.45cm}|p{1.06cm}<{\centering}|p{1.08cm}<{\centering}|p{1.06cm}<{\centering}|p{1.06cm}<{\centering}|p{1.06cm}<{\centering}|p{1.06cm}<{\centering}|p{1.2cm}<{\centering}|p{1.2cm}<{\centering}|p{1.3cm}<{\centering}|p{1.09cm}<{\centering}|p{1.12cm}<{\centering}|}
\hline
 Fun & Value  & $ Be(3,2)$  & $ Be(2.5,2.5)$ & $Be(2,3)$ & $U(0,1)$ & $N(0,1)$ & $N(0.5,1)$ & $N(0.5,0.5)$ & $logn(0,1)$ & $logn(.69,.25)$ & $logn(0,0.5)$ & $logn(0,2/3)$ \\
\hline
 \multirow{2}{*}{${f_1}$} &$\overline \Delta$ &49.68&	55.83&	71.94&	77.85&	281.77&	241.58&	121.80&	 362.31&	436.91	&170.55	&216.18\\
                             &Dist  &0.20&	0.20&	0.60&	0.50&	0.30&	0.60&	0.80&	0.20&	 0.20&	0.20&	0.20\\
\hline
\multirow{2}{*}{${f_2}$}  &$\overline \Delta$ &112.27&	101.99&	112.42&	150.39&	536.64&	478.01&	237.95&	 751.80&	933.73&	389.20&	473.33\\
                          &Dist & 5.33e-14	&5.32e-14&	5.50e-14&	0.08&	0.08&	5.19e-14&	 5.77e-14&	5.61e-14&	5.76e-14&	5.64e-14&	5.19e-14\\
\hline
\multirow{2}{*}{${f_3}$}  & $\overline \Delta$ & 56.40&	50.77&	55.97&	75.09&	267.53&	240.01&	119.68&	 369.47&	466.27&	194.37&	238.51 \\
                             &Dist  & 9.39e-94&	9.69e-94&	8.54e-94&	5.78e-94&	5.78e-94&	 2.18e-93&	6.32e-94&	1.09e-93&	2.28e-93&	1.36e-93&	3.34e-94\\
\hline
\multirow{2}{*}{${f_4}$}    &$\overline \Delta$ & 57.35&	52.26&	57.83&	76.71&	275.66&	245.93&	 122.35&	382.70&	478.07&	199.97&	242.47\\
                             &Dist  & 13.58&	15.47&	12.25&	12.70&	12.89&	15.04&	13.57&	13.90&	 12.76&	14.65&	12.51\\
\hline
\multirow{2}{*}{${f_5}$}    &$\overline \Delta$ & 6.77e+03&	6.11e+03&	6.76e+03&	8.96e+03&	 3.22e+04&	2.89e+04&	1.43e+04&	4.44e+04&	5.62e+04&	2.33e+04&	2.86e+04 \\
                             &Dist  &2.64&	4.08&	3.14&	2.65&	2.67&	1.63&	3.68&	2.98&	 4.03&	4.76&	3.87\\
\hline
\multirow{2}{*}{${f_6}$ } &$\overline \Delta$ &1.12e+03&	1.02e+03&	1.12e+03&	1.50e+03&	 5.36e+03&	4.77e+03&	2.39e+03&	7.44e+03&	9.33e+03&	3.88e+03&	4.74e+03\\
                             &Dist  &1.06&	0.61&	1.60&	2.28&	2.16&	2.39&	4.44&	1.77&	 3.84&	2.62&	2.19\\
\hline
 \multirow{2}{*}{${f_7}$ } &$\overline \Delta$ & 112.21&	101.87&	111.88&	149.37&	532.96&	479.66&	 239.24&	739.01&	932.91&	389.61&	473.91\\
                             &Dist  & 5.40&	2.43&	4.10&	4.44&	3.69&	3.91&	3.15&	3.47&	 4.09&	5.41&	3.78\\
 \hline
 \multirow{2}{*}{${f_8}$ } &$\overline \Delta$ &1.09e+03&	1.02e+03&	1.16e+03&	1.49e+03&	 5.43e+03&	4.79e+03&	2.41e+03&	7.46e+03&	9.25e+03&	3.81e+03&	4.64e+03\\
                             &Dist &1.43e+03&	1.35e+03&	1.37e+03&	1.31e+03&	1.36e+03&	 1.35e+03&	1.33e+03&	1.35e+03&	1.35e+03&	1.37e+03&	1.36e+03 \\
  \hline
 \multirow{2}{*}{${f_9}$ } &$\overline \Delta$ &9.69e+03&	1.26e+04&	1.56e+04&	1.28e+04&	 3.35e+04&	2.60e+04&	1.60e+04&	3.43e+04&	3.39e+04&	1.32e+04&	1.86e+04\\
                             &Dist  &1.44e+03&	1.59e+03&	2.76e+03&	1.77e+03&	3.76e+04&	 2.77e+04&	1.28e+03&	5.79e+04&	3.51e+04&	1.20e+04&	2.49e+04\\
\hline
\hline
Fun &   & $ E(0.5)$  & $E(0.1) $ & $E(0.8)$ & $Rayl(0.4)$ & $Rayl(0.8)$ & $Rayl(0.1)$ & $Weib(1,1.5)$ & $Weib(1.5,1)$ & $Weib(1,1)$ & $random$ & $LHS$ \\
\hline
 \multirow{2}{*}{${f_1}$} &$\overline \Delta$ &120.89&	150.23&	166.66&	69.82&	151.70&	142.42&	149.66&	 334.02&	211.59&	77.96&	77.99\\
                             &Dist  &0.30&	0.50&	0.30&	0.40&	0.40&	0.30&	0.40&	0.20&	 0.50&	0.20&	2.42e-15\\
\hline
\multirow{2}{*}{${f_2}$}  &$\overline \Delta$ &220.39&	241.23&	335.92&	126.67&	337.15&	224.77&	318.54&	 691.83&	429.84&	150.48&	150.01\\
                          &Dist & 5.41e-14&	5.46e-14&	4.94e-14&	5.75e-14&	5.35e-14&	5.17e-14&	 5.62e-14&	5.46e-14&	5.46e-14&	5.77e-14&	5.30e-14\\
\hline
\multirow{2}{*}{${f_3}$}  & $\overline \Delta$ & 110.39&	120.58&	166.38&	62.85&	169.67&	112.37&	 158.61&	347.37&	213.49&	74.80&	75.00 \\
                             &Dist  & 5.82e-94&	6.52e-94&	1.31e-93&	6.80e-94&	1.23e-93&	 1.07e-93&	8.57e-94&	1.28e-94&	6.02e-94&	4.70e-94&	7.76e-94\\
\hline
\multirow{2}{*}{${f_4}$}    &$\overline \Delta$ & 113.31&	123.09&	172.12&	64.60&	172.07&	115.14&	 163.03&	351.33&	221.10&	77.04&	76.79\\
                             &Dist  & 15.36&	12.66&	14.31&	16.23&	13.23&	12.92&	14.06&	10.91&	 13.75&	12.58&	15.34\\
\hline
\multirow{2}{*}{${f_5}$}    &$\overline \Delta$ & 1.32e+04&	1.44e+04&	2.00e+04&	7.58e+03&	 2.03e+04&	1.35e+04&	1.91e+04&	4.10e+04&	2.54e+04&	8.98e+03&	8.99e+03 \\
                             &Dist  &1.87&	3.23&	4.83&	3.97&	1.72&	3.54&	4.36&	3.07&	 3.44&	4.14&	1.68\\
\hline
\multirow{2}{*}{${f_6}$ } &$\overline \Delta$ &2.21e+03&	2.41e+03&	3.32e+03&	1.26e+03&	 3.37e+03&	2.25e+03&	3.19e+03&	6.81e+03&	4.27e+03&	1.49e+03&	1.50e+03\\
                             &Dist  &4.17&	2.03&	2.81&	1.37&	1.73&	1.58&	2.95&	2.17&	 2.23&	2.75&	3.63\\
\hline
 \multirow{2}{*}{${f_7}$ } &$\overline \Delta$ & 220.73&	240.91&	333.45&	126.38&	339.20&	224.52&	 316.49&	687.03&	427.83&	150.72&	149.99\\
                             &Dist  & 5.15&	3.91&	3.53&	2.88&	4.40&	6.25&	4.04&	4.74&	 4.56&	4.04&	5.20\\
 \hline
 \multirow{2}{*}{${f_8}$ } &$\overline \Delta$ &2.23e+03&	2.50e+03&	3.35e+03&	1.27e+03&	 3.31e+03&	2.34e+03&	3.12e+03&	6.85e+03&	4.29e+03&	1.50e+03&	1.50e+03\\
                             &Dist & 1.34e+03&	1.37e+03&	1.37e+03&	1.30e+03&	1.37e+03&	 1.34e+03&	1.35e+03&	1.39e+03&	1.31e+03&	1.32e+03&	1.37e+03\\
  \hline
 \multirow{2}{*}{${f_9}$ } &$\overline \Delta$ &1.73e+04&	2.46e+04&	1.88e+04&	1.33e+04&	 1.26e+04&	2.39e+04&	1.46e+04&	3.15e+04&	2.15e+04&	1.28e+04&	1.28e+04\\
                             &Dist  &1.86e+03&	2.71e+03&	1.14e+04&	2.47e+03&	796.78&	9.39e+03&	 5.82e+03&	3.84e+04&	2.44e+04&	1.60e+03&	1.74e+03\\
\hline
\end{tabular}
\end{adjustwidth}
\end{table}

\begin{table}[h]
\begin{adjustwidth}{-1.7cm}{}
\centering
\tiny
\caption{The result of $\overline \Delta$ and $Dist$ on PSO-w. \label{Table-17} }
\begin{tabular}{|p{0.2cm}|p{0.45cm}|p{1.06cm}<{\centering}|p{1.08cm}<{\centering}|p{1.06cm}<{\centering}|p{1.06cm}<{\centering}|p{1.06cm}<{\centering}|p{1.06cm}<{\centering}|p{1.2cm}<{\centering}|p{1.2cm}<{\centering}|p{1.3cm}<{\centering}|p{1.09cm}<{\centering}|p{1.12cm}<{\centering}|}
\hline
 Fun & Value  & $ Be(3,2)$  & $ Be(2.5,2.5)$ & $Be(2,3)$ & $U(0,1)$ & $N(0,1)$ & $N(0.5,1)$ & $N(0.5,0.5)$ & $logn(0,1)$ & $logn(.69,.25)$ & $logn(0,0.5)$ & $logn(0,2/3)$ \\
\hline
 \multirow{2}{*}{${f_1}$} &$\overline \Delta$ &49.747&	56.00&	71.59&	77.96&	281.06&	240.68&	121.98&	 359.49&	437.16&	171.58&	217.81\\
                             &Dist  &2.55&	19.89&	20.26&	18.88&	20.69&	19.93&	19.38&	18.62&	 20.02&	17.10&	15.12\\
\hline
\multirow{2}{*}{${f_2}$}  &$\overline \Delta$ &120.28&	150.15&	167.03&	69.78&	151.88&	142.40&	149.09&	 333.41&	209.28&	78.02&	78.00\\
                             &Dist  &20.40&	19.59&	19.87&	19.63&	17.26&	19.96&	16.49&	17.79&	 20.01&	20.02&	20.15\\
\hline
\multirow{2}{*}{${f_3}$}  & $\overline \Delta$ & 56.23&	50.89&	56.29&	74.98&	268.60&	239.30&	119.66&	 373.32&	467.00&	194.48&	236.53 \\
                             &Dist  & 1.90e-06&	1.58e-06&	1.70e-06&	1.73e-06&	1.71e-06&	 1.61e-06&	1.80e-06&	1.90e-06&	1.83e-06&	1.73e-06&	1.87e-06\\
\hline
\multirow{2}{*}{${f_4}$}    &$\overline \Delta$ & 56.25&	50.92&	56.29&	75.03&	268.84&	239.34&	 119.67&	372.52&	467.11&	194.41&	236.46\\
                             &Dist  & 20.35&	14.14&	20.71&	17.02&	17.67&	15.93&	17.12&	17.87&	 16.28&	22.05&	19.11\\
\hline
\multirow{2}{*}{${f_5}$}    &$\overline \Delta$ & 6.75e+03&	6.11e+03&	6.75e+03&	9.00e+03&	 3.22e+04&	2.87e+04&	1.44e+04&	4.48e+04&	5.60e+04&	2.34e+04&	2.84e+04\\
                             &Dist  &6.25&	5.78&	5.22&	6.87&	5.15&	7.94&	4.67&	5.47&	 7.24&	8.22&	6.36\\
\hline
\multirow{2}{*}{${f_6}$ } &$\overline \Delta$ &1.12e+03&	1.02e+03&	1.12e+03&	1.50e+03&	 5.37e+03&	4.79e+03&	2.39e+03&	7.46e+03&	9.34e+03&	3.89e+03&	4.73e+03\\
                             &Dist  &0.15&	0.03&	0.16&	0.08&	0.10&	0.10&	0.09&	0.08&	 0.09&	0.19&	0.15\\
\hline
 \multirow{2}{*}{${f_7}$ } &$\overline \Delta$ & 112.52&	101.83&	112.55&	150.02&	537.84&	478.52&	 239.50&	748.06&	934.13&	389.29&	472.24\\
                             &Dist  & 39.44&	14.34&	37.13&	17.29&	22.75&	18.77&	24.64&	22.36&	 16.92&	30.09&	28.27\\
 \hline
 \multirow{2}{*}{${f_8}$ } &$\overline \Delta$ &1.09e+03&	1.02e+03&	1.16e+03&	1.50e+03&	 5.41e+03&	4.79e+03&	2.39e+03&	7.41e+03&	9.25e+03&	3.82e+03&	4.67e+03\\
                             &Dist &1.10e+03&	1.02e+03&	1.14e+03&	1.47e+03&	5.44e+03&	 4.87e+03&	2.45e+03&	7.09e+03&	9.22e+03&	3.74e+03&	4.41e+03 \\
  \hline
 \multirow{2}{*}{${f_9}$ } &$\overline \Delta$ &9.68e+03&	1.26e+04&	1.56e+04&	1.28e+04&	 3.34e+04&	2.60e+04&	1.60e+04&	3.38e+04&	3.41e+04&	1.32e+04&	1.87e+04\\
                             &Dist  &6.22e+03&	9.97e+03&	1.39e+04&	8.69e+03&	4.01e+04&	 2.98e+04&	1.84e+04&	1.03e+05&	3.65e+04&	1.83e+04&	3.03e+04\\
\hline
\hline
Fun &   & $ E(0.5)$  & $E(0.1) $ & $E(0.8)$ & $Rayl(0.4)$ & $Rayl(0.8)$ & $Rayl(0.1)$ & $Weib(1,1.5)$ & $Weib(1.5,1)$ & $Weib(1,1)$ & $random$ & $LHS$ \\
\hline
 \multirow{2}{*}{${f_1}$} &$\overline \Delta$ &120.28&	150.15&	167.03&	69.78&	151.88&	142.40&	149.09&	 333.41&	209.28&	78.02&	78.00\\
                             &Dist  &20.40&	19.59&	19.87&	19.63&	17.26&	19.96&	16.49&	17.79&	 20.01&	20.02&	20.15\\
\hline
\multirow{2}{*}{${f_2}$}  &$\overline \Delta$ &220.66&	240.79&	333.92&	126.32&	338.74&	224.82&	317.97&	 690.47&	427.24&	149.94&	150.00\\
                          &Dist & 3.36e-04&	0.36&	0.26&	0.23&	0.56&	0.43&	0.11&	0.32&	 0.50&	0.09&	0.17\\
\hline
\multirow{2}{*}{${f_3}$}  & $\overline \Delta$ & 110.38&	120.43&	166.65&	63.18&	169.11&	112.39&	 158.93&	344.79&	213.83&	74.99&	75.00\\
                             &Dist  & 1.58e-06&	1.65e-06&	1.77e-06&	1.47e-06&	1.85e-06&	 1.79e-06&	1.73e-06&	1.86e-06&	1.76e-06&	1.73e-06&	1.69e-06\\
\hline
\multirow{2}{*}{${f_4}$}    &$\overline \Delta$ & 110.39&	120.39&	167.03&	63.18&	169.16&	112.39&	 158.89&	345.13&	214.05&	75.02&	75.00\\
                             &Dist  & 18.56&	23.09&	17.42&	15.13&	20.80&	23.34&	18.07&	18.01&	 16.22&	16.27&	16.92\\
\hline
\multirow{2}{*}{${f_5}$}    &$\overline \Delta$ & 1.32e+04&	1.44e+04&	2.00e+04&	7.58e+03&	 2.03e+04&	1.35e+04&	1.91e+04&	4.14e+04&	2.57e+04&	9.01e+03&	9.00e+03\\
                             &Dist  &7.27&	7.29&	5.73&	8.25&	9.39&	8.74&	7.24&	5.19&	 7.47&	7.51&	6.13\\
\hline
\multirow{2}{*}{${f_6}$ } &$\overline \Delta$ &2.21e+03&	2.41e+03&	3.34e+03&	1.26e+03&	 3.38e+03&	2.25e+03&	3.18e+03&	6.89e+03&	4.28e+03&	1.50e+03&	1.50e+03\\
                             &Dist  &0.12&	0.23&	0.11&	0.07&	0.16&	0.22&	0.15&	0.10&	 0.11&	0.07&	0.06\\
\hline
 \multirow{2}{*}{${f_7}$ } &$\overline \Delta$ & 220.64&	240.79&	333.92&	126.25&	338.46&	224.74&	 317.27&	690.00&	427.88&	149.96&	149.99\\
                             &Dist  & 25.05&	43.34&	22.18&	18.09&	31.77&	45.85&	29.09&	19.63&	 23.20&	16.61&	18.17\\
 \hline
 \multirow{2}{*}{${f_8}$ } &$\overline \Delta$ &2.23e+03&	2.50e+03&	3.34e+03&	1.27e+03&	 3.32e+03&	2.34e+03&	3.15e+03&	6.86e+03&	4.26e+03&	1.50e+03&	1.50e+03\\
                             &Dist & 2.19e+03&	2.51e+03&	3.80e+03&	1.26e+03&	3.18e+03&	 2.33e+03&	3.11e+03&	6.87e+03&	4.18e+03&	1.55e+03&	1.51e+03\\
  \hline
 \multirow{2}{*}{${f_9}$ } &$\overline \Delta$ &1.74e+04&	2.46e+04&	1.88e+04&	1.32e+04&	 1.26e+04&	2.39e+04&	1.46e+04&	3.13e+04&	2.15e+04&	1.28e+04&	1.28e+04\\
                             &Dist  &1.98e+04&	1.19e+04&	2.69e+04&	1.01e+04&	1.48e+04&	 1.62e+04&	1.77e+04&	4.58e+04&	3.00e+04&	9.07e+03&	8.05e+03\\
\hline
\end{tabular}
\end{adjustwidth}
\end{table}

\begin{table}[h]
\begin{adjustwidth}{-1.7cm}{}
\centering
\tiny
\caption{The result of $\overline \Delta$ and $Dist$ on CS. \label{Table-18} }
\begin{tabular}{|p{0.2cm}|p{0.45cm}|p{1.06cm}<{\centering}|p{1.08cm}<{\centering}|p{1.06cm}<{\centering}|p{1.06cm}<{\centering}|p{1.06cm}<{\centering}|p{1.06cm}<{\centering}|p{1.2cm}<{\centering}|p{1.2cm}<{\centering}|p{1.3cm}<{\centering}|p{1.09cm}<{\centering}|p{1.12cm}<{\centering}|}
\hline
 Fun & Value  & $ Be(3,2)$  & $ Be(2.5,2.5)$ & $Be(2,3)$ & $U(0,1)$ & $N(0,1)$ & $N(0.5,1)$ & $N(0.5,0.5)$ & $logn(0,1)$ & $logn(.69,.25)$ & $logn(0,0.5)$ & $logn(0,2/3)$ \\
\hline
 \multirow{2}{*}{${f_1}$} &$\overline \Delta$ &49.73&	56.26&	71.94&	78.38&	281.48&	241.61&	122.37&	 353.43&	437.41&	172.19&	218.41\\
                             &Dist  &9.86e-04&	1.22e-05&	1.87e-05&	3.16e-05&	0.20&	0.10&	 0.10&	0.15&	114.56&	0.21&	0.02\\
\hline
\multirow{2}{*}{${f_2}$}  &$\overline \Delta$ &113.18&	101.92&	112.73&	149.55&	537.84&	478.92&	240.10&	 751.52&	929.21&	392.43&	476.58\\
                             &Dist  &0.29&	0.08&	3.19e-14&	0.0821&	0.1307&	0.08&	3.23e-14&	 3.17e-14&	276.49&	0.18&	0.08\\
\hline
\multirow{2}{*}{${f_3}$}  & $\overline \Delta$ & 56.28&	50.79&	56.12&	75.29&	270.07&	242.34&	120.46&	 361.59&	469.61&	193.96&	239.33\\
                             &Dist  & 1.68e-33&	9.69e-34&	1.07e-33&	1.23e-33&	1.94e-33&	 1.38e-33&	1.71e-33&	3.11e-33&	138.75&	1.45e-33&	2.11e-33\\
\hline
\multirow{2}{*}{${f_4}$}    &$\overline \Delta$ &57.77&	52.27&	57.58&	77.14&	275.17&	244.98&	122.82&	 379.29&	478.37&	199.52&	238.07 \\
                             &Dist  & 7.47&	6.71&	7.20&	7.94&	9.39&	7.99&	8.24&	9.19&	 145.21&	10.08&	9.19\\
\hline
\multirow{2}{*}{${f_5}$}    &$\overline \Delta$ & 6.79e+03&	6.07e+03&	6.75e+03&	9.01e+03&	 3.18e+04&	2.88e+04&	1.44e+04&	4.53e+04&	5.57e+04&	2.34e+04&	2.85e+04\\
                             &Dist  &5.49e-07&	0.47&	5.58e-07&	5.28e-07&	5.39e-07&	5.63e-07&	 5.44e-07&	5.56e-07&	1.66e+04&	5.56e-07&	5.65e-07 \\
\hline
\multirow{2}{*}{${f_6}$ } &$\overline \Delta$ &1.13e+03&	1.02e+03&	1.13e+03&	1.50e+03&	 5.42e+03&	4.80e+03&	2.41e+03&	7.45e+03&	9.34e+03&	3.88e+03&	4.68e+03\\
                             &Dist  &0.41&	0.14&	0.50&	0.55&	1.03&	1.21&	0.72&	1.08&	 3000&	1.05&	0.83\\
\hline
 \multirow{2}{*}{${f_7}$ } &$\overline \Delta$ & 112.33&	102.37&	114.10&	150.21&	535.35&	473.63&	 240.64&	745.48&	935.03&	392.70&	470.15\\
                             &Dist  & 52.99&	32.62&	48.38&	62.53&	92.89&	93.18&	74.23&	127.28&	 297.82&	130.38&	136.33\\
 \hline
 \multirow{2}{*}{${f_8}$ } &$\overline \Delta$ &1089.3&	1019.1&	1165.5&	1510.4&	5379.6&	4753&	2398.2&	 7335.7&	9233.2&	3834.6&	4611.2\\
                             &Dist & 1652.9	&1484&	1744.2&	1771&	1790.2&	1913.5&	1928&	1859.1&	 1918.7&	1913.9&	1654.5\\
  \hline
 \multirow{2}{*}{${f_9}$ } &$\overline \Delta$ &9669.3&	12671&	15598&	12716&	33576&	26275&	16028&	 33901&	33986&	13126&	18727\\
                             &Dist  &1486.8	&4068.6&	6337.3&	3297.6&	11811&	5084&	3071.5&	48085&	 23146&	415.86&	9495.9\\
\hline
\hline
Fun &   & $ E(0.5)$  & $E(0.1) $ & $E(0.8)$ & $Rayl(0.4)$ & $Rayl(0.8)$ & $Rayl(0.1)$ & $Weib(1,1.5)$ & $Weib(1.5,1)$ & $Weib(1,1)$ & $random$ & $LHS$ \\
\hline
 \multirow{2}{*}{${f_1}$} &$\overline \Delta$ &119.31&	150.81&	166.93&	70.51&	151.75&	142.37&	149.90&	 330.09&	208.52&	78.20&	77.98\\
                             &Dist  &0.10&	0.20&	0.10&	1.87e-05&	0.02&	0.20&	1.01&	0.11&	 0.10&	1.59e-05&	4.34e-05\\
\hline
\multirow{2}{*}{${f_2}$}  &$\overline \Delta$ &221.56&	241.44&	332.23&	126.70&	339.83&	224.47&	313.59&	 693.58&	425.49&	149.70&	150.08\\
                          &Dist &0.13&	0.08&	3.18e-14&	0.08&	3.21e-14&	3.09e-14&	3.18e-14&	 0.08&	3.25e-14&	3.09e-14&	3.15e-14\\
\hline
\multirow{2}{*}{${f_3}$}  & $\overline \Delta$ &109.79&	120.15&	165.66&	63.07&	169.69&	112.01&	157.64&	 348.43&	211.76&	75.20&	74.97\\
                             &Dist  & 1.68e-33&	2.45e-33&	1.01e-33&	1.20e-33&	1.77e-33&	 3.33e-33&	1.71e-33&	1.73e-33&	1.61e-33&	1.29e-33&	1.66e-33\\
\hline
\multirow{2}{*}{${f_4}$}    &$\overline \Delta$ & 112.79&	123.01&	170.93&	64.89&	172.29&	114.89&	 162.25&	351.19&	214.81&	76.44&	76.77\\
                             &Dist  & 8.68&	10.10&	6.69&	6.83&	9.65&	9.63&	9.35&	8.73&	 8.33&	7.11&	8.51\\
\hline
\multirow{2}{*}{${f_5}$}    &$\overline \Delta$ &1.33e+04&	1.44e+04&	2.01e+04&	7.59e+03&	 2.03e+04&	1.35e+04&	1.89e+04&	4.12e+04&	2.57e+04&	9.04e+03&	9.00e+03 \\
                             &Dist  &5.48e-07&	0.31&	5.42e-07&	5.31e-07&	5.23e-07&	5.37e-07&	 5.57e-07&	0.76&	5.50e-07&	5.60e-07&	0.38\\
\hline
\multirow{2}{*}{${f_6}$ } &$\overline \Delta$ &2.19e+03&	2.41e+03&	3.35e+03&	1.25e+03&	 3.39e+03&	2.25e+03&	3.21e+03&	6.89e+03&	4.23e+03&	1.50e+03&	1.50e+03\\
                             &Dist  &0.76&	0.95&	0.72&	0.29&	0.69&	1.09&	0.83&	0.89&	 0.82&	0.41&	0.43\\
\hline
 \multirow{2}{*}{${f_7}$ } &$\overline \Delta$ & 221.64&	240.06&	339.18&	127.31&	340.17&	224.67&	 315.12&	691.94&	422.38&	150.08&	150.06\\
                             &Dist  &87.03&	159.94&	91.07&	51.97& 119.77&	142.70&	108.97&	116.65&	 100.92&	64.26&	57.20 \\
 \hline
 \multirow{2}{*}{${f_8}$ } &$\overline \Delta$ &2215.8&	2498.2&	3377.4&	1272.6&	3337.9&	2343.3&	3107.9&	 6838.5&	4289.5&	1504.8&	1501.4\\
                             &Dist & 1863.4&	1882.5&	1793.1&	1721.3&	1826.3&	1784.9&	1864.2&	1802.4&	 1869.5&	1745.2&	1783\\
  \hline
 \multirow{2}{*}{${f_9}$ } &$\overline \Delta$ &17351&	24632&	18814&	13227&	12749&	23862&	14486&	 31290&	21511&	12858&	12815\\
                             &Dist  &4601&	9859.8&	3021.2&	3756.6&	697.04&	10847&	970.16&	32208&	 9003.2&	3756.6&	2992.8\\
\hline
\end{tabular}
\end{adjustwidth}
\end{table}

\begin{figure}[htbp]
\centering
\subfigure[]{\includegraphics[width=4cm]{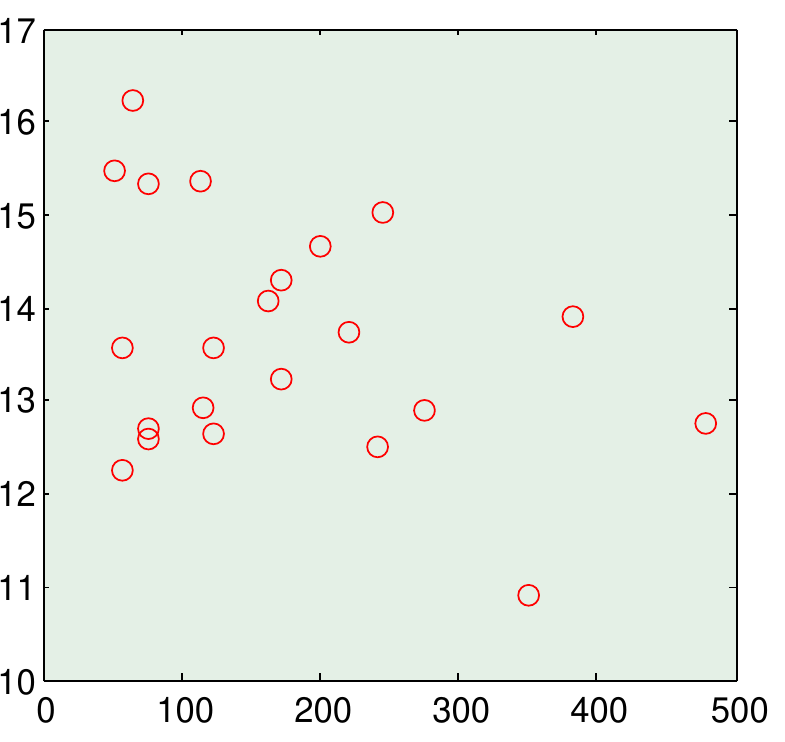}{\label{DE_f4}}}
\subfigure[]{\includegraphics[width=4.3cm]{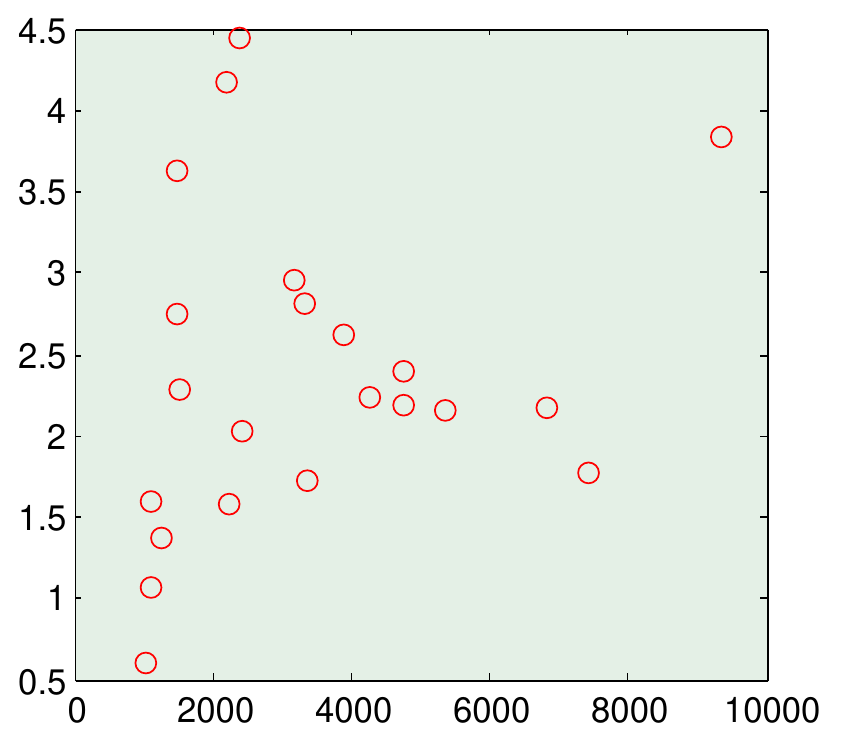}{\label{DE_f6}}}
\subfigure[]{\includegraphics[width=4cm]{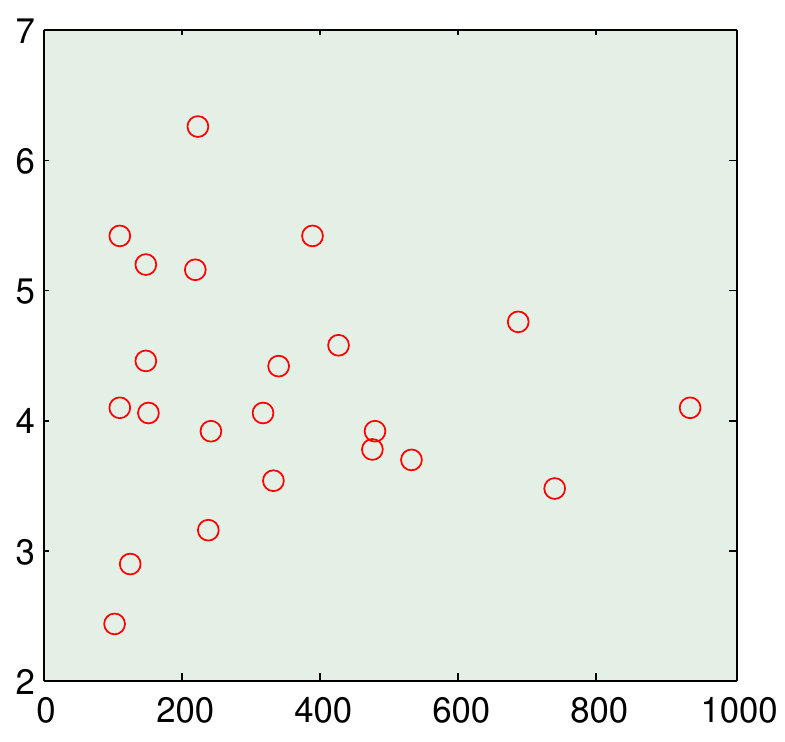}{\label{DE_f7}}}
\caption{The relationship of $\overline \Delta$ and `Dist': (a) $f_4$ of DE-a. (b)$f_6$ of DE-a. (c)$f_7$ of DE-a. The x-axis represents $\overline \Delta$ : the average value of $\Delta $ (the average distance between the initial population and the real optimal solution) in 20 experiments, and the y-axis represents `Dist' : average value of the distance between the best obtained solution and the real optimal solution in 20 experiments. It shows that there is no obvious correlation. }
\label{Fig-10}
\end{figure}

\begin{figure}[htbp]
\centering
\subfigure[]{\includegraphics[width=4cm]{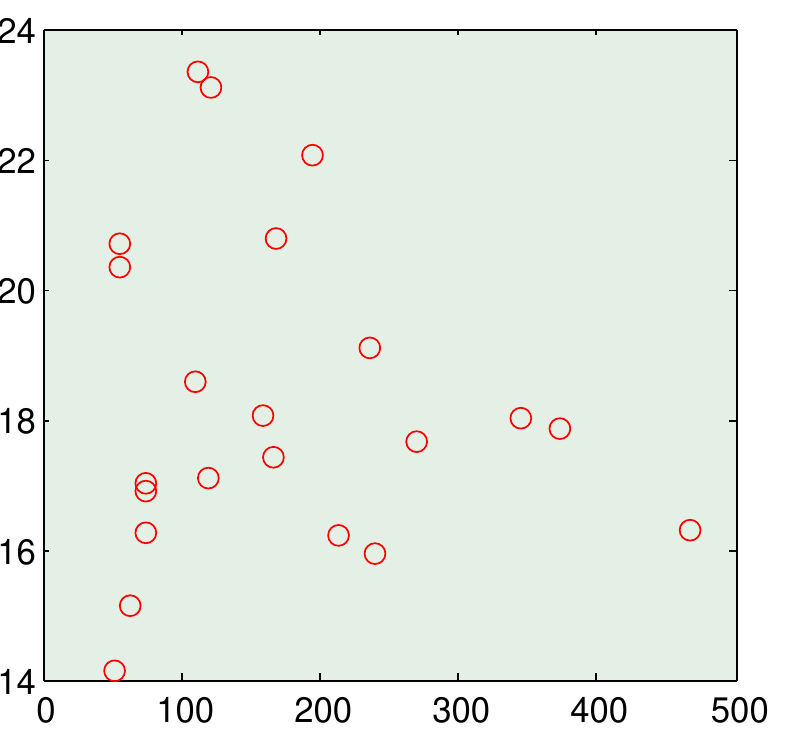}{\label{PSO_f4}}}
\subfigure[]{\includegraphics[width=4.3cm]{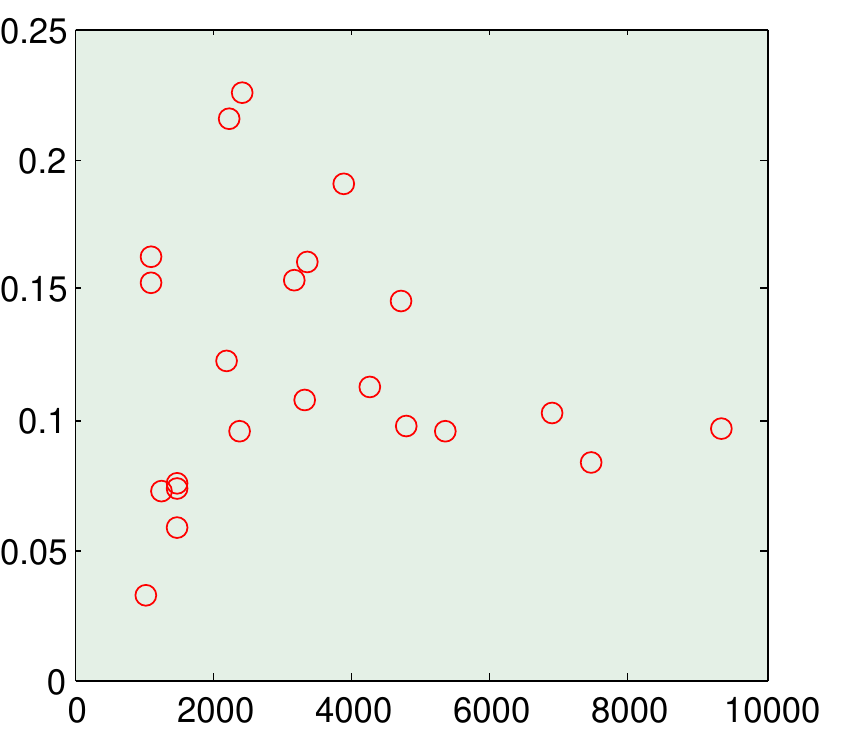}{\label{PSO_f6}}}
\subfigure[]{\includegraphics[width=4cm]{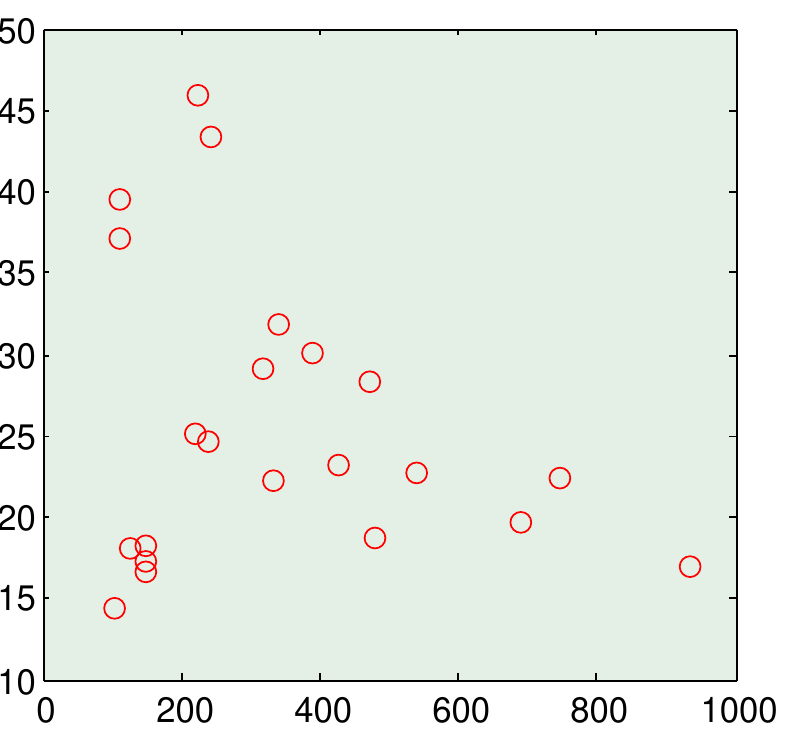}{\label{PSO_f7}}}
\caption{The relationship of $\overline \Delta$ and `Dist': (a)$f_4$ of PSO-w. (b)$f_6$ of PSO-w. (c)$f_7$ of PSO-w. The x-axis represents $\overline \Delta$, and the y-axis represents `Dist'. It can be seen that there is no obvious correlation.}
\label{Fig-11}
\end{figure}

\begin{figure}[htbp]
\centering
\subfigure[]{\includegraphics[width=4cm]{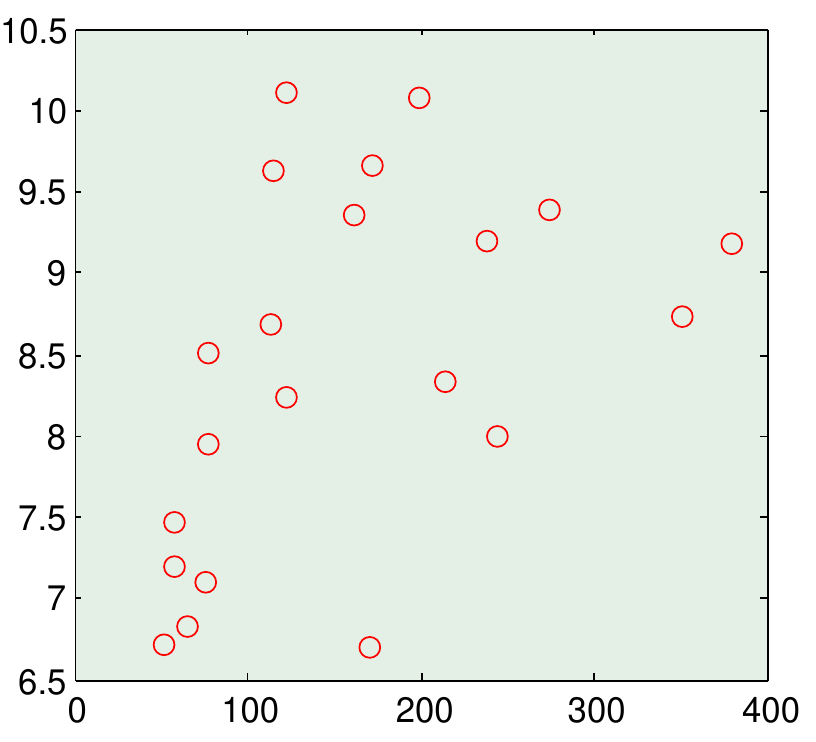}{\label{CS_f4}}}
\subfigure[]{\includegraphics[width=4cm]{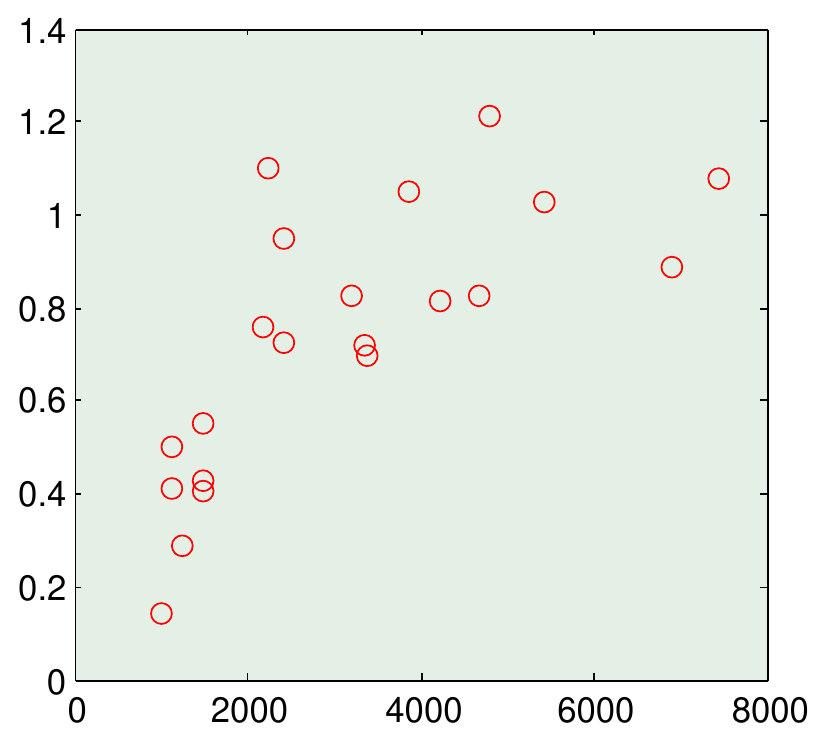}{\label{CS_f6}}}
\subfigure[]{\includegraphics[width=4cm]{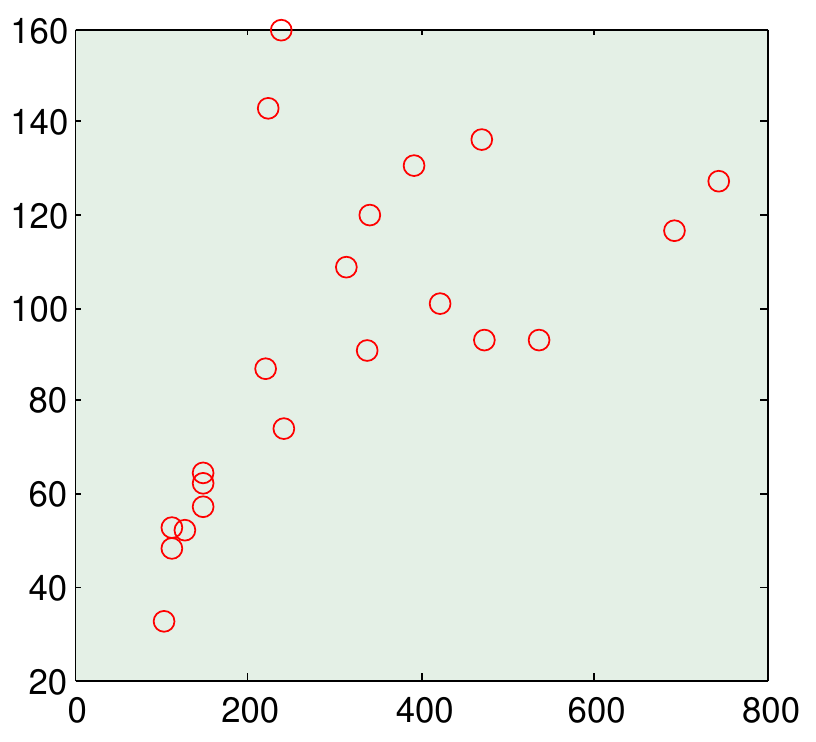}{\label{CS_f7}}}
\caption{The relationship of $\overline \Delta$ and `Dist': (a)$f_4$ of CS. (b)$f_6$ of CS. (c)$f_7$ of CS. Among these three figures, $f_4$ has no significant correlation, while $f_6$ and $f_7$ may have some weak positive correlation.}
\label{Fig-12}
\end{figure}

{{As seen from Tables~\ref{Table-16} to~\ref{Table-18}, $\overline \Delta$ and `Dist' have no significant correlation. For the ease of observation, the relationship between the $\overline \Delta$ and `Dist' of the partial functions is given in Figs.~\ref{Fig-10} to \ref{Fig-12}. It is worth noting that, due to the larger value of $logn(0.69, 0.25)$ used in the CS algorithm, we can essentially consider such an extreme value as an outlier and thus remove it in the correlation test.}}

From a statistical perspective, the correlation tests have been undertaken, with the results shown in Fig.~\ref{Table-19}. Using a significance level of $0.01$, when $p<0.01$ means that the $\overline \Delta$ and `$Dist$' of the corresponding function has a significant correlation, mark with $**$. Another symbol $*$ means that the $\overline \Delta$ and `$Dist$' may have a weak correlation. The results show that except for one function $f_9$, there is no significant correlation for the DE-a algorithm. {{For the PSO-w, the $\overline \Delta$ and `$Dist$' of $f_3$ may have a weak correlation, and those of $f_8$ and $f_9$ have significant correlations between $\overline \Delta$ and `$Dist$'. For other functions, there are no noticeable correlations. For the CS, there are six functions that may have some correlations between $\overline \Delta$ and `$Dist$'. }}

\begin{table}[h]
\begin{adjustwidth}{-0.5cm}{}
\centering
\tiny
\caption{Correlation test.\label{Table-19} }
\begin{tabular}{|p{1.1cm}<{\centering}|p{1.1cm}<{\centering}|p{0.9cm}<{\centering}|p{0.9cm}<{\centering}|p{0.9cm}<{\centering}|p{0.9cm}<{\centering}|p{1.2cm}<{\centering}|p{1.2cm}<{\centering}|p{1.3cm}<{\centering}|p{1.1cm}<{\centering}|p{1.1cm}<{\centering}|}
\hline
 Algorithm&     &$ f_1$  & $f_2$ & $f_3$ & $f_4$ & $f_5$ & $f_6$ & $f_7$ & $f_8$ & $f_9$  \\
\hline
 \multirow{2}{*}{DE-a }& $r$	&-0.163 &0.003	&  0.368    &  -0.328     &0.050	& 0.214	&-0.054	&	 0.117 &	0.872**  \\
                       & $p$    & 0.47  &0.988  &  0.092   &   0.136   &  0.825  &  0.338 & 0.813   &  0.604  &  0.000 \\
\hline
 \multirow{2}{*}{PSO-w }& $r$	&-0.184	&-0.179	&  0.469*    &  -0.104     &-0.108	& -0.056	&-0.243	 &	0.998** &	0.752**  \\
                       & $p$    & 0.411  &0.425  &  0.028    &  0.646     &  0.633  &  0.803 & 0.275  &  0.000  &  0.000 \\
\hline
 \multirow{2}{*}{CS}& $r$	&0.774**	&-0.071	&  0.588**    &  -0.466*    &0.126	& 0.727**	 &0.598**	&	0.397 &	0.774**  \\
                       & $p$    & 0.000  & 0.759 &  0.004   &   0.033   &  0.586  &  0.000 &  0.004  &  0.074  &  0.000 \\
\hline
\end{tabular}
\end{adjustwidth}
\end{table}

As demonstrated by our experiments, the relationship between $\overline \Delta$ and `$Dist$' for the DE-a algorithm is not very significant. The difference between the final solutions obtained by different initialization methods is due to the distribution characteristics of the initial population. In most cases, the PSO-w algorithm is not sensitive to the distribution range of the initial population. The average distance between the initial population and the real optimal has no positive correlation with the final solutions. {{The $\overline \Delta$ has a great influence on the performance of the CS algorithm. The closer the initial points are to the real optimal solution, the better the algorithm result. This may be part of the reason why the CS algorithm is sensitive to initialization.}} This is also in good agreement with the previous experimental results. On the influence of initialization on the performance of the three algorithms, we can conclude that CS $>$ PSO $>$ DE.

{{
\subsection{Experiments on ABC and GA}

The above experiments and analyses have focused on the three algorithms, and the conclusions have been drawn accordingly. In order to see if these conclusions are still valid for other algorithms, we  have carried out more tests on two other algorithms: the artificial bee colony (ABC) algorithm~\citep{karaboga2005idea} and the genetic algorithm (GA)~\citep{pal2017genetic}.

By using the same 22 initialization methods mentioned above, we have carried out some numerical experiments on the original ABC algorithm for all 19 test functions with the dimensionality of $D=30$. To make a fair comparison, the parameters of $NP$ and $limit$ are set to 50 and $D \cdot NP$ respectively~\citep{cui2018enhanced,lin2018novel}. Experiments of each initialization method have been executed independently for 20 times, and the maximum number of function evaluations (FEs) is set to 150000. In our experimental studies, the `Best', `Mean', `Var' and `Dist' values were recorded for the 9 basic functions to measuring the performance of the algorithm for each initialization method. In addition, the  `Best', `Mean', `Var' values were also recorded for the 10 CEC functions. Then, the experimental results are sorted out and analyzed in the same ways as discussed in the previous section, and the results of Friedman rank test of different initialization methods are given in Table~\ref{Table-25}.

\begin{table}[h]
\begin{adjustwidth}{-1.3cm}{}
\centering
\tiny
\caption{Friedman ranks of different initialization methods for the ABC algorithm.\label{Table-25} }
\begin{tabular}{|p{0.85cm}<{\centering}|p{0.85cm}<{\centering}|p{0.85cm}<{\centering}|p{1.1cm}<{\centering}|p{0.9cm}<{\centering}|p{0.9cm}<{\centering}|p{0.9cm}<{\centering}|p{0.9cm}<{\centering}|p{1.2cm}<{\centering}|p{1.2cm}<{\centering}|p{1.3cm}<{\centering}|p{1.1cm}<{\centering}|p{1.1cm}<{\centering}|}
\hline
\multirow{4}{*}{basic}
 &$p$ & $ Be(3,2)$  & $ Be(2.5,2.5)$ & $Be(2,3)$ & $U(0,1)$ & $N(0,1)$ & $N(0.5,1)$ & $N(0.5,0.5)$ & $logn(0,1)$ & $logn(.69,.25)$ & $logn(0,0.5)$ & $logn(0,2/3)$ \\
\cline{2-13}
 &0.00&  1& 3 & 5&  7& 17&12 &9 & 16 & 21& 15& 18\\
\cline{2-13}
 &  &$ E(0.5)$  & $E(0.1) $ & $E(0.8)$ & $Rayl(0.4)$ & $Rayl(0.8)$ & $Rayl(0.1)$ & $Weib(1,1.5)$ & $Weib(1.5,1)$ & $Weib(1,1)$ & $random$ & $LHS$ \\
\cline{2-13}
 & & 13 & 22 &  8 &   2 & 14  &20  &  10 & 19  & 11  &  6  &  4\\
\hline
\hline
\multirow{4}{*}{CEC}
&  $p$ & $ Be(3,2)$  & $ Be(2.5,2.5)$ & $Be(2,3)$ & $U(0,1)$ & $N(0,1)$ & $N(0.5,1)$ & $N(0.5,0.5)$ & $logn(0,1)$ & $logn(.69,.25)$ & $logn(0,0.5)$ & $logn(0,2/3)$ \\
\cline{2-13}
 & 0.00& 4 &	1 & 7	&	6&	21&20	&9	&16	&22	&11	&17	\\
\cline{2-13}
  &  &$ E(0.5)$  & $E(0.1) $ & $E(0.8)$ & $Rayl(0.4)$ & $Rayl(0.8)$ & $Rayl(0.1)$ & $Weib(1,1.5)$ & $Weib(1.5,1)$ & $Weib(1,1)$ & $random$ & $LHS$ \\
\cline{2-13}
  &  &12 	& 18 &8	&2	&10	&15	& 14  &19	&  13  &	5 &3	 \\
\hline
\end{tabular}
\end{adjustwidth}
\end{table}

 From Table~\ref{Table-25}, we can see that, for all 9 basic functions or CEC functions, the experimental results are basically the similar as before. Both the $p$-values are far less than 0.05, so the null hypothesis can be rejected, which indicates that ABC is greatly affected by initialization. For the ABC, the initialization methods: $ Be(2.5,2.5)$, $Rayl(0.4)$, $LHS$, $Be(3,2)$ seem to lead to better performance.

Similarly, we have carried out the same numerical experiments on the GA. It is worth pointing out that there are many GA variants, and the variant used in this paper is to keep half of the population with better fitness to be passed onto the next generation. The mutation probability has been set to 0.1. The maximum number of iterations and the population size are 100 and 3000, respectively. Each initialization method has been executed independently for 20 times for  all 19 test functions with $D=30$. The experimental results are summarized in Table~\ref{Table-26}.

\begin{table}[h]
\begin{adjustwidth}{-1.3cm}{}
\centering
\tiny
\caption{Friedman ranks of different initialization methods for the GA algorithm.\label{Table-26} }
\begin{tabular}{|p{0.85cm}<{\centering}|p{0.85cm}<{\centering}|p{0.85cm}<{\centering}|p{1.1cm}<{\centering}|p{0.9cm}<{\centering}|p{0.9cm}<{\centering}|p{0.9cm}<{\centering}|p{0.9cm}<{\centering}|p{1.2cm}<{\centering}|p{1.2cm}<{\centering}|p{1.3cm}<{\centering}|p{1.1cm}<{\centering}|p{1.1cm}<{\centering}|}
\hline
\multirow{4}{*}{basic}
 &$p$ & $ Be(3,2)$  & $ Be(2.5,2.5)$ & $Be(2,3)$ & $U(0,1)$ & $N(0,1)$ & $N(0.5,1)$ & $N(0.5,0.5)$ & $logn(0,1)$ & $logn(.69,.25)$ & $logn(0,0.5)$ & $logn(0,2/3)$ \\
\cline{2-13}
 &0.88&   2  &  12   &  9    &  17   &    19  &   22   &   1   &   8    &   6    &   4    &   7  \\
\cline{2-13}
 &  &$ E(0.5)$  & $E(0.1) $ & $E(0.8)$ & $Rayl(0.4)$ & $Rayl(0.8)$ & $Rayl(0.1)$ & $Weib(1,1.5)$ & $Weib(1.5,1)$ & $Weib(1,1)$ & $random$ & $LHS$ \\
\cline{2-13}
 &   &  5   &  10    &  20    &   21  &   3   &   14   &   15    &  18     &   11   &  13   &  16    \\
\hline
\hline
\multirow{4}{*}{CEC}
&  $p$ & $ Be(3,2)$  & $ Be(2.5,2.5)$ & $Be(2,3)$ & $U(0,1)$ & $N(0,1)$ & $N(0.5,1)$ & $N(0.5,0.5)$ & $logn(0,1)$ & $logn(.69,.25)$ & $logn(0,0.5)$ & $logn(0,2/3)$ \\
\cline{2-13}
 & 0.00&  3 &	5  &  4	&2	 &	22 &21	 &  8 	&18	  & 20  	&17	  & 16  	\\
\cline{2-13}
  &  &$ E(0.5)$  & $E(0.1) $ & $E(0.8)$ & $Rayl(0.4)$ & $Rayl(0.8)$ & $Rayl(0.1)$ & $Weib(1,1.5)$ & $Weib(1.5,1)$ & $Weib(1,1)$ & $random$ & $LHS$ \\
\cline{2-13}
  &  & 9	& 15 &11	&7	 &  10	&14	& 12  &19	& 13   &	1 &	6 \\
\hline
\end{tabular}
\end{adjustwidth}
\end{table}

As indicated by the results in Table~\ref{Table-26}, the Friedman rank test shown that for the 9 basic functions, the $p$-value is 0.878, which is greater than 0.05. Thus, we can essentially conclude that, for these problems, different initialization methods have little influence on the GA algorithm. For the complex CEC functions, the $p$ value is far less than 0.05. This shows that the GA is affected by initialization methods when problems are complex. However, the most appropriate initialization methods for the GA are Bate distribution, $LHS$, $random$, Uniform distribution.

}}

\subsection{Findings and implications}

{{The main contribution of this work is to study systematically the influence of different initialization approaches on algorithms, so as to gain some insight on this topic. Based on the extensive simulations and statistical tests, we can now discuss our findings and their implications.

One surprise finding is that some algorithms are more sensitive to initialization than others. However,
such sensitivity can also be problem-dependent. For example, differential evolution is not quite sensitive to initialization, while particle swarm optimization, cuckoo search and artificial bee colony algorithm are greatly affected by initialization. In addition, the genetic algorithm is less sensitive to initialization for many easy and smooth functions, but it becomes more sensitive to initialization for highly complex functions. }} Another surprise finding is that the commonly used technique in terms of uniform distributions for initialization is not necessarily the best approach. {{For example, for the PSO, our recommendation is to use the random, beta distribution and LHS as the main initialization methods. But for the CS, the preferred initialization methods are the beta distribution, exponential distribution, and Rayleigh distribution.}}

In addition, the population size can also have a significant effect. For the PSO, a larger population size is usually required, while a smaller population with more iterations can give better results for the DE. However, only a very small population size is sufficient for the CS. Furthermore, the above conclusions may also depend on the objective landscapes and thus may be problem-dependent, the correlation between the initialization methods and the premature convergence is very weak. Consequently, as long as the diversity of the population is high enough and the iterations are long enough, the optimal solutions can be found by all the algorithms.

Though these findings are preliminary, they do have some interesting implications. Firstly, for a given new algorithm, some parametric study is always needed to see if it is sensitive to initialization and its population size. Secondly, different initialization methods, especially a combination of uniform distributions and long-tailed distributions such as the exponential distribution and Rayleigh distribution
should be explored. Finally, different types of benchmark problems with different properties should be used to validate new algorithms, especially those with multimodal and optima-shifted functions.

Despite the above findings, we have not focused on how the selection mechanism of an algorithm may influence the diversity of the population in later iterations. In addition, almost all real-world problems have nonlinear constraints. We have not considered if the handling of constraints may affect the above findings. These will form the topics for further research.

\section{Conclusions}

Initialization has some significant influence on the performance of an algorithm for a given set of problems. {{In the current literature, the widely used initialization methods are the random methods, uniform distributions and LHS.}} However, there is no systematic comparison for different initialization techniques. In this paper, we have compared 22 different initialization methods based on different probability distributions for five algorithms over a set of 19 diverse benchmark functions. Based on our simulations and analyses, we can draw some conclusions:
\begin{itemize}
  \item {{The accuracy of the metaheuristic algorithm is closely related to two parameters: the population size and the maximum number of iterations. However, the dependence of different algorithms on their population sizes is different. Under the same conditions (the same total number of function evaluations), the DE-a algorithm usually requires a small population size with a larger number of iterations. A clear advantage of the CS is that it requires even a smaller population size, typically less than 100. On the other hand, the population size for the PSO-w should be larger, in comparison with both the DE and CS.}}

  \item The DE-a algorithm is not particularly sensitive to initialization, which means that DE is more robust. On the other hand, the PSO-w performs differently for different initialization methods, and the most appropriate initialization methods are random, beta distribution, and LHS. This may explain why many PSO variants performed well for uniform random initialization in the literature.
      Similarly, the CS is also sensitive to different initialization methods, and the most suitable methods for general problems are the beta distribution, followed by the Rayleigh and uniform distributions. For more complex problems, the most suitable initialization methods of the CS algorithm are the exponential distribution and Rayleigh distribution. {{Similarly,
      the ABC algorithm is sensitive to initialization, while the GA is less sensitive. However,
      such sensitivity can be problem-dependent. For example, the sensitivity of the GA is
      low for easy and
      smooth functions, while this sensitivity increases significantly for highly complex functions. }}

  \item The average distance between the initial population and the real optimal solution does not have any significant correlation with the quality of the final solution for the algorithms. That is to say, the final solutions obtained are not usually affected by the locality of the optimal solutions. Thus, as long as the diversity of the population is high and the number of iterations is large, all these algorithms are capable of finding the optimal solutions.
\end{itemize}

{{
The above conclusions are preliminary, and different algorithms can have different performance for different initialization methods and for different problems. Thus, there is a strong need to figure out what is the best initialization method for a given algorithm for a given set of problems. The present work paves a way for further investigation for a vast number of algorithms exist in the current literature. However, there are still some issues that need to be addressed so as to gain further insight into different algorithms and different initialization strategies. }} These can form the topics for further research.
\begin{itemize}
  \item For more complex problems, two or more distribution methods can be used as a hybrid, so as to
  enhance the overall diversity of the population. This should be tested using rigorous statistical techniques to see if they can indeed affect the statistical properties of the population of an algorithm at different stages of iterations.

  \item In all our tests, we have used the benchmark problems with simple bounds on regular domains. It would be useful to test more complicated problems with nonlinear constraints on irregular search domains to see if the same conclusions still hold. Such problems can be drawn from real-world applications.

  \item An automatic and self-adaptive method can be developed to automatically find the most suitable initialization method(s) for a given type of problems with a given algorithm. This may be attempted by following a similar approach as the self-tuning algorithm framework~\citep{Yang2013STA}. It is hoped that this work can inspire more studies concerning algorithm performance, robustness and different initialization techniques.
\end{itemize}

\section*{Acknowledgement}
This work was supported by the National Natural Science Foundation of China (Grant Nos.61877046 and 61877047), Shaanxi Provincial Natural Science Foundation of China (Grant No.2017JM1001).

\section*{Declaration of interests}
The authors declare that they have no known competing financial interests or personal relationships that could have appeared to influence the work reported in this paper.

\bibliographystyle{model1-num-names}
\bibliography{Qian_bib}

\end{document}